\pdfoutput=1
\documentclass[11pt]{article}

\usepackage[table]{xcolor}

\usepackage[preprint]{acl_latex}

\usepackage{times}
\usepackage{latexsym}

\usepackage[T1]{fontenc}

\usepackage[utf8]{inputenc}

\usepackage{microtype}

\usepackage{inconsolata}

\usepackage{graphicx}
\usepackage{amssymb} 

\definecolor{gold}{rgb}{0.83, 0.69, 0.22}
\definecolor{purple}{rgb}{0.54, 0.17, 0.89}
\usepackage{nicefrac}

\usepackage{multirow}
\usepackage{listings} 
\colorlet{punct}{red!60!black}
\definecolor{background}{HTML}{EEEEEE}
\definecolor{delim}{RGB}{20,105,176}
\colorlet{numb}{magenta!60!black}
\lstdefinelanguage{json}{
 basicstyle=\normalfont\ttfamily,
 numbers=left,
 numberstyle=\scriptsize,
 basicstyle=\scriptsize,
 stepnumber=1,
 numbersep=8pt,
 showstringspaces=false,
 breaklines=true,
 frame=lines,
 backgroundcolor=\color{background},
 literate=
 *{0}{{{\color{numb}0}}}{1}
 {1}{{{\color{numb}1}}}{1}
 {2}{{{\color{numb}2}}}{1}
 {3}{{{\color{numb}3}}}{1}
 {4}{{{\color{numb}4}}}{1}
 {5}{{{\color{numb}5}}}{1}
 {6}{{{\color{numb}6}}}{1}
 {7}{{{\color{numb}7}}}{1}
 {8}{{{\color{numb}8}}}{1}
 {9}{{{\color{numb}9}}}{1}
 {:}{{{\color{punct}{:}}}}{1}
 {,}{{{\color{punct}{,}}}}{1}
 {\{}{{{\color{delim}{\{}}}}{1}
 {\}}{{{\color{delim}{\}}}}}{1}
 {[}{{{\color{delim}{[}}}}{1}
 {]}{{{\color{delim}{]}}}}{1},
}
\usepackage[most,skins,theorems]{tcolorbox}
\usepackage{mathtools}
\tcbset{
 aibox/.style={
 width=\linewidth,
 top=8pt,
 bottom=4pt,
 colback=blue!6!white,
 colframe=black,
 colbacktitle=black,
 enhanced,
 center,
 attach boxed title to top left={yshift=-0.1in,xshift=0.15in},
 boxed title style={boxrule=0pt,colframe=white,},
 }
}
\newtcolorbox{dmbox}[2][]{aibox,title=#2,#1}

\usepackage{dsfont} 

\usepackage{booktabs}

\usepackage{subcaption}

\usepackage{enumitem}
\usepackage{gb4e}
\noautomath
\usepackage{bbm} 
\usepackage{amsmath}

\usepackage{xfrac}
\usepackage{cleveref}

\usepackage{pgfplots}
\usepgfplotslibrary{groupplots}
\usetikzlibrary{patterns}
\usepackage{ulem}

\definecolor{C_HIGHER}{HTML}{1E88E5}
\definecolor{C_LOWER}{HTML}{D81B60}
\definecolor{C_FIRST}{HTML}{FF8F07}
\definecolor{C_SECOND}{HTML}{004D40}
\definecolor{C_NULL}{HTML}{626464}
\definecolor{C_CALIBRATED}{HTML}{993893}
\definecolor{PURPLE}{HTML}{993893}
\definecolor{BLACK}{HTML}{000000}
\definecolor{goldenrod}{rgb}{0.85, 0.65, 0.13}

\usepackage{pgfplotstable}
\pgfplotsset{every tick label/.append style={font=\footnotesize},compat=1.17}
\pgfplotsset{compat = newest}

\usepgfplotslibrary{colorbrewer}

\definecolor{titlebg}{HTML}{111111} 
\definecolor{titletext}{HTML}{FEFEFE} 
\definecolor{boxbg}{HTML}{E6E6FA} 

\definecolor{PINK}{HTML}{FE6100}
\definecolor{BLUE}{HTML}{1A85FF}

\definecolor{RoyalBlue}{rgb}{0.25, 0.41, 0.88}
\definecolor{brickred}{rgb}{0.8, 0.25, 0.33}
\definecolor{forestgreen}{rgb}{0.13, 0.55, 0.13}

\definecolor{codegreen}{rgb}{0,0.6,0}
\lstdefinestyle{promptstyle}{  
    commentstyle=\color{codegreen},
    keywordstyle=\color{blue},
    basicstyle=\ttfamily\small,
    breakatwhitespace=false,        
    breaklines=true,                 
    captionpos=b,                    
    keepspaces=true,                 
    showspaces=false,                
    showstringspaces=false,
    showtabs=false,                  
    tabsize=2
}
\newcommand{\ccell}[2]{%
  \begingroup
  \setlength{\fboxsep}{1pt} 
  \colorbox{#1}{{\small #2}}
  \endgroup
}

\newcommand{\RC}{\ccell{PINK!25}}
\newcommand{\GC}{\ccell{BLUE!25}}
\newcommand{\WC}{\ccell{white}}

\newcommand{\mistral}[1]{\texttt{Mistral-7B-v0.{#1}}}
\newcommand{\mixtral}{Mixtral-8x7B-v0.1}
\newcommand{\yi}[1]{\texttt{Yi-1.5-{#1}B}}
\newcommand{\llama}{\texttt{Llama-3.1-8B}}
\newcommand{\gemma}[1]{\texttt{gemma-2-{#1}b}}
\newcommand{\gpt}[1]{\texttt{gpt-{#1}}}

\title{Language Model Probabilities are \\  \textit{Not} Calibrated in Numeric Contexts}

\newcommand{\spade}{{\tiny $\spadesuit$}}

\newcommand{\aspace}{\hspace{1em}}
\newcommand{\kensho}{$^{\heartsuit}$}
\newcommand{\adobe}{$^{\spadesuit}$}
\newcommand{\rit}{$^{\clubsuit}$}
\newcommand{\apple}{$^{\diamondsuit}$}
\newcommand{\xkensho}{$^{\dag}$}

\author{%
Charles J. Lovering\kensho \aspace\\
\textbf{Michael Krumdick}\kensho \aspace
  \textbf{Viet Dac Lai}\adobe{}\xkensho \aspace
  \textbf{Varshini Reddy}\kensho  \aspace\\ 
\textbf{Seth Ebner}\kensho \aspace
\textbf{Nilesh Kumar}\rit\xkensho \aspace
  \textbf{Rik Koncel-Kedziorski}\apple\xkensho \aspace
  \textbf{Chris Tanner}\kensho \aspace \\
 \kensho{} Kensho Technologies  \aspace
  \adobe{}Adobe \aspace \rit{} RIT \aspace \apple{}Apple \aspace\\
  \xkensho{} Previously at Kensho Technologies\\
\texttt{charles.lovering@kensho.com}
}

\begin{document}
\maketitle
\begin{abstract}
Some statements have one well-defined continuation (e.g., ``the Eiffel Tower is in [\textit{Paris}]''), whereas  others have a natural distribution over multiple options (e.g., ``the weighted coin flip was [\textit{Heads/Tails}].'') We argue that language model (LM) outputs should capture these natural distributions. Our work specifically tests whether LM output probabilities are \textit{calibrated} to numeric information within their textual contexts. For example, if the context (the prompt) concerns two equally likely options (e.g., heads or tails for a fair coin), the LM output probabilities should also be equal. Likewise, in a context with nonuniformly likely events (e.g., rolling a pair with two dice) an LM should output  proportionate probabilities. However, we find that even in simple settings, the best LMs (1) are poorly calibrated and (2) have systematic biases: artifacts like word identity, word order, and word frequency all impact calibration. For example, \texttt{gpt-4o-mini} often picks the first of two options presented in the prompt regardless of the options' implied likelihoods, whereas \texttt{Llama-3.1-8B} picks the second. Models do not  allocate probability mass \textit{among} valid options in a calibrated manner.

\end{abstract}

\section{Introduction}
We investigate if language model (LM) outputs are calibrated to the numeric content of their contexts. Consider the context below:
\begin{exe}
 \ex \label{intro:example1} {From \textbf{98} \textcolor{blue}{\textit{blue}} marbles and \textbf{99} \textcolor{red}{\textit{red}} marbles, Tommy reached blindly into a bag and grabbed a marble the color [\textit{\textcolor{blue}{\textit{blue}}/\textcolor{red}{\textit{red}}}]} 
 \end{exe}

\begin{figure}[h]
 \centering
 \includegraphics[width=\linewidth]{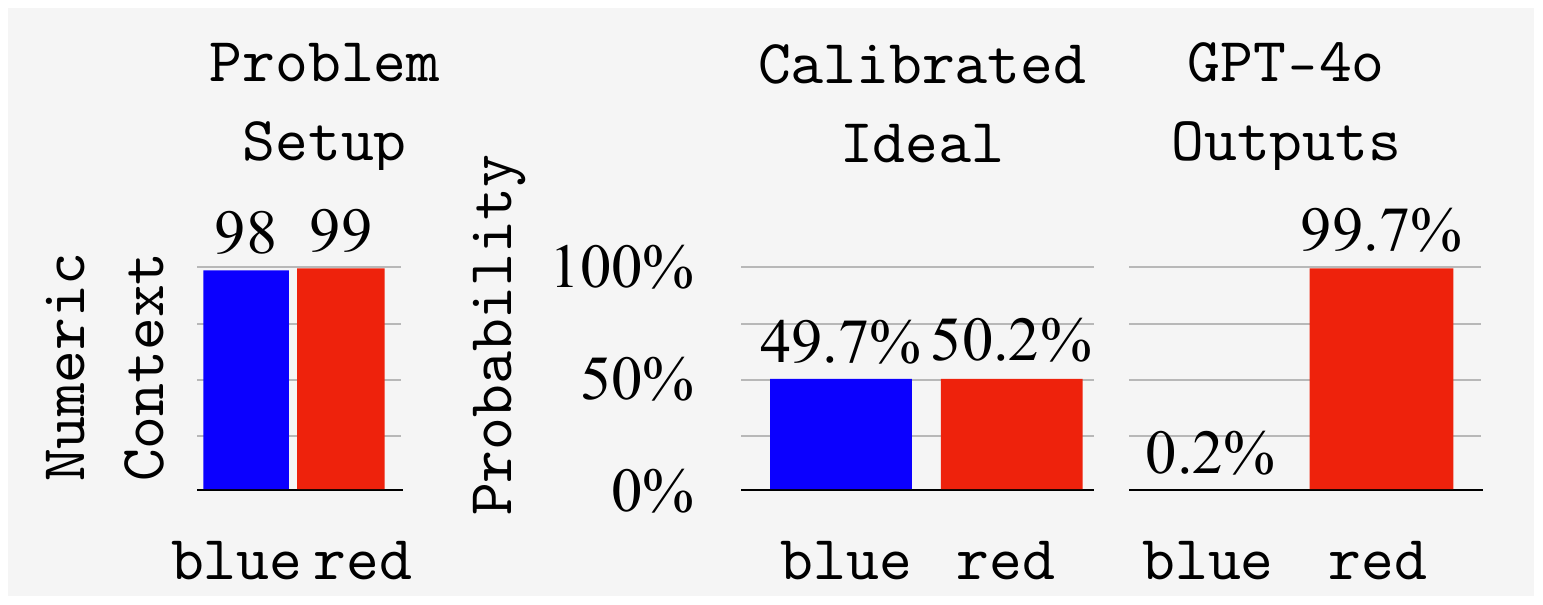}
 \caption{\textbf{Models are un-calibrated}. In this example, \gpt{4o} over-weights the option with a higher count of items beyond the calibrated probability, predicting \textcolor{red}{\textit{red}} with 99.7\% probability when 50.2\% is appropriate. We find consistent patterns of uncalibrated behavior.
}
 \label{fig:big-fig}
\end{figure}

This work tests the normative assumption that model probabilities for the next generated token should be calibrated to the relevant numeric content. For example, in \Cref{fig:big-fig}, the probability that an LM continues with one of the \textit{bracketed tokens} should be proportional to the respective \textbf{bolded} values. If the context defines or implies a probability to hold within the context, then subsequent text generation should condition on that probability.

\begin{dmbox}{What Do We Mean By Calibration?}
LMs' output probabilities are \textit{calibrated} to their contexts if the probabilities of relevant tokens correspond to the numeric content implicitly or explicitly defined in those contexts. If tokens $t_1, t_2, \dots, t_n$ are indicated by context $C$ to have probabilities $P = p_1, p_2, \dots, p_n$, a calibrated LM $m$ outputs corresponding probabilities $\pi_i = m(t_i | C)$: 
 \[ p_i \propto	 \pi_i,\quad i\- \in \- 
 1, 2, \dots, n. \]
We set calibration as the distance between $P$ and $\Pi = \pi_1, \pi_2, \dots, \pi_n$ (\Cref{sec:design:metrics}).
\end{dmbox}

Calibration within numeric contexts is a simple concept, but there is good reason to believe that models may struggle with it. Numbers within large pre-training datasets do not all occur with identical frequencies (e.g., numbers ending in 5 appear more than numbers ending in 7), as shown by \citep{zhou-etal-2023-navigating} for the Pile dataset and our findings with Dolmo \citep{soldaini2024dolma} (\Cref{sec:analysis:freq}). These differences may have downstream effects on LMs' biases across different numbers, making it difficult for language models to exhibit calibrated outputs.

Model calibration is related to and impacted by models' mathematical capabilities. Since language models are notably poor in their mathematical capabilities (compared to traditional Question-Answering tasks), it can be challenging for models to be calibrated properly. Roughly speaking, if a model struggles to answer basic math questions, it likely has poor representation and usage of numbers, which are requirements for being properly calibrated. Although models are quickly improving at mathematical reasoning \citep{lin2024rho,shao2024deepseekmath,mathstral2025,chervonyi2025goldmedalistperformancesolvingolympiad}, our work demonstrates that models are not well calibrated even in simple scenarios that require little reasoning.

Because LMs are increasingly used, calibrating LM outputs is increasingly important. We find motivation in a recent explanation method \citep{chen2023models}, Counterfactual simulatability, that requires explanations to capture the ``reason'' a model arrived at an answer. There should be an analogous expectation for model output probabilities. Not reaching this calibration standard may not impact a user over a single interaction; any given outcome is compatible with a calibrated distribution (assuming some probability mass on all options).

However, failed calibration may have an impact across multiple users or repeat interactions. For example, an uncalibrated model may always recommend visiting a particular restaurant despite the city having hundreds of equally good options. This model's calibration failure could result from something as idiosyncratic as the restaurant's name. In this scenario, the uncalibrated model negatively impacts many parties (e.g., the other restaurants, customers, etc). This restaurant example is simply one of the many scenarios where users rely on accurate and helpful recommendations (e.g., for movies, job candidates, vacations, colleges, etc). Having correctly calibrated models is critical for situations involving medical or health services, such as making radiology predictions \citep{shreekumar2025x}.

We find that many LMs are poorly calibrated in numeric contexts. Moreover, they fail in systematic ways. In cases like Example \ref{intro:example1}, some uncalibrated models prefer one particular option based on position (e.g., always the first option). Other uncalibrated models may consistently pick the option with the higher number. Word identity of the options and the vocabulary frequency of the numeric information appear to impact these order biases and, thus, the calibration. Our work demonstrates that LMs are often consistently biased in these simple ways. This suggests that there is a risk in relying on LMs for scenarios that involve probabilistic outcomes.

\begin{dmbox}{Contributions}
\begin{enumerate}[leftmargin=*]
 \item Instruction-tuning helps by increasing the model's cumulative probability for all relevant options. Yet, models generally remain poorly calibrated -- sometimes worse than simple baselines.
 \item Models exhibit mode collapse: instruction-tuning overly reduces the entropy of the output distribution.
 \item Models exhibit systematic biases over word identity and order, often overriding the numeric information in the context.
 \item Frequency effects of the number tokens in the training data impact calibration.
\end{enumerate}
\end{dmbox}

\section{Related Work}
\paragraph{Models struggle to calibrate internal uncertainty with textual outputs.}
A related line of work studies how well models calibrate prediction confidence with prediction accuracy: a calibrated model with $0.80$ confidence on each prediction should classify $80\%$ of examples correctly \citep{guo2017calibration, minderer2021revisiting}. Under this definition, \citet{wei2024measuring} reports GPT models have good calibration, whereas, \citet{phan2025humanity} reports the opposite.
Another line of work investigates how well model outputs align with uncertainty \citep{yona2024can,kumar2024confidence,zhou-etal-2023-navigating,lin2022teaching}, termed linguistic calibration \citep{mielke2022reducing}. \citet{yona2024can} reports that models struggle to exhibit internal uncertainty in text. \citet{kumar2024confidence} measures how well confidence is aligned internally (model probabilities) and externally (via eliciting a model to choose among a Likert scale). In our work, we look at the agreement between the probability of predicted outputs and the contextually defined likelihood of outputs in the text. We find that models are often internally \textit{certain} when they should be \textit{uncertain}; this {mode collapse} arises both via supervised fine-tuning and reward-based fine-tuning \citep{oattributing, janus2024modecollapse}. \citet{zhou-etal-2024-relying} also finds that reward-based fine-tuning biases models towards certainty, generating strengtheners (e.g., ``I am certain\dots'') over weakeners (e.g., ``Maybe\dots'').
Notably, faithful calibration is not always desirable; to simplify our experiments, we set aside phenomena like hyperbole and pragmatic effects \citep{tsvilodub2025non}.

\paragraph{Recent work finds that language models struggle to simulate randomness:} coin flip predictions are biased in favor of heads and the first outcome mentioned in context \citep{van2024random}.
\citet{hopkins2023can}, similar to our work, shows that LMs struggle to induce uniform distributions in the 0 to 1 interval.

\paragraph{Biases in language models.} There are large-scale efforts to remove biases from models \citep{rudinger2018,bai2022training}. \citet{pezeshkpour-hruschka-2024-large} shows that LMs prefer multiple-choice options based on their position. Using methods similar to our's, \citet{kusner2017counterfactual} studies probability differences in the setting of counterfactual fairness. We find evidence of similar biases impacting calibration.

\section{Experimental Design}\label{sec:design}
\subsection{Problem Setup}\label{sec:design:problem_setup}
We introduce three templated datasets. Each problem instance is a context $C$ to be continued by a relevant token among $t_1, t_2, \dots, t_n = T \subset V$, where $V$ is the full vocabulary. Each token $t_i$ is associated with an implied probability $p_i$, forming a distribution over $T$, $P = p_1, p_2, \dots, p_n$. In Example \ref{intro:example1}, $T$ is $\{\textrm{\textit{red}}, \textrm{\textit{blue}}\}$ with $P = \{\textrm{\textit{red}} : \sfrac{98}{197}, \textrm{\textit{blue}} : \sfrac{99}{197}\}$. For a model $m$, we define $\Pi = \pi_1, \pi_2, \dots, \pi_n$ where $\pi_i \approx m(t_i | C)$, with one modification. We sum the probabilities of common tokenizations for a given word, namely, capitalization and spaces: $\pi_i = \sum_{s \in \textrm{Tokenizations}(t_i)} m(s | C)$. For example, for the option \textit{red}, we sum probabilities for ``red'', ``Red'', ``\_red'', and ``\_Red.'' This approach does not account for the problem of there being many ways of grammatically continuing the sentence. However, empirically, instruction-tuned models put almost all probability mass on $T$. This admittedly imperfect approach still allows us to study the problem meaningfully.\footnote{See \Cref{app:details} for details on reproducibility.}

\subsection{Metrics and Evaluation}\label{sec:design:metrics}
We use three metrics to test if models calibrate to numeric contexts and where they go wrong.

\textbf{Probability Mass (PM)}: A model's calibration, the relative probability mass across tokens in \textit{T}, is well-founded only if there is sufficient probability mass upon the tokens in $T$. We measure this directly with PM,
$\textrm{PM}(T) \doteq \sum_{t\,\in\,T} \pi_t.$
If $\textrm{PM}(T)$ for a model is low (say, 0.30), then it is unclear if the model is capturing the intended relationship between the context and $T$. While meaningful distribution across $T$ could still exist if these values were normalized, we set this case aside. When $\textrm{PM}$ is high (say, $>0.75$), we can start to meaningfully ask questions about how probability mass is allocated among, say, \textit{red} and \textit{blue}. 

\textbf{Wasserstein Distance (WD)}: To measure {calibration}, we use {WD} \citep{kantorovich1939mathematical}, which captures the movement between one distribution (or set of values) and another.\footnote{Kullback–Leibler (KL) divergence is also a natural choice. However, some problems in our datasets include options with the value zero where KL would be undefined. We also compute Mean-Squared Error (MSE) but don't find a material difference in the findings so we only report the WD.} We use SciPy's \texttt{wasserstein\_distance\_nd} implementation.\footnote{\url{https://docs.scipy.org/doc/scipy/reference/generated/scipy.stats.wasserstein\_distance\_nd.html}}

\textbf{Relative Entropy (RE):} To help understand a model's behavior, we propose $\textrm{RE} = \textrm{H}(\Pi) - \textrm{H}(P)$, the difference in entropy $\textrm{H}$ between model and ideal calibrated probabilities. $\textrm{RE}<0$ means the model's probabilities are overly concentrated and $\textrm{RE}>0$ means they are overly diffuse.

\subsection{Models} Our experiments assume access to model logits.
We test range of open-source models with both base and instruction-tuned versions available: \mistral{1}, \mistral{3}, \mixtral{}, \yi{9}, \yi{34}, \llama{}, \gemma{9}, \gemma{27} \citep{jiang2023mistral,jiang2024mixtral,young2024yi,dubey2024llama,team2024gemma}; and four proprietary models, \gpt{3.5}, \gpt{4-turbo}, \gpt{4o-mini}, and \gpt{4o} \citep{brown2020language,achiam2023gpt,openai2023gpt4o}. We prompt instruction-tuned models to calibrate outputs to the numbers in context. We refer to the base version of the models as ``Base'' and the instruction-tuned, prompted models as ``Chat.''

\subsection{Datasets}
We introduce three datasets \textit{colors}, \textit{wordproblems}, and \textit{distributions}. The following are representative examples, respectively: \footnote{Our datasets test basic ratios and probability. As models improve, it would be exciting if datasets targeted concepts like Bayes' theorem, De Morgan's laws, event independence, etc.}

\begin{exe}
 \ex \label{experiments:ex2:1} {From \textbf{17} \textit{red} marbles and \textbf{99} \textit{blue} marbles, Billy reached blindly into the bag and grabbed a marble with the color $[\textit{red}/\textit{blue}]$}
\ex \label{experiments:ex2:2} There was a grove with \textbf{17} \textit{spruce} and \textbf{99} \textit{cedar}. During a thunderstorm lightning struck a tree. All the trees were about the same height and elevation. Turned out, the species of the tree was [\textit{spruce/cedar}]
\ex \label{experiments:ex2:3} Sampling from a uniform distribution with support from \textbf{2} inclusive to \textbf{5} exclusive, I sampled the number $[\textit{0}/\textit{1}/\textit{\textbf{2}}/\textit{\textbf{3}}\dots\textit{9}]$
\end{exe}

For \textbf{(a)} \textit{colors}, we use 5 templates, 3 numeric scales with 100 configurations each, and 110 permutations of color options (e.g., \textcolor{red}{red}/\textcolor{blue}{blue} and \textcolor{blue}{blue}/\textcolor{red}{red}).\footnote{Calibration could be well-studied by sampling non-exhaustively across these permutations but doing so lets us highlight biases across different options.
} The numeric scales are (1) numbers 1 to 10, (2) ten numbers sampled from 10 to 100, and (3) ten numbers sampled from 100 to 999. Pilot experiments with larger numbers ($>1000$) did not see marked differences in calibration. Still, future work should study a wider range of numbers. For \textbf{(b)} \textit{wordproblems}, there are 10 templates, the same 3 numeric scales with 100 configurations each, and 4-10 pairs of options (e.g., \textit{spruce/cedar}, or \textit{nurse/doctor}) depending on the template. For \textbf{(c)} \textit{distributions}, the configuration is different. The task is to pick a number uniformly from an interval. The valid options in \textit{T} are all integers in the interval. There are 5 templates, different range inclusivities, and 320 pairs of numbers that define the intervals ranging from 10 to 1000. Each problem is set to occur within an interval length of less than 10 and requires only a single token prediction.

When we compute standard errors we cluster option pairs together \citep{miller2024adding}. All our results show the 95\% confidence interval (not the standard error) either explicitly or using $\pm$ notation.

\begin{table}[ht!]
\small
    \centering
    \begin{tabular}{@{}l@{}rrr@{}}
    \toprule
    Dataset & \# Problems & \# Option Sets & Entropy (bits) \\
    \midrule
    \textit{colors}         &   165.0K & 55 (110) & $0.81_{0.06}$\\
    \textit{wordproblems}   &   33.6K & 55 (110) & $0.81_{ 0.06}$\\
    \textit{distributions}   &   4.5K & 320 & $2.27_{0.17}$\\
    \bottomrule
    \end{tabular}
    \caption{\textbf{Datasets.} The first two datasets have 55 pairs of options; 110 when order is considered. For the the third, there are 320  pairs of numbers that define the distribution intervals. The first two datasets use the same underlying numbers, sharing the same average entropy per problem instance. The entropy is shown as an average and sub-scripted by the variance across problem instances. \textit{Distributions} problems are across more options (2+) and expect a uniform distribution.} 
    \label{tab:dataset}
\end{table}

\subsection{Reference Behaviors and Baselines}
\noindent\textit{Reference Behaviors.} Understanding a model's behavior across many instances is difficult. However, models typically produce one of six behaviors in practice, making them a useful way to summarize model behavior. We categorize the model behavior into these as reference points.

Summaries are limited: \gpt{4o-mini}, when the color \textit{white} is listed second in the prompt, tends to pick the  colo listed first. This is classified as Pick First but could be understood instead as ``pick the first color before options starting with \textit{w}.'' Our reference points describe behavior compatible with a model's behavior over many examples and settings, but it is still only descriptive. Finally, for a single instance, there are multiple valid explanations (e.g. in Example \ref{intro:example1} could be Pick Higher or Pick Lower). We can only determine the behavior over a large set of balanced instances.
\begin{enumerate}[itemsep=0em]
 \item [0.] \texttt{Null:} PM is closer to zero than to any reference behavior.
 \item [1.] \texttt{Calibrated:} Sets $\Pi = P$, the best case.
 \item [2.] \texttt{Pick Higher:} Sets all probability mass on $\arg\max P$.
 \item [3.] \texttt{Pick lower:} analogously sets all probability mass on $\arg\min P$.
 \item [4.] \texttt{Pick First:} Sets all probability mass on the first option, ignoring numeric information.
 \item [5.] \texttt{Pick Second:} analogously sets all probability mass on the second option.
\end{enumerate}

\noindent\textit{Baselines.} We introduce two more baselines besides using the reference behaviors above as baselines. First, $\texttt{Pick Higher}_{p = q}$ places $q$ (say, 0.7) on $\arg\max P$ and the remainder of the probability mass evenly across other tokens in $T$. Second, \texttt{Random} baselines to contextualize our results. For every problem in the dataset, the random baseline $\textrm{Random}_{\tau}$ produces a $\Pi$ by uniformly sampling two random numbers from 0 to 1, which are then transformed into a distribution parameterized by a temperature $\tau$ using the softmax function.\\

\noindent\textit{Example with Baseline Scores.} \Cref{tab:baseline:performances} provides baseline scores for a single example (numbers 51 and 98) to provide reference metric scores. Beyond showing how WD scores range, this example again highlights how different baselines (say, Pick First and Pick Higher) will have the same behavior for some instances; 
 references are only delineated over a pool of examples were the behaviors are mutually exclusive.  See \Cref{app:results:demo} to see how different models behave on this example. 
\begin{table}[ht!]
    \centering
    \small
\begin{tabular}{@{}lr@{\hspace{4pt}}r@{\hspace{4pt}}r@{\hspace{4pt}}r@{\hspace{4pt}}r@{}}
\toprule
{} &    $P(t_1)$ &    $P(t_2)$ &  PM &  RE &  WD \\
\midrule
Ideal & 0.34 & 0.66 & 1.00& 1.00& 0.00 \\
\midrule
Null    &  0.00 &  0.00 &      0.00 &      \textit{n/a} &         0.74 \\
Pick First                &  1.00 &  0.00 &      1.00&            -0.93 &          0.93 \\
Pick Lower                &  1.00 &  0.00 &      1.00&            -0.93 &          0.93 \\
Pick Second               &  0.00 &  1.00 &      1.00&            -0.93 &          0.48 \\
Pick Higher               &  0.00 &  1.00 &      1.00&            -0.93 &          0.48 \\
Pick Higher$_{p =0.7}$    &  0.30 &  0.70 &      1.00&            -0.05 &          0.06 \\
Pick Higher$_{p =0.6}$    &  0.40 &  0.60 &      1.00&             0.04 &          0.08 \\
Pick Higher$_{p =0.8}$    &  0.20 &  0.80 &      1.00&            -0.21 &          0.20 \\
Pick Higher$_{p =0.9}$    &  0.10 &  0.90 &      1.00&            -0.46 &          0.34 \\
\bottomrule
\end{tabular}
\caption{Each row shows the performance of the given baseline an example with numbers 51 and 98. This table is meant to be a helpful reference to build an intuition about performance score magnitudes.}
\label{tab:baseline:performances}
\end{table}

\section{Results}\label{sec:results:exp2}

\subsection{Probability Mass}
First, we closely examine the probability mass (PM) on the relevant tokens. Consider Example \ref{intro:example1} where \gpt{4o} is to pick either \textit{Red} or \textit{Blue}.
There, a high PM means that the model successfully allocated probability mass over these tokens. Success depends on the design of our problem scenarios and model capability. The high scores suggest that the models correctly interpret the basic task.

\begin{table}[ht!]
\small
\centering
\begin{tabular}{@{}lr@{\hspace{4pt}}r@{\hspace{4pt}}r@{}}
\toprule
    &  \multicolumn{1}{c}{Base} &   \multicolumn{1}{c}{Chat}   &     Chat - Base \\
    &   &     &     $\Delta$ (95\% CI) \\
    \midrule
\textit{colors} \\ 
Llama-3.1-8B &  0.38$_{\pm{0.01}}$ &  0.80$_{\pm{0.00}}$ &    \GC{\small (+0.41, +0.43)} \\
Mistral-7B-v0.1   &  0.30$_{\pm{0.01}}$ &  0.54$_{\pm{0.01}}$ &    \GC{\small (+0.22, +0.25)} \\
Mistral-7B-v0.3   &  0.33$_{\pm{0.01}}$ &  0.76$_{\pm{0.02}}$ &    \GC{\small (+0.42, +0.44)} \\
Mixtral-8x7B &  0.36$_{\pm{0.01}}$ &  0.99$_{\pm{0.00}}$ &    \GC{\small (+0.62, +0.63)} \\
Yi-1.5-9B         &  0.31$_{\pm{0.01}}$ &  0.99$_{\pm{0.00}}$ &     \GC{\small (+0.67, +0.70)} \\
Yi-1.5-34B        &  0.42$_{\pm{0.01}}$ &  0.62$_{\pm{0.02}}$ &    \GC{\small (+0.17, +0.22)} \\
gemma-2-9b        &  0.39$_{\pm{0.01}}$ &  0.99$_{\pm{0.00}}$ &    \GC{\small (+0.59, +0.61)} \\
gemma-2-27b       &  0.54$_{\pm{0.01}}$ &  1.00$_{\pm{0.00}}$ &    \GC{\small (+0.45, +0.46)} \\
gpt-3.5           &                  - &  1.00$_{\pm{0.00}}$ &               - \\
gpt-4-turbo       &                  - &  1.00$_{\pm{0.00}}$ &               - \\
gpt-4o-mini       &                  - &  1.00$_{\pm{0.00}}$ &               - \\
gpt-4o            &                  - &  1.00$_{\pm{0.00}}$ &               - \\
\midrule
\textit{wordproblems} \\ 
Llama-3.1-8B &  0.54$_{\pm{0.05}}$ &  0.86$_{\pm{0.03}}$ &    \GC{\small (+0.25, +0.38)} \\
Mistral-7B-v0.1   &  0.44$_{\pm{0.04}}$ &  0.69$_{\pm{0.03}}$ &    \GC{\small (+0.22, +0.28)} \\
Mistral-7B-v0.3   &  0.42$_{\pm{0.04}}$ &  0.84$_{\pm{0.06}}$ &    \GC{\small (+0.33, +0.51)} \\
Mixtral-8x7B&  0.57$_{\pm{0.03}}$ &  0.97$_{\pm{0.02}}$ &    \GC{\small (+0.37, +0.44)} \\
Yi-1.5-9B         &  0.58$_{\pm{0.03}}$ &  0.88$_{\pm{0.03}}$ &    \GC{\small (+0.25, +0.36)} \\
Yi-1.5-34B        &  0.53$_{\pm{0.05}}$ &  0.72$_{\pm{0.06}}$ &    \GC{\small (+0.14, +0.25)} \\
gemma-2-9b        &  0.63$_{\pm{0.04}}$ &  0.98$_{\pm{0.01}}$ &     \GC{\small (+0.31, +0.40)} \\
gemma-2-27b       &  0.59$_{\pm{0.03}}$ &  1.00$_{\pm{0.00}}$ &    \GC{\small (+0.38, +0.45)} \\
gpt-3.5           &                  - &  0.60$_{\pm{0.05}}$ &               - \\
gpt-4-turbo       &                  - &  0.46$_{\pm{0.07}}$ &               - \\
gpt-4o-mini       &                  - &  0.55$_{\pm{0.05}}$ &               - \\
gpt-4o            &                  - &  0.60$_{\pm{0.05}}$ &               - \\
\midrule
\textit{distributions} \\ 
Llama-3.1-8B &  0.82$_{\pm{0.00}}$ &  0.78$_{\pm{0.01}}$ &  \RC{\small (-0.06, -0.04)} \\
Mistral-7B-v0.1   &  0.92$_{\pm{0.00}}$ &  0.95$_{\pm{0.00}}$ &    \GC{\small (+0.03, +0.03)} \\
Mistral-7B-v0.3   &  0.91$_{\pm{0.00}}$ &  1.00$_{\pm{0.00}}$ &    \GC{\small (+0.09, +0.09)} \\
Mixtral-8x7B &  0.96$_{\pm{0.00}}$ &  1.00$_{\pm{0.00}}$ &    \GC{\small (+0.03, +0.04)} \\
Yi-1.5-9B         &  0.90$_{\pm{0.00}}$ &  0.96$_{\pm{0.01}}$ &    \GC{\small (+0.06, +0.07)} \\
Yi-1.5-34B        &  0.95$_{\pm{0.00}}$ &  0.91$_{\pm{0.02}}$ &  \RC{\small (-0.06, -0.02)} \\
gemma-2-9b        &  0.84$_{\pm{0.01}}$ &  0.97$_{\pm{0.01}}$ &    \GC{\small (+0.13, +0.15)} \\
gemma-2-27b       &  0.96$_{\pm{0.00}}$ &  1.00$_{\pm{0.00}}$ &    \GC{\small (+0.03, +0.04)} \\
gpt-3.5           &                  - &  0.99$_{\pm{0.00}}$ &               - \\
gpt-4-turbo       &                  - &  1.00$_{\pm{0.00}}$ &               - \\
gpt-4o-mini       &                  - &  0.68$_{\pm{0.01}}$ &               - \\
gpt-4o            &                  - &  0.95$_{\pm{0.01}}$ &               - \\
\bottomrule
\end{tabular}

\caption{\textbf{Probability Mass Results} for the base and chat versions of models, along with 95\% confidence intervals of their paired difference.    Statistically significant increases in probability mass are \textcolor{BLUE}{highlighted blue (better)}; decreases are \textcolor{PINK}{highlighted orange (worse)}.}
\label{tab:results:pm}
\end{table}

\Cref{tab:results:pm} reports PM for the base and chat versions of all tested models. \textbf{Instruction-tuned models have statistically significantly higher PMs over their base versions.} This can be seen by the strictly positive 95\% confidence intervals of the difference between the chat and base scores in the rightmost column. Except for \llama{} and \yi{34} on \textit{distributions}, this holds for all models. The instruction-tuned PM scores are all reasonably high, suggesting that the datasets' sentence templates are well-formed. The primary exception is \mistral{1}. Notably, \gpt{*} models had lower PM on \textit{wordproblems}.

\begin{table}[ht!]
\small
\centering
\begin{tabular}{@{}lrrr@{}}
\toprule
  &            \multicolumn{1}{c}{\textit{colors}} & \multicolumn{1}{c}{\textit{wordp.}} & \multicolumn{1}{c}{\textit{ distr.}} \\
\midrule
Pick Higher      &  0.47$_{\pm0.01}$ &  0.44$_{\pm0.00}$ &               - \\
Pick $\textrm{Higher}_\textrm{p=0.7}$  &  0.15$_{\pm0.01}$ &  0.17$_{\pm0.00}$ &               - \\
Pick Lower       &  0.95$_{\pm0.01}$ &  0.98$_{\pm0.00}$ &               - \\
Pick First/Second      &  0.71$_{\pm0.02}$ &  0.71$_{\pm0.00}$ &              0.38$_{\pm0.00}$ \\
$\textrm{Random}_{\tau = 0.01}$    &  0.71$_{\pm0.02}$ &  0.69$_{\pm0.00}$ &  0.86$_{\pm0.00}$ \\
$\textrm{Random}_{\tau = 1.00}$     &  0.27$_{\pm0.01}$ &  0.29$_{\pm0.00}$ &  0.38$_{\pm0.00}$ \\
\midrule
Llama-3.1-8B & 0.40$_{\pm0.01}^\spadesuit$ & 0.48$_{\pm0.02}$ & 0.43$_{\pm0.01}$ \\
  Mistral-7B-v0.1 & 0.50$_{\pm0.01}$ & 0.49$_{\pm0.01}$ & 0.22$_{\pm0.01}$ \\
  Mistral-7B-v0.3 & 0.48$_{\pm0.02}$ & 0.55$_{\pm0.01}$ & 0.56$_{\pm0.01}$ \\
Mixtral-8x7B & 0.51$_{\pm0.01}$ & 0.60$_{\pm0.03}$ & 0.71$_{\pm0.01}$ \\
        Yi-1.5-9B & 0.49$_{\pm0.01}$ & 0.56$_{\pm0.01}$ & 0.61$_{\pm0.01}$ \\
       Yi-1.5-34B & 0.55$_{\pm0.01}$ & 0.57$_{\pm0.01}$ & 0.62$_{\pm0.01}$ \\
       gemma-2-9b & 0.50$_{\pm0.01}$ & 0.52$_{\pm0.02}$ & 0.59$_{\pm0.01}$ \\
      gemma-2-27b & 0.40$_{\pm0.00}^\spadesuit$ & 0.48$_{\pm0.01}$ & 0.59$_{\pm0.01}$ \\
          gpt-3.5 & 0.30$_{\pm0.01}^\spadesuit$ & 0.57$_{\pm0.02}$ & 0.42$_{\pm0.01}$ \\
      gpt-4-turbo & 0.42$_{\pm0.01}^\spadesuit$ & 0.62$_{\pm0.02}$ & 0.69$_{\pm0.01}$ \\
      gpt-4o-mini & 0.40$_{\pm0.01}^\spadesuit$ & 0.57$_{\pm0.02}$ & 0.57$_{\pm0.01}$ \\
           gpt-4o & 0.40$_{\pm0.00}^\spadesuit$ & 0.57$_{\pm0.02}$ & 0.49$_{\pm0.01}$ \\

\bottomrule
\end{tabular}
\caption{\textbf{Calibration Results.} This table shows calibration scores (WD) for Chat models along with 95\% confidence intervals ($\pm$). For \textit{colors}, six models have a statistically significantly better calibration than the baseline Pick Higher, denoted {\scriptsize${ \spadesuit}$}. For \textit{wordproblems}, no model is better than this simple baseline.
}
\label{tab:results:wd}
\end{table}

\subsection{Calibration}
Next, we examine the calibration scores of instruction-tuned models in \Cref{tab:results:wd}. Given the low PM scores of the base models, we leave those results to Appendix \Cref{tab:results:wd:all}. Our high-level finding is simple: all tested models are poorly calibrated, and their performance is typically worse than simple baselines. Some models are mildly better than a baseline that sets 100\% probability to the option associated with the higher number, but most are not. Another baseline, Pick $\textrm{Higher}_{p = 0.7}$, which allocates 70\% probability to the option associated with the higher number, outperforms all models. This result, more than anything, describes the best average output probabilities for \textit{colors} and \textit{wordproblems}.
This all goes to say: \textbf{Although the models can allocate probability mass on valid options, they cannot properly allocate probability mass \textit{among} these options in a calibrated manner.}
In \Cref{app:results:reproducibility}, we find evidence that our calibration results are stable across a broader range of templates and template variations.

\subsection{Relative Entropy}
\begin{table}[ht!]
\small
\centering
\begin{tabular}{@{}lrrr@{}}
\toprule
  &            \multicolumn{1}{c}{\textit{colors}} & \multicolumn{1}{c}{\textit{wordp.}} & \multicolumn{1}{c}{\textit{ distr.}} \\
\midrule
Pick $\textrm{Higher}_\textrm{p=0.7}$  &   0.02$_{\pm0.01}$ &   0.07$_{\pm0.00}$ &               - \\
Pick Higher / Lower       &  -0.86$_{\pm0.01}$ &  -0.81$_{\pm0.00}$ &               - \\
Pick First / Second     &  -0.86$_{\pm0.01}$ &  -0.81$_{\pm0.00}$ &  -2.26$_{\pm0.01}$ \\
$\textrm{Random}_{\tau = 0.01}$    &  -0.80$_{\pm0.01}$ &  -0.76$_{\pm0.00}$ & -2.07$_{\pm0.01}$ \\
$\textrm{Random}_{\tau = 1.0}$     &   0.11$_{\pm0.01}$ &   0.16$_{\pm0.00}$ & 1.35$_{\pm0.01}$  \\
\midrule
Llama-3.1-8B & -0.11$_{\pm0.02}$ & -0.33$_{\pm0.06}$ & -0.70$_{\pm0.03}$ \\
  Mistral-7B-v0.1 & -0.21$_{\pm0.02}$ & -0.24$_{\pm0.04}$ &  0.05$_{\pm0.02}$ \\
  Mistral-7B-v0.3 & -0.36$_{\pm0.03}$ & -0.47$_{\pm0.04}$ & -1.07$_{\pm0.04}$ \\
Mixtral-8x7B-v0.1 & -0.67$_{\pm0.01}$ & -0.65$_{\pm0.05}$ & -1.59$_{\pm0.04}$ \\
        Yi-1.5-9B & -0.55$_{\pm0.01}$ & -0.55$_{\pm0.03}$ & -1.13$_{\pm0.05}$ \\
       Yi-1.5-34B & -0.45$_{\pm0.02}$ & -0.53$_{\pm0.04}$ & -1.32$_{\pm0.04}$ \\
       gemma-2-9b & -0.44$_{\pm0.01}$ & -0.45$_{\pm0.03}$ & -1.11$_{\pm0.04}$ \\
      gemma-2-27b & -0.63$_{\pm0.01}$ & -0.57$_{\pm0.02}$ & -1.14$_{\pm0.05}$ \\
          gpt-3.5 & -0.35$_{\pm0.03}$ & -0.68$_{\pm0.04}$ & -0.69$_{\pm0.02}$ \\
      gpt-4-turbo & -0.65$_{\pm0.01}$ & -0.69$_{\pm0.04}$ & -1.54$_{\pm0.03}$ \\
      gpt-4o-mini & -0.53$_{\pm0.01}$ & -0.58$_{\pm0.05}$ & -1.30$_{\pm0.02}$ \\
           gpt-4o & -0.61$_{\pm0.00}$ & -0.63$_{\pm0.04}$ & -0.92$_{\pm0.02}$ \\
\bottomrule
\end{tabular}

\caption{\textbf{Relative Entropy (RE) Results.} The RE is far below a calibrated level across all datasets.}
\label{tab:results:re}
\end{table}
The section above demonstrated that the models cannot allocate the probability mass \textit{among} the valid options in a calibrated manner. Here, as shown in \Cref{tab:results:re}, we find that these models fail primarily by over-allocating probability mass on one of the two options: mode collapse. Every model's RE was statistically significantly less than the calibrated result, often with large effects for \textit{colors} and \textit{wordproblems}. Models with the best calibration scores (gpt-* models) also have the lowest relative entropy, suggesting that no model approaches well-calibrated behavior.

Instruction-tuning appears to cause this mode collapse. Across all three datasets, there is a drastic reduction in entropy between the base and chat model versions, on average, 0.50/0.36/1.19 bits across the three datasets respectively. This drop in bits reduces the entropy to 47/42/55\% percent of the entropy of the ideal calibrated result. (See \Cref{app:results:mode} for more details.)

\section{Analysis: Option Identity and Order}\label{sec:analysis:option}
\begin{figure}[ht!]
 \centering
 \includegraphics[width=\linewidth]{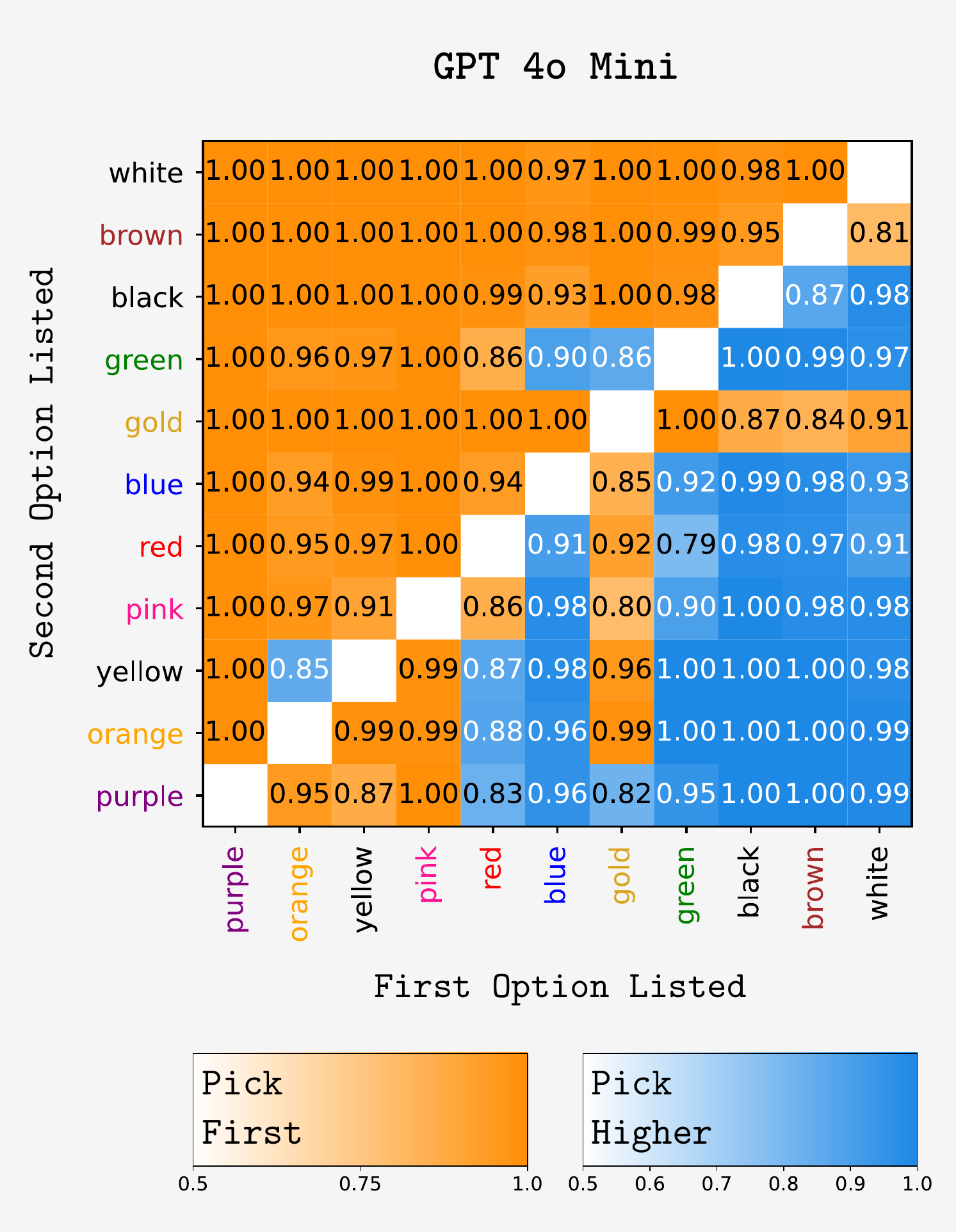}
 \caption{
 \textbf{Systematic Patterns in Model Behavior.} Each cell corresponds to model behavior across 100 examples. The number is the rate the outputs are compatible with the given behavior. For example, 1.00 in the first top-left cell means that for 100/100 instances a majority of the probability mass is on the first option. The top-left cell corresponds to instances where \textcolor{PURPLE}{purple} is first in the prompt and \textcolor{gray}{white} is second. High rates across multiple behaviors are impossible; they are mutually exclusive across the 100 instances. See \Cref{fig:examples-fig} for a representative instance drawn from the top-left cell.
 }
\label{fig:hard-fig}
\end{figure}
In this section, we study the particular behaviors that models exhibit on the \textit{colors} dataset, finding that \textbf{the models exhibit systematic biases and preferences based upon word identity and word order.}
Understanding that (1) the models are poorly calibrated and (2) exhibit low entropy falls directly out of our proposed metrics. However, understanding how the models behave is more complicated. To start, we characterize the behavior of \gpt{4o-mini} in \Cref{fig:hard-fig} (more in \Cref{app:results:ordering}). \Cref{fig:model_behavior_bar_plot} shows model behavior across all settings.

\begin{figure*}[ht!]
 \centering
 \ttfamily
\begin{tikzpicture}
    \begin{axis}[
ybar stacked,
ymajorgrids,
tick align=inside,
major grid style={draw=white},
symbolic x coords={0,1,2,3,4,5,6,7,8,9,10,11},
xtick=data,
ymin=0, 
ymax=100,
height=2.25in,
width=5.5in,
ylabel=Percent (\%),
enlarge x limits={abs=5},
bar width=8mm,
    enlargelimits=0.05,
      legend style={at={(1,0)},anchor=south east,},
xticklabel style={align=center,font=\scriptsize}, 
xticklabels={%
{Mistral \\7Bv0.1}, {Mistral \\7Bv0.3}, {Mixtral\\8x7B}, 
{Yi-1.5\\9B}, {Yi-1.5\\34B}, {Llama \\3.1 8B}, 
{gemma\\2 9B}, {gemma\\2 27B}, {gpt-3.5}, {gpt-4\\turbo}, 
{gpt-4o\\mini}, {gpt-4o},
},
ylabel style={font=\small}, 
yticklabel style={font=\scriptsize}, 
legend style={font=\scriptsize, at={(1.01,1)}, anchor=north west}, 
    legend cell align={left}, 
    reverse legend,
legend image code/.code={
        \draw [#1, draw=none] (0cm,-0.1cm) rectangle  (0.2cm,0.365cm); }
    ]

\addplot+[ybar, draw=none, fill=C_NULL] plot coordinates
{(0,20.0) (1,42.0) (2,16.0) (3,1.0) (4,0.0) (5,33.0) (6,0.0) (7,0.0) (8,0.0) (9,0.0) (10,0.0) (11,0.0)};
\addplot+[ybar, draw=none, fill=C_CALIBRATED] plot coordinates 
{(0,33.0) (1,18.0) (2,18.0) (3,1.0) (4,4.0) (5,10.0) (6,7.0) (7,0.0) (8,8.0) (9,0.0) (10,0.0) (11,0.0)};
\addplot+[ybar, draw=none, fill=C_HIGHER] plot coordinates  
{(0,11.0) (1,0.0) (2,10.0) (3,40.0) (4,43.0) (5,16.0) (6,6.0) (7,87.0) (8,75.0) (9,84.0) (10,70.0) (11,84.0)};
\addplot+[ybar, draw=none, fill=C_FIRST] plot coordinates 
{(0,18.0) (1,40.0) (2,36.0) (3,43.0) (4,41.0) (5,32.0) (6,86.0) (7,13.0) (8,15.0) (9,0.0) (10,27.0) (11,16.0)};
\addplot+[ybar, draw=none, fill=C_SECOND] plot coordinates
{(0,18.0) (1,0.0) (2,19.0) (3,15.0) (4,13.0) (5,10.0) (6,1.0) (7,1.0) (8,1.0) (9,16.0) (10,3.0) (11,0.0)};    \legend{null, calibrated, higher,  first,  second}
    \end{axis}
    \end{tikzpicture}
 \caption{\textbf{Systematic Patterns in Model Behavior} for the \textit{colors} dataset. Each bar shows the percent of the different behaviors models exhibit averaged across templates, color pairs, and numeric scales.
}
 \label{fig:model_behavior_bar_plot}
\end{figure*}

Beyond the specific systematic patterns we uncover, we want to stress that the existence of these patterns is important. Every use case of LMs that requires such choices may have similar patterns. \textbf{That these systematic errors arise even with innocuous features like color should give us pause about using models to make probabilistic choices in more complicated scenarios.}

\Cref{fig:hard-fig} captures model behavior across all tested pairs of color options. The most apparent note to make is the diagonal asymmetry. \textbf{This asymmetry shows a hierarchical, systematic bias across word identity (specific colors) and word order.} Each cell within the heatmap reports the rate at which the model's output probabilities are compatible with the given behavior across a full grid of number pairs. The behavior with the highest compatibility rate is displayed and indicated by the color. For example, if the model had rates of 100 for Pick first, 50 for Pick higher, and 50 for pick second, the plot would show the rate of 1.00 (100/100) for Pick first. A high rate of two behaviors cannot happen because the 100 examples are balanced (in half, the first number is higher, in the other half, the second number is higher.) 

\begin{figure}[ht!]
 \centering
 \includegraphics[width=\linewidth]{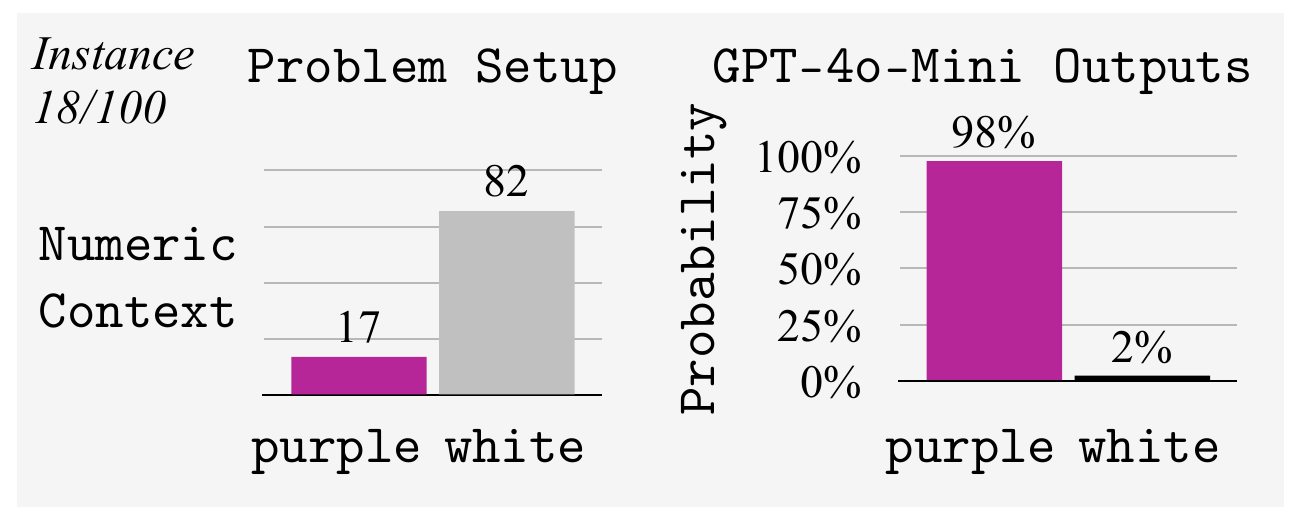}
 \caption{\textbf{Representative Examples of Model Behaviors.}
 This instance is from \gpt{4o-mini} results in \Cref{fig:hard-fig}, \texttt{Pick First} behavior from the top-left cell.}
 \label{fig:examples-fig}
\end{figure}

The behaviors exhibited across the diagonal are often different.
This means that option order, listing a color first or second, greatly affects calibration. For example, in \Cref{fig:hard-fig}, we can see that when ``white'' is mentioned first, the model tends to Pick Higher and when it is mentioned second, the model tends to Pick First. Because Pick Higher is a more calibrated behavior than Pick First, mentioning ``white'' first leads to lower calibration scores.

We measure these differences in calibration across options and models, see detaisl in \Cref{app:results:ordering}. Color orderings are correlated among similar models (\gpt{4o}/\gpt{4o-mini}), suggesting that training data may play a large role in forming these patterns of behavior.

\Cref{fig:model_behavior_bar_plot}, reports the overall behavior rates for each model. \Cref{app:results:ordering} details how we compute the proportions of different behaviors across the dataset.
\gpt{*} models (and \gemma{27b}) tend to \texttt{Pick Higher}, others also \texttt{Pick First} and \texttt{Second}. Smaller models also exhibit \texttt{null}. These patterns (1) differ across models (whether it is Pick First or Second vs Pick Higher or something else) but (2) the word identity (\textcolor{red}{\textit{red}}/\textcolor{gray}{\textit{white}} with the same numeric information) consistently interacts with word order (\textcolor{red}{\textit{red}}/\textcolor{gray}{\textit{white}} vs \textcolor{gray}{\textit{white}}/\textcolor{red}{\textit{red}}) to create systematic error patterns per model.

\begin{table*}[ht!]
\small
    \centering
\begin{tabular}{@{}l@{\hspace{4pt}}r@{\hspace{4pt}}r@{\hspace{4pt}}r@{\hspace{4pt}}|r@{\hspace{4pt}}r@{\hspace{4pt}}r@{}}
\toprule
 & \multicolumn{3}{c}{Base} & \multicolumn{3}{|c}{Chat} \\
  &            \textit{colors} & \textit{wordp.} & \textit{distr.} &       \textit{colors} & \textit{wordp.} & \textit{distr.} \\
\midrule
Llama-3.1-8B &       \RC{\small  +0.68} &  \RC{\small  +0.46} &   \GC{\small  -0.26} &  \RC{\small  +0.57} &  \WC{\small  +0.03} &   \WC{\small  -0.09} \\
Mistral-7B-v0.1   &       \RC{\small  +0.58} &  \RC{\small  +0.60} &   \GC{\small  -0.40} &  \RC{\small  +0.17} &  \WC{\small  +0.08} &   \GC{\small  -0.25} \\
Mistral-7B-v0.3   &       \RC{\small  +0.57} &  \RC{\small  +0.59} &   \GC{\small  -0.44} &  \RC{\small  +0.21} &  \WC{\small  +0.00} &   \WC{\small  +0.02} \\
Mixtral-8x7B &       \RC{\small  +0.71} &  \RC{\small  +0.52} &   \GC{\small  -0.34} &  \GC{\small  -0.44} &  \WC{\small  -0.04} &   \WC{\small  +0.00} \\
Yi-1.5-34B        &       \RC{\small  +0.66} &  \RC{\small  +0.49} &   \GC{\small  -0.35} &  \WC{\small  -0.01} &  \WC{\small  +0.02} &   \WC{\small  -0.00} \\
Yi-1.5-9B         &       \RC{\small  +0.57} &  \RC{\small  +0.49} &   \GC{\small  -0.48} &  \GC{\small  -0.24} &  \WC{\small  -0.06} &   \GC{\small  -0.20} \\
gemma-2-27b       &       \RC{\small  +0.61} &  \RC{\small  +0.52} &   \GC{\small  -0.44} &  \GC{\small  -0.55} &  \GC{\small  -0.57} &   \WC{\small  -0.12} \\
gemma-2-9b        &       \RC{\small  +0.55} &  \RC{\small  +0.65} &   \GC{\small  -0.47} &  \WC{\small  -0.08} &  \WC{\small  -0.02} &   \WC{\small  -0.16} \\
gpt-3.5           &                 - &            - &             - &  \GC{\small  -0.40} &  \GC{\small  -0.38} &   \WC{\small  -0.12} \\
gpt-4-turbo       &                 - &            - &             - &  \GC{\small  -0.54} &  \GC{\small  -0.47} &   \WC{\small  -0.01} \\
gpt-4o            &                 - &            - &             - &  \GC{\small  -0.46} &  \GC{\small  -0.34} &   \GC{\small  -0.24} \\
gpt-4o-mini       &                 - &            - &             - &  \GC{\small  -0.61} &  \WC{\small  -0.07} &   \GC{\small  -0.31} \\
\bottomrule
    \end{tabular}
    \caption{\textbf{Frequency Effects.} Each cell reports the Spearman’s correlation between the \textit{Frequency Gap} between the numbers in context and the calibration scores. Positive correlations mean that larger  word frequency gaps predict worse calibration, \textcolor{PINK}{highlighted orange (worse)}. Negative correlations mean that larger frequency gaps predict improved calibration, \textcolor{BLUE}{\strut highlighted blue (better)}. Cells not highlighted are not statistically significant at $p < 0.01$.  } 
    \label{tab:freq}
\end{table*}

\section{Analysis: Number Token Frequency}\label{sec:analysis:freq}
\paragraph{Numbers appear at different frequencies in training data. Does frequency explain the differences in calibration?} We find that the models exhibit different behaviors, but there are three primary patterns. Across almost all models, there is a high and significant correlation between number frequencies and average calibration scores.
\begin{exe}
 \ex {From \textbf{50} {\textit{blue}} marbles and \textbf{84} \textit{red}$\dots$}
\end{exe}
\noindent Repeating a partial example similar to those above, here, in this section, we focus our analysis on the numeric content, here $N_1=50 $/$ N_2=84$, particularly the frequencies of those numbers in the training data. Because the models we test do not publish their training data details, we approximate the frequency of numbers in their training data using infini-gram\footnote{\url{https://infini-gram.io/api_doc.html}} \citep{liu2024infini} over Dolma \citep{soldaini2024dolma}, a 3 trillion token corpus. Given the limited amount of natural text available online, we expect pre-training datasets to overlap significantly and be similarly distributed.\footnote{ We query for the number counts with spaces both before and after each number: ``\_$N$\_'', to account for numbers appearing within other numbers. This assumes the noise we introduce here is distributed similarly across numbers.} 

We use the Spearman's correlation between the calibration scores and the \textit{Frequency Gap}, the log absolute difference of the number frequencies, $\log|(\textrm{Freq}(N_1) - {\textrm{Freq}(N_2)}|$.\footnote{See \Cref{app:results:freq} for more results and correlations.} Using the Frequency Gap is a simple way to capture if there is a relationship between the frequencies of the two numbers.

We present the correlation results in \Cref{tab:freq}.
Frequency effects appear to negatively impact the base models: Larger Frequency Gaps correlate with worse calibration. Conversely, frequency effects appear to positively impact instruction-tuned models: Larger Frequency Gaps correlate with improved calibration. Frequency effects on \textit{distributions} are different. Effects are weaker, and increased frequency tends to improve calibration.

Frequency effects are a known issue for LMs \citep{wei2021frequency,lovering2023training,mccoy2024embers}; \textbf{Our results reinforce the understanding that increases in model scale and performance have not canceled out such biases.}

\section{Conclusion}\label{sec:conclusion}
All models we test are not only \textit{not} calibrated; they are generally less calibrated than simple baselines.
Word identity, word order, and the vocabulary frequency of the numeric information heavily modulate model behavior. Even in the basic problem settings we test in this paper, LMs are heavily biased, which suggests there are significant risks in using models in probabilistic scenarios to make decisions.

\section*{Limitations}
This work strikes a balance between sampling across a large space of options--more numbers \citep{shrestha2025mathematicalreasoninglargelanguage}, more unique templates--and uncovering biases present in those chosen options, which benefits from a ``grid-search'' approach. Future work that increases the diversity and size of these datasets could only improve this line of work.

We report Spearman's correlation in \Cref{sec:analysis:freq}. Using mixed-effects models may provide additional nuance. Increasing the range of numbers used, especially in this section, could improve the robustness of our results.

\section*{Potential Risks}
This work does not pose a risk. Instead, this work highlights several risks of using uncalibrated models and shows that current models are uncalibrated even in simple settings.

\bibliography{custom}
\appendix
\clearpage
\section*{\textsc{Basics}}

\section{Housekeeping}\label{app:details}
\begin{itemize}
    \item First off, we apologize to any color-blind readers; we make heavy use of color in this paper. While we do our best to use color-blind aware color palettes, this paper doubtlessly is relatively unfriendly from that point of view. (We found David Nichols's resources helpful: \url{https://davidmathlogic.com/colorblind/#%23D81B60-%231E88E5-%23FFC107-%23004D40}.)
    \item We used ChatGPT to assist with \LaTeX~ debugging. (Editorial comment: it was net helpful if not always right.)
    \item {Because we have a large number of figures in this appendix, we front-load the (almost) text-only sections first and then follow this with a single column, mirrored section structure, with only figures. }

\end{itemize}
\section{Reproducibility}\label{app:results:reproducibility}
All data and experiments will be released under The MIT License. All details required for reproducing our results, like problem templates and prompts, will also be released at the link. Data was created with research settings in mind, but can be used in any setting (research/professional).    \url{https://drive.google.com/drive/folders/11mY5vbNanqMA2wwci84I5cipwh7FPPPB} 

\Cref{tab:prompts:dist,tab:prompts:prob,tab:templates:dist,tab:templates:prob} report the prompts (formatted as messages) and problem templates.

We use the models listed in \Cref{tab:models}. For visual acuity, we only put the full model codes in this table, rather than throughout the paper.
All experiments can be run on a P4de-24 (8 NVIDIA A100) in approximately 48 hours.

\section*{\textsc{More Results}}
\section{{More Results on What Do Metric Values Mean?}}\label{app:results:demo}

In this section, we show reference performance scores, \Cref{app:tab:more:best:b}. Second, we show that most models are capable of doing well and sometimes (albeit rarely) do in fact produce well-calibrated outputs. The results here are over a small, new set of numbers for demonstration purposes: 8, 9, 48, 51, 98, 99. Figures are in \Cref{app:results:demo:figures}.

The scores in \Cref{app:tab:more:best:b} show WD on the high end will be around 1.00. What these results don't show well is what is a ``good'' calibration score. We find that in a small number of cases the model achieves calibration scores near 0, less than 0.05. We consider these examples solved.

\section{{More Results on Is It Possible For The Models To Do Well?}}\label{app:results:solved}

\begin{figure*}[ht!]
 \centering \frame{\includegraphics[width=0.95\linewidth]{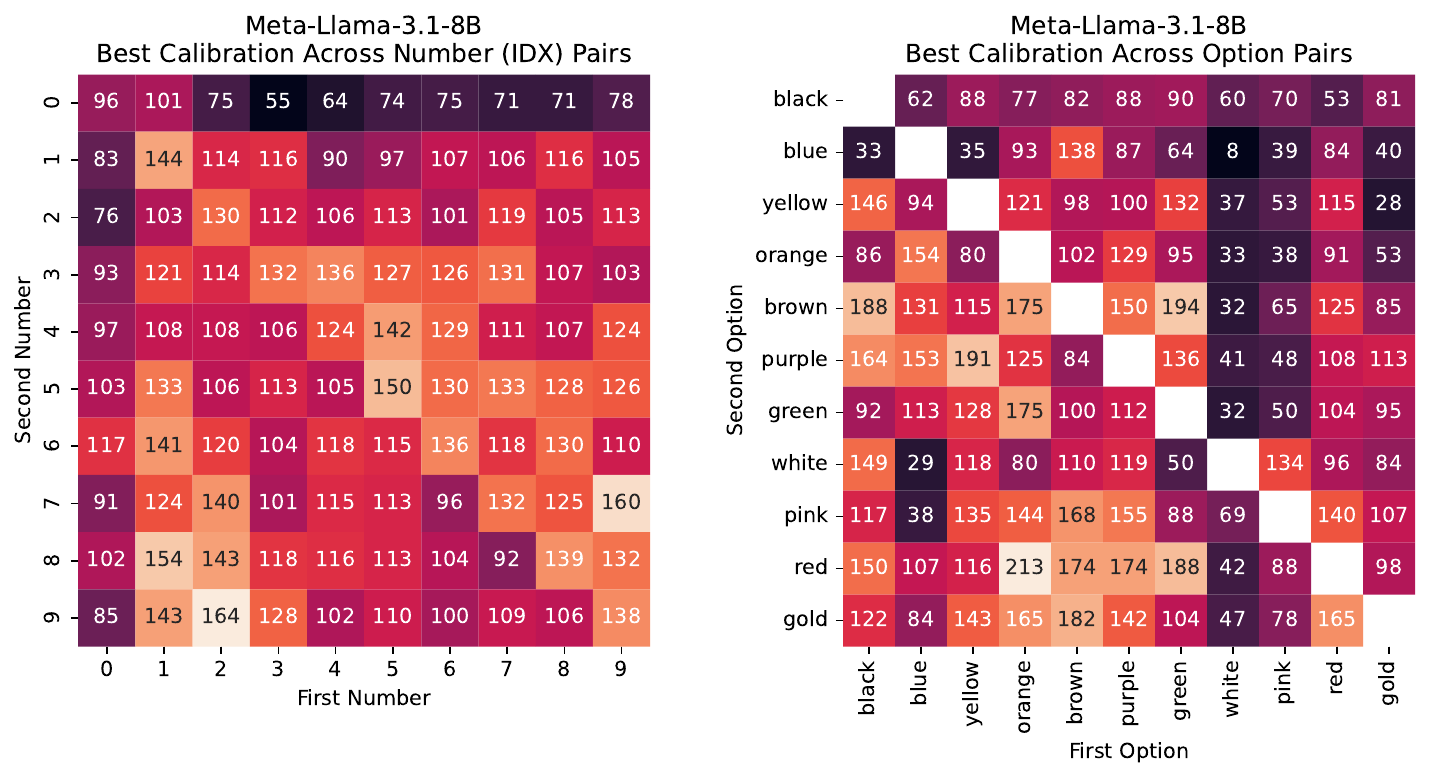}}
\frame{\includegraphics[width=0.95\linewidth]{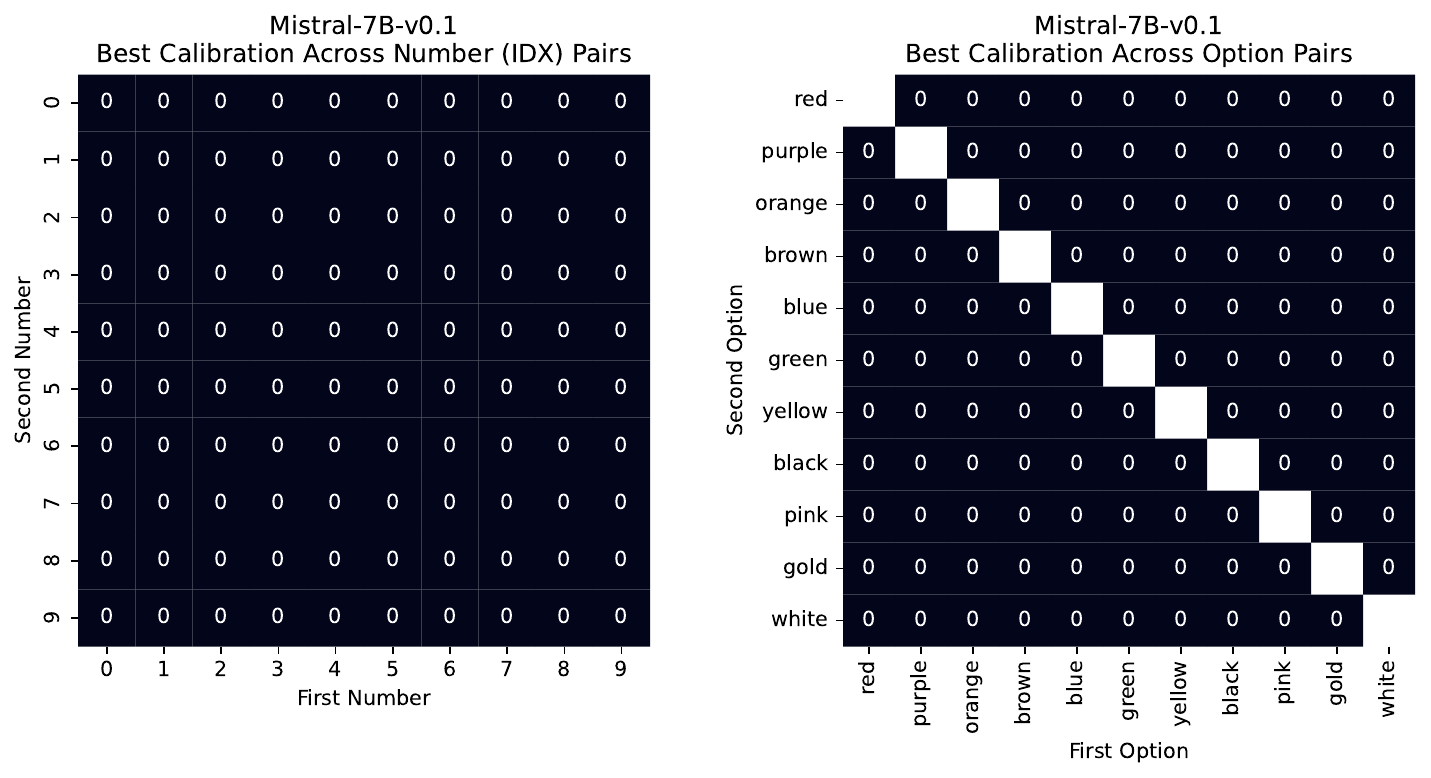}}
\frame{\includegraphics[width=0.95\linewidth]{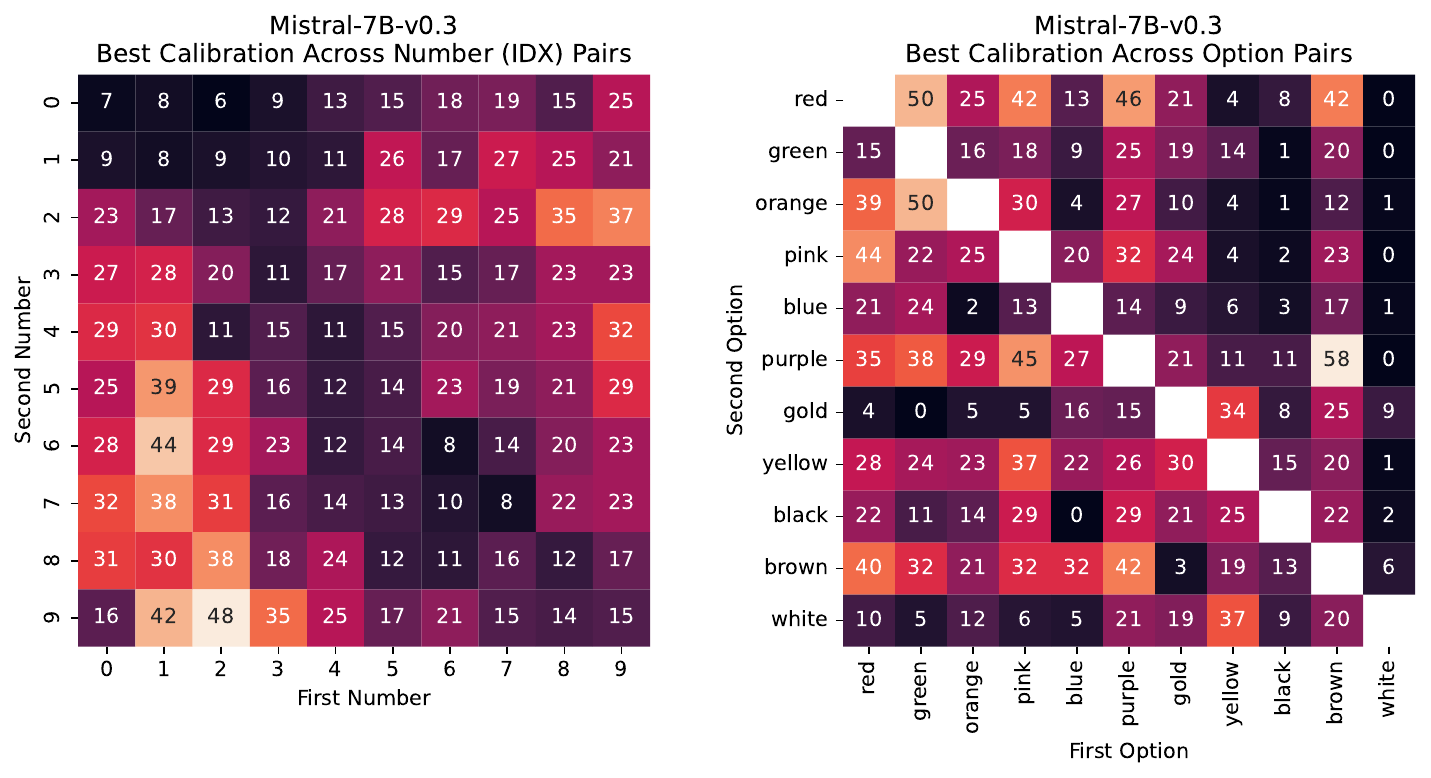}}
 \caption{\textbf{Best Results Over Different Settings for \llama{}, \mistral{1},  \mistral{3}.} 
 }
 \label{app:fig:solved:1}
 \end{figure*}

\begin{figure*}[ht!]
 \centering \frame{\includegraphics[width=0.95\linewidth]{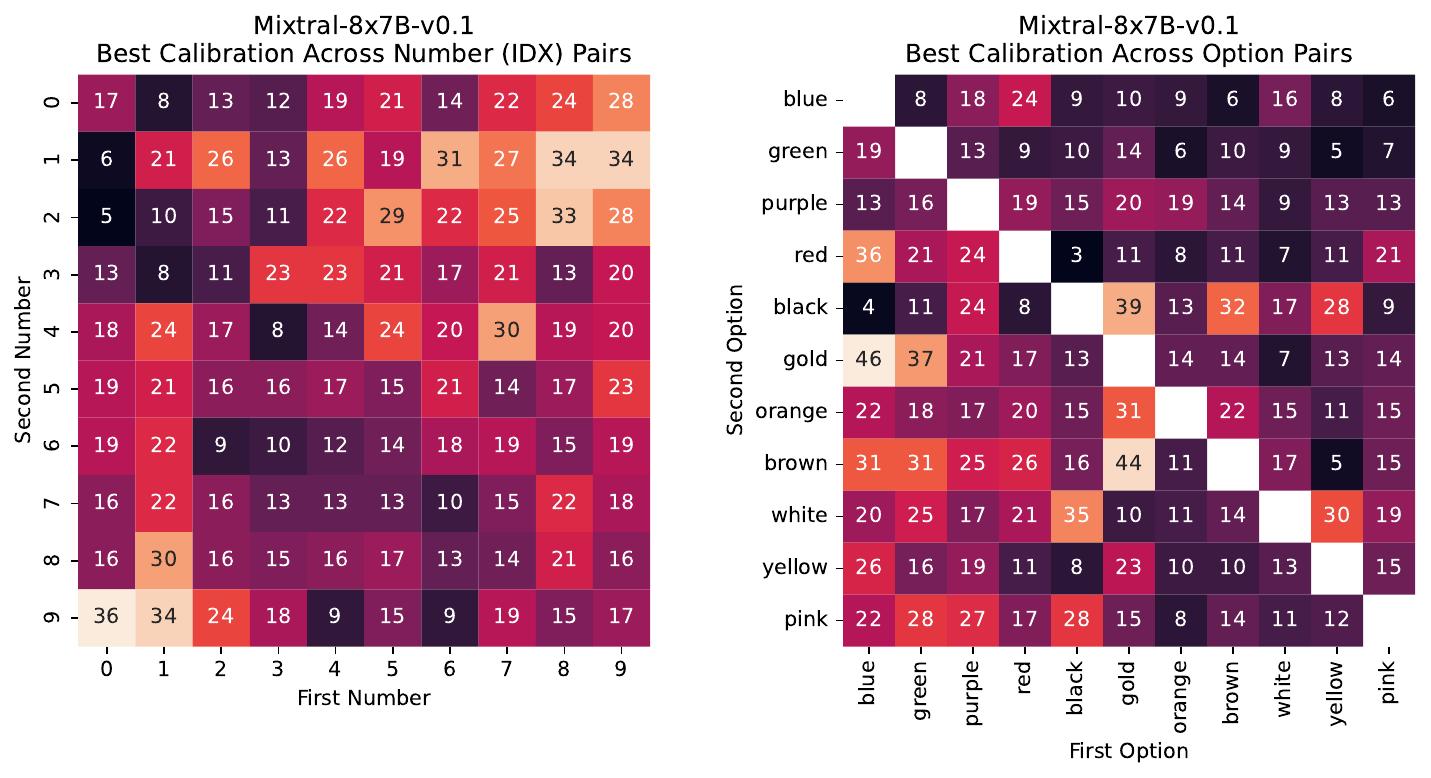}}
\frame{\includegraphics[width=0.95\linewidth]{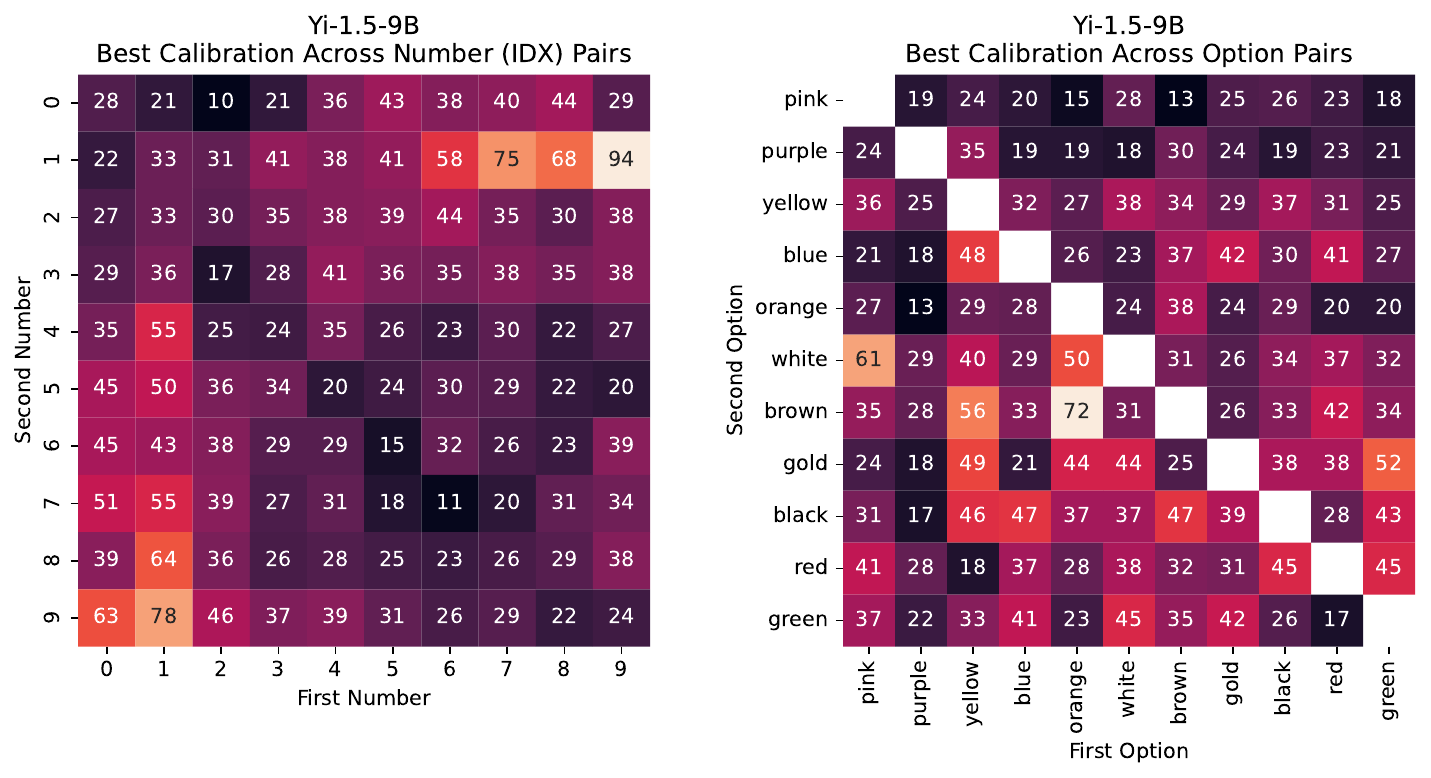}}
\frame{\includegraphics[width=0.95\linewidth]{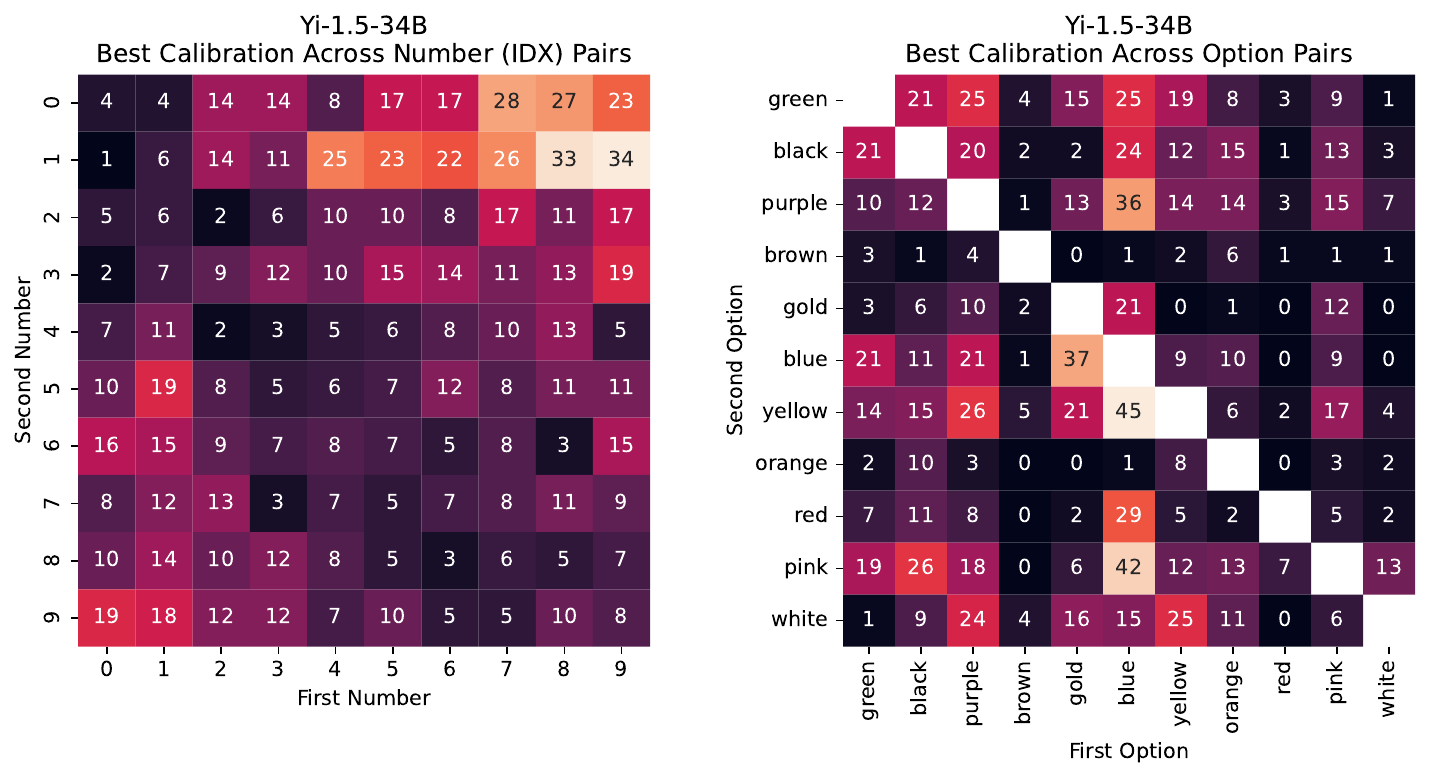}}
 \caption{\textbf{Best Results Over Different Settings for \mixtral{}, \yi{9},  \yi{34}.} 
 }
  \label{app:fig:solved:2}
 \end{figure*}

\begin{figure*}[ht!]
 \centering \frame{\includegraphics[width=0.95\linewidth]{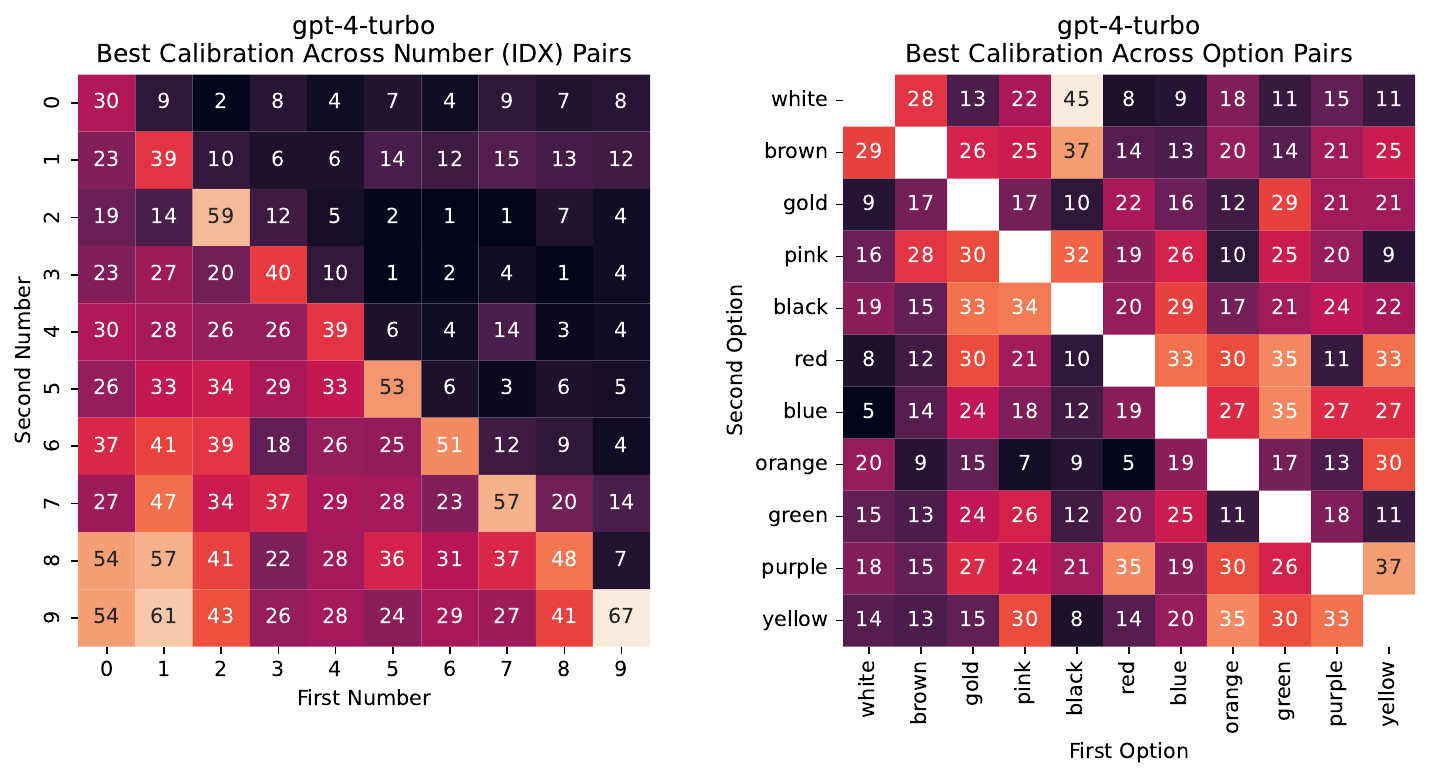}}
\frame{\includegraphics[width=0.95\linewidth]{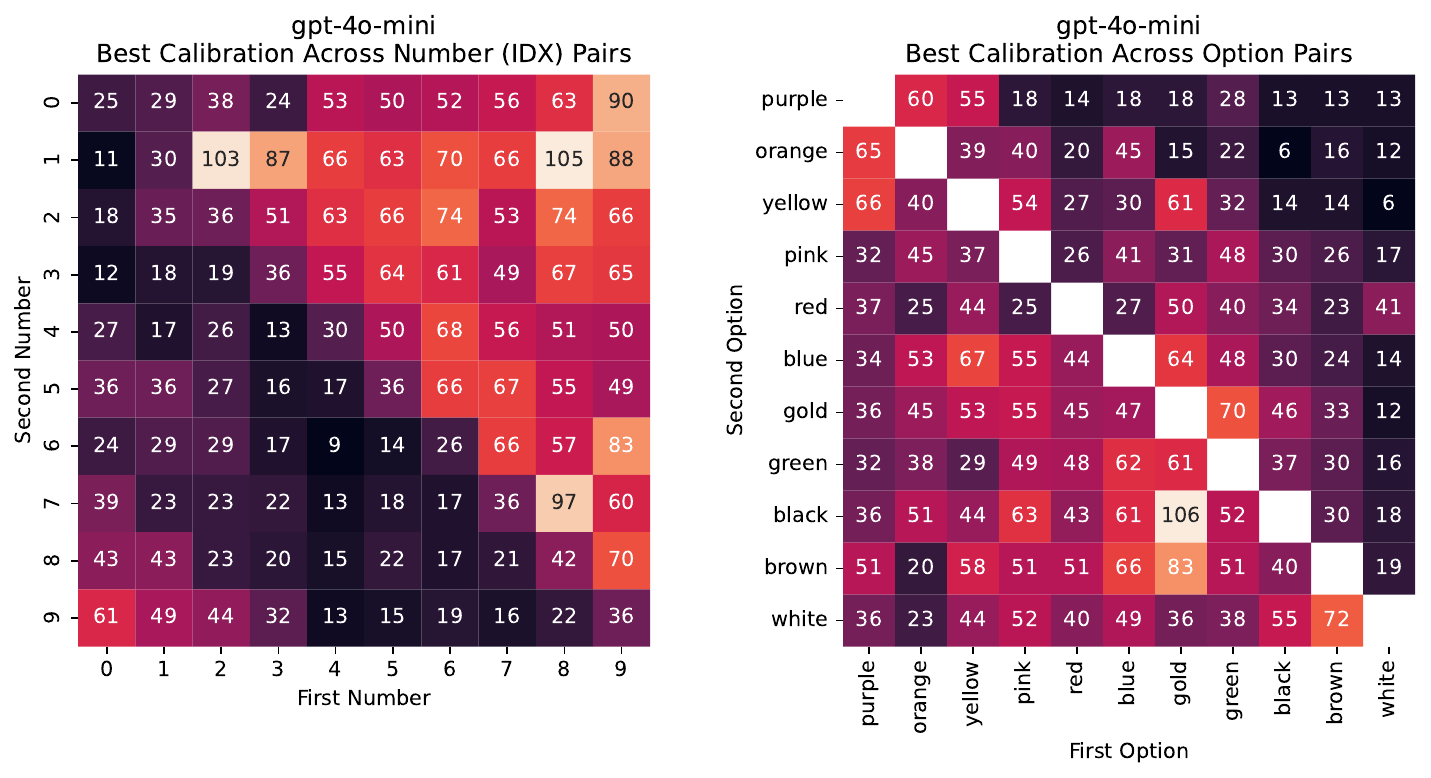}}
\frame{\includegraphics[width=0.95\linewidth]{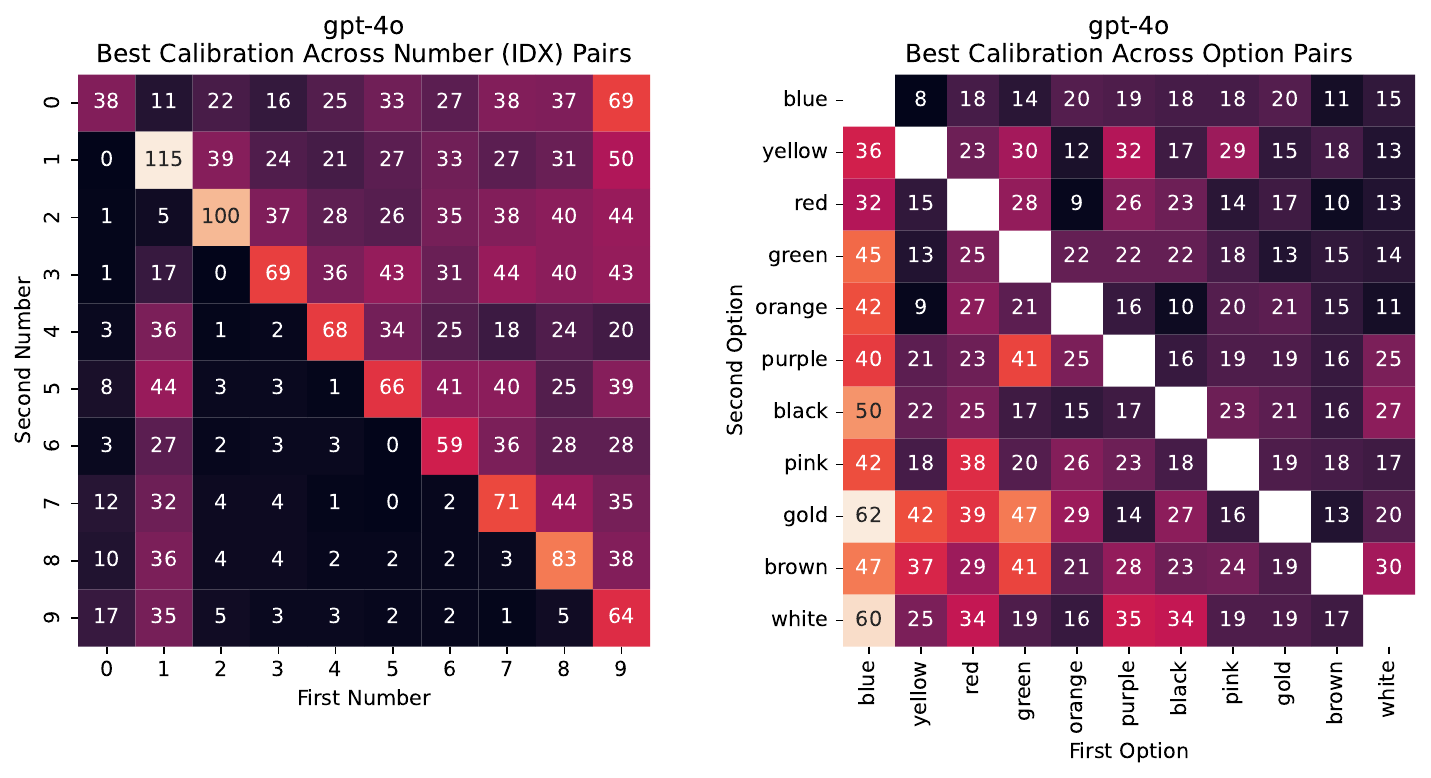}}
 \caption{\textbf{Best Results Over Different Settings for \gpt{4-turbo}, \gpt{4o-mini},  \gpt{4o}.} 
 }
 \label{app:fig:solved:3}
 \end{figure*}

\Cref{app:fig:solved:1,app:fig:solved:2,app:fig:solved:3} shows the count of solved inputs across either pairs of numbers (left) or pairs of colors (right). Each number is a raw count out of 500, shown as such for visual acuity. For these plots we considered the model to have solved an instance if the $\textrm{WD} < 0.05$. This tends to constitute a good fit. Making this threshold more strict, say, $0.01$, maintains the trends present in the figure. Mostly, these plots show that (1) in some cases the models do calibrate well even if in general they tend not to; (2) the number pair and option pair have a large impact on whether a model ever yields a solved result, mirroring previous results; (3) the patterns differ per model but again mirror previous results. The most important result is the first. \textbf{Models are clearly capable of yielding calibrated probabilties, but do not do so in the (large) majority of cases.} 

\section{{More Results on Calibration And Low Probability Mass}}\label{app:results:calibration}
In this section the first question we ask is  if our calibration results are stable across a greater diversity of data. We do this by 1) introducing variations of each template, and 2) creating more templates.
Specifically, using \texttt{\gpt{4o}-2024-08-06}, we create 500 leading introductions to prefix our templates in the \textit{colors} dataset, ranging from one to ten sentences. Next, we create 500 more templates for the \textit{colors} dataset using \texttt{\gpt{4o}-2024-08-06}. For both new splits we test a subset of our instruction-tuned models. The calibration scores did changed little from  the original performances, \Cref{tab:results:intros}. In two cases, the calibration was worse (the confidence interval was strictly positive), though the effect was not large. Using synthetically generated data has its risks, nonetheless, we take these results as evidence that calibration over a wider distribution of similar data remains poor.
\begin{table}[ht!]
\small
    \centering
    \begin{tabular}{@{}l@{\hspace{4pt}}l@{\hspace{4pt}}l@{\hspace{4pt}}l@{}}
    \toprule
    {} &   &  \textit{Intros} - WD  & \textit{Templates}  - WD \\
    {} &  WD & $\Delta$ (95\% CI)  & $\Delta$ (95\% CI) \\
    \midrule
Llama-3.1-8B &  0.40 &      \WC{\small   (-0.04, +0.00) }&           \RC{\small  (+0.02, +0.05) }\\
Yi-1.5-9B         &  0.49 &       \WC{\small  (-0.03, +0.02) }&           \RC{\small  (+0.09,+ 0.13) } \\
gemma-2-9b        &  0.50 &      \WC{\small   (-0.02, +0.04) }&          \WC{\small  (-0.00, +0.04) }\\
    \bottomrule
    \end{tabular}
    \caption{Calibration Scores for Additional Experiments. The first column WD is the original calibration score (repeated from \Cref{tab:results:wd}). The remaining columns show the 95\% confidence intervals of the change in performance between the original results and the alternate settings, \textit{intros} and \textit{templates}. 
    Statistically significant increases in calibration scores are  \textcolor{PINK}{highlighted orange (worse calibration)}.
    }
    \label{tab:results:intros}
\end{table}

\section{{More Results on Mode Collapse}}\label{app:results:mode}
\Cref{app:tab:results:re} shows the drop in RE from base models to chat models. This occurs across all datasets.

We also provide a more in-depth analysis on mode collapse in the \textit{distributions} dataset. 

\paragraph{Additional Metrics for \textit{distributions}.} We introduce three metrics to understand if there are patterns in the mode.

\textbf{Mode Probability} is the mean max probability of $\Pi$ over the dataset, $\textrm{AVG}_{\textrm{dataset}} \max \Pi$.

\textbf{Mode Stability} is the rate at which the most likely token is preserved between base and chat versions of a model averaged over the dataset,
$\textrm{AVG}_{\textrm{dataset}}(\mathds{1}(\arg\max \Pi_{\textrm{base}} = \arg\max \Pi_{\textrm{chat}}))$.

\textbf{Mode Frequency} examines whether there are biases for particular tokens (numbers). This metric measures the frequency of the mode averaged across distributions. Because different distributions cover different digits, we group the distributions by underlying problem range and inclusivity: E.g., [2, 5) and [132, 135) are grouped. This is equivalent to: (1) setting the temperature to 0 (greedy sampling), (2) averaging over the groups, keeping the digits dimension, (3) taking the maximum for each group, and (4) averaging across all maximums,
$\textrm{AVG}(\max_\textrm{digits}\textrm{AVG}_{\textrm{distributions}, \textrm{ keepdim}}(\Pi^{\tau = 0}))$.

\begin{figure*}
 \centering
 \includegraphics[width=\linewidth]{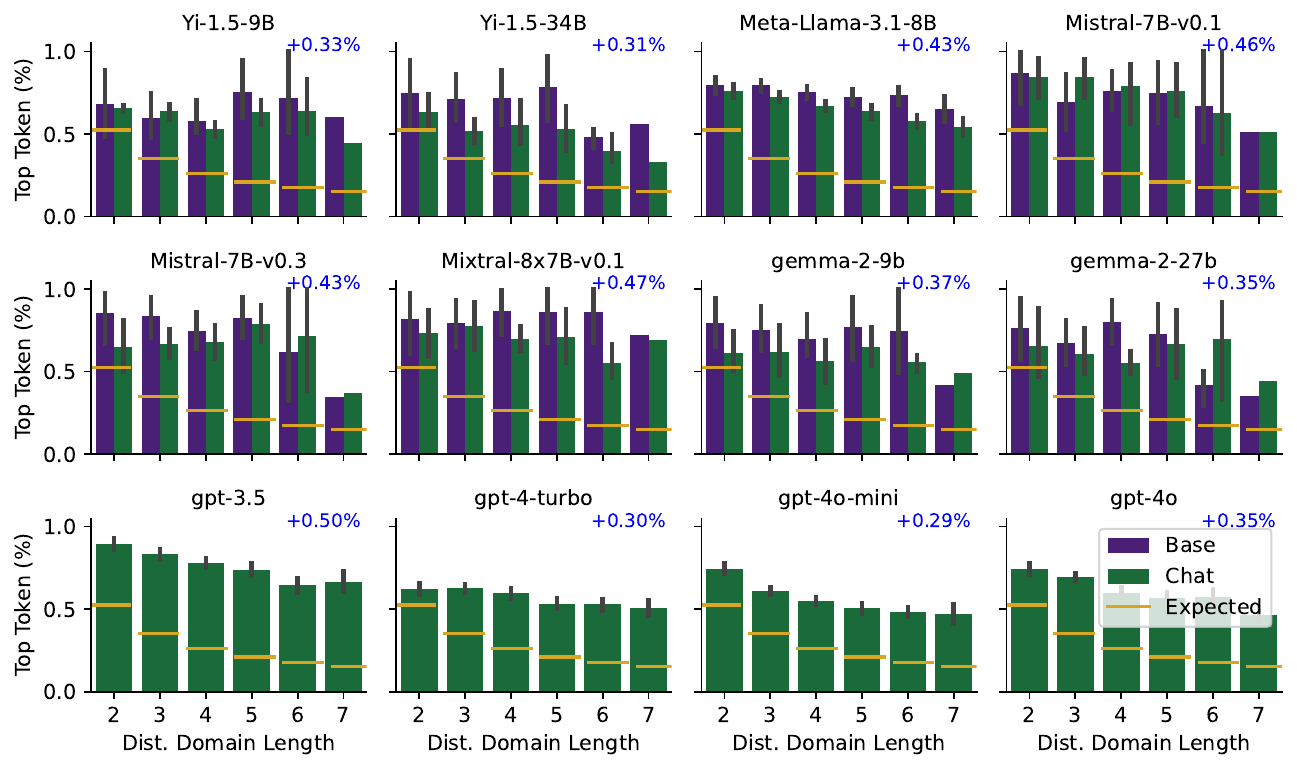}
 \caption{\textbf{Models Over-represent Numbers.} Each bar shows how often the top-chosen token is chosen (percent). The \textcolor{goldenrod}{gold} bars mark the expected rate for a calibrated model; the \textcolor{blue}{blue} annotation marks the average excess probability on the top-chosen token, ranging from $29-50\%$ across models. Takeaways \textbf{(1)}: Models over-represent a token (number) over other valid options. This token is not always the same but the pattern of over-representing a number irrespective of the numeric context holds.}
 \label{fig:top-tokens-full}
\end{figure*}

\paragraph{Results} We find evidence that the reduced performance of chat models results from their tendency to over-allocate probability on a subset of valid tokens. \Cref{app:tab:dist:mode} provides additional metrics that suggest mode-collapse. For chat models, the most likely token receives, on average, $66\%$ of the probability mass (up from $28\%$ for base models, beyond the empirical ideal $23\%$, on this dataset). The most likely token remains the same for $48\%$ instances across base and chat models, suggesting that the mode is often preserved.
\begin{figure}[ht!]
 \centering

\ttfamily
\begin{tikzpicture}
    \begin{groupplot}[
group style={columns=1},
scale only axis,
view={0}{90},
width=4.9cm,
height=2.5cm,
ymin=-0.01, ymax=1.00,
group style={
        group size=1 by 1,
        horizontal sep=10pt,
        vertical sep=25pt,
        x descriptions at=edge bottom,
        y descriptions at=edge left},
]
  \nextgroupplot[
  ybar,
    ylabel style={at={(-0.15,0.5)}, font=\footnotesize},
          ylabel={Mode Frequency (\%)},
      title=gpt-4o-mini,
      bar width=6mm,
  ymajorgrids,
xlabel style={font=\footnotesize,align=center},
xlabel={Distribution Domain Length},
xtick={2,3,4,5,6,7},
  legend image code/.code={
        \draw [#1, draw=none] (0cm,-0.1cm) rectangle  (0.2cm,0.365cm); },
legend style={font=\scriptsize, at={(0.99,0.98)}, anchor=north east}, 
legend columns=1
  ]
  \addplot[draw=none, style={C_FIRST,fill=C_FIRST,mark=none}] coordinates {(2,0.75) (3,0.61) (4,0.55) (5,0.51) (6,0.48) (7,0.47)};
\draw [-, very thick, black] (axis cs:1.65,1/2) -- (axis cs:2.35,1/2);
\draw [-, very thick, black] (axis cs:2.65,1/3) -- (axis cs:3.35,1/3);
\draw [-, very thick, black] (axis cs:3.65,1/4) -- (axis cs:4.35,1/4);
\draw [-, very thick, black] (axis cs:4.65,1/5) -- (axis cs:5.35,1/5);
\draw [-, very thick, black] (axis cs:5.65,1/6) -- (axis cs:6.35,1/6);
\draw [-, very thick, black] (axis cs:6.65,1/7) -- (axis cs:7.35,1/7);
\legend{Base}

\end{groupplot}
\end{tikzpicture}
 \caption{\textbf{Models Over-represent Some Numbers}; the modes are heavily over-represented. Each bar is the mode frequency, i.e., how often the top-chosen token is chosen averaged over distributions. The \textcolor{black}{black} lines mark the expected rate for a calibrated model.}
 \label{fig:top-tokens}
\end{figure}
\Cref{fig:top-tokens} reports the mode frequency for \gpt{4o-mini}. (For all models see \Cref{fig:top-tokens-full}.) If all probability mass were on valid tokens, the minimum mode frequency would be the same as the rate of the uniform distribution; we see much higher rates, ranging from $+29$ to $+50\%$. The implication is that models are also uncalibrated at an outcome level with systematic biases (preferences) for certain numbers. 

\section{{More Results on Ordering and Colors}}\label{app:results:ordering}

\begin{figure*}[ht!]
 \centering
 \includegraphics[width=0.55\linewidth]{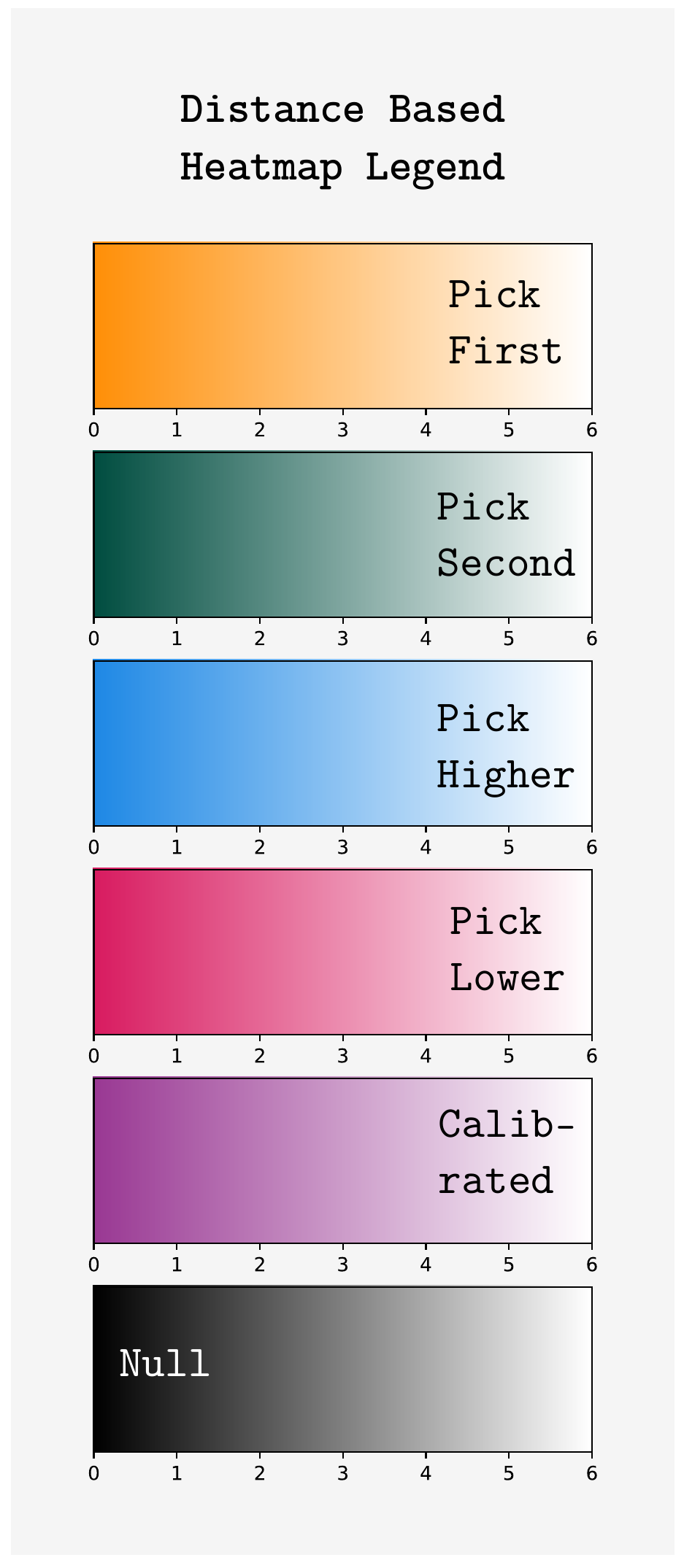}
 \caption{\textbf{Legend for heatmaps on the following page.} 
 }
 \label{app:fig:legend:distance}
\end{figure*}

\begin{figure*}[ht!]
 \centering \includegraphics[width=0.85\linewidth]{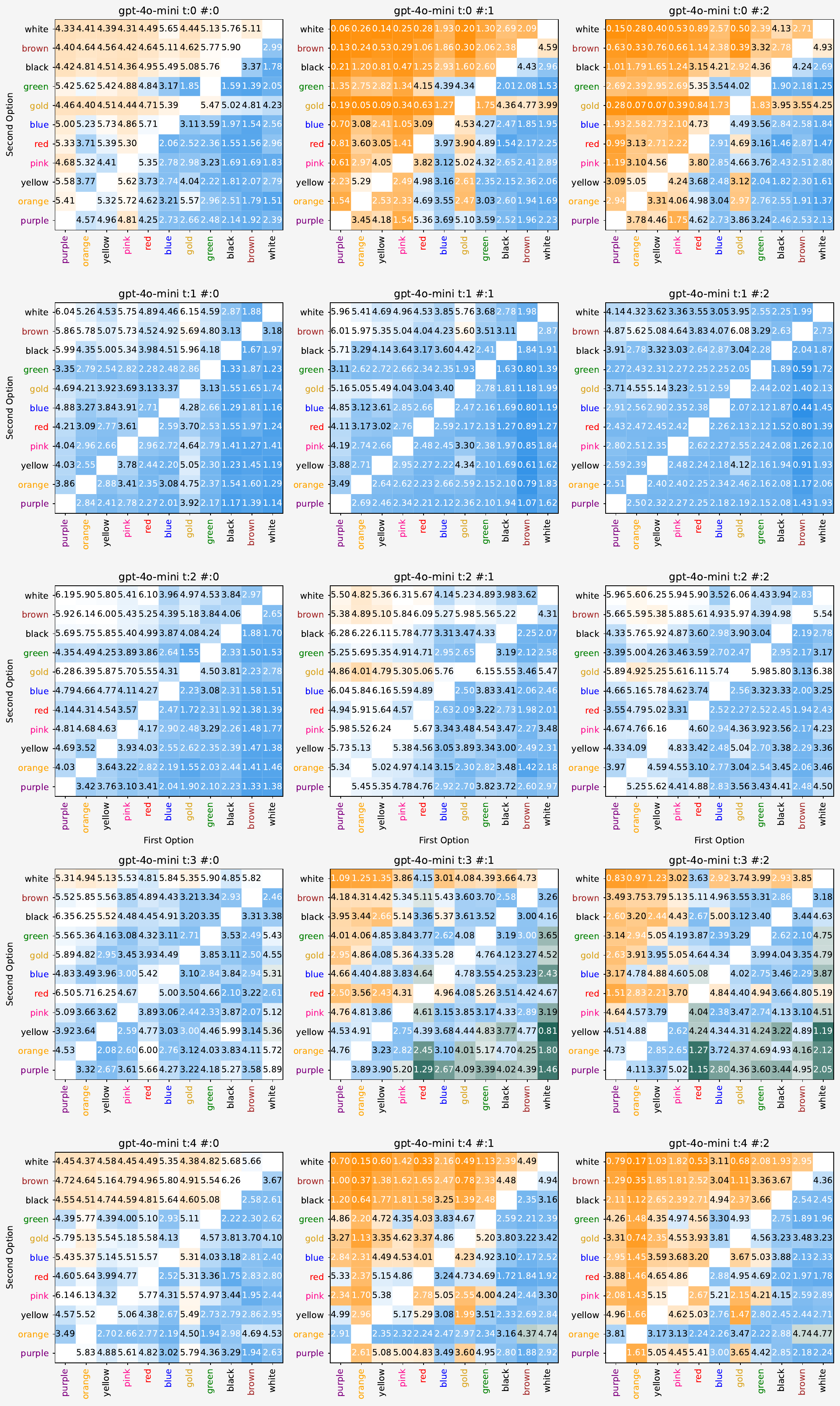}
 \caption{\textbf{Distance-based Heatmap for \gpt{4o-mini}.} 
 }
  \label{app:fig:gpt-4o-mini-distance}
 \end{figure*}

\begin{figure*}[ht]
 \centering
 \frame{\includegraphics[width=0.24\linewidth]{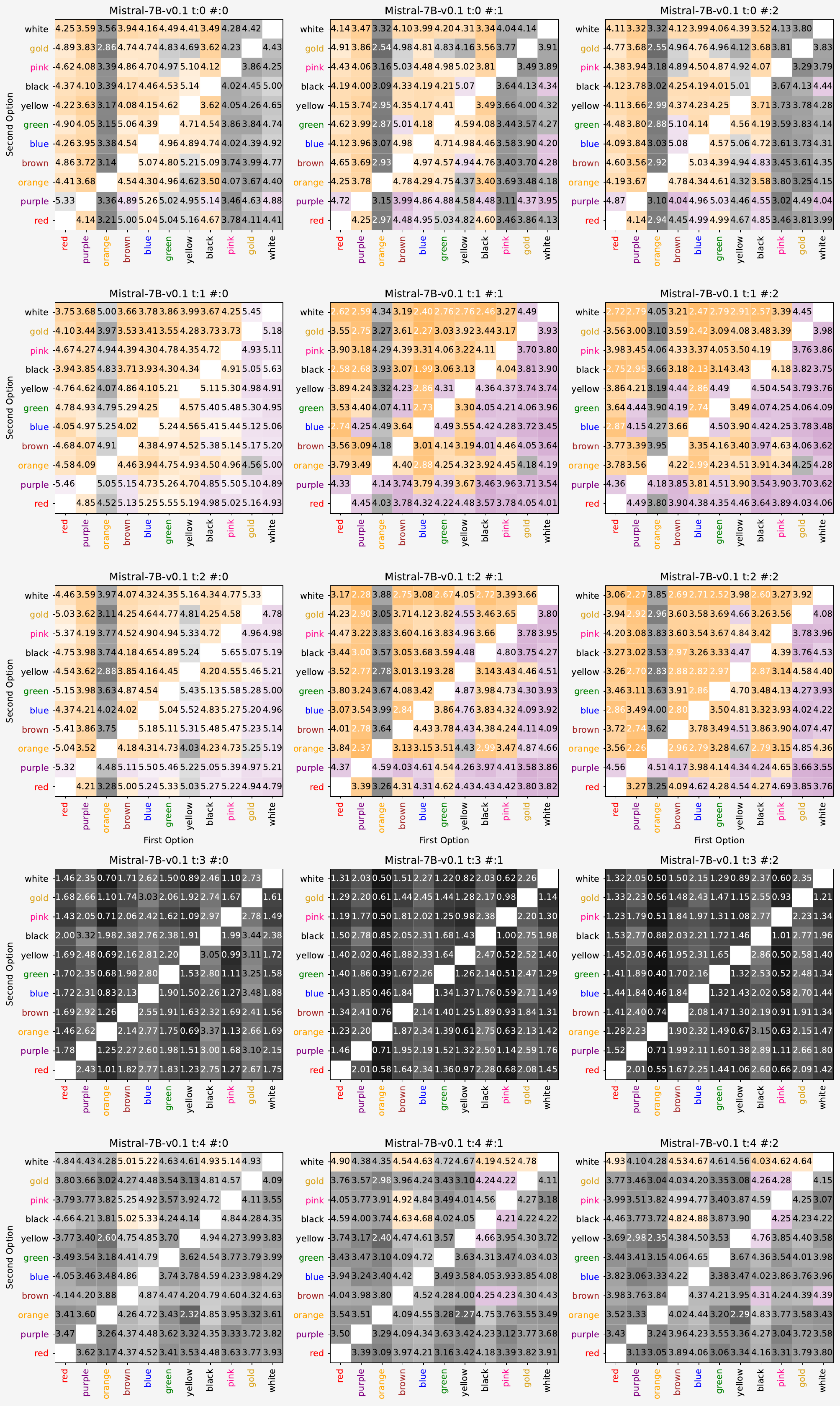}}
 \frame{\includegraphics[width=0.24\linewidth]{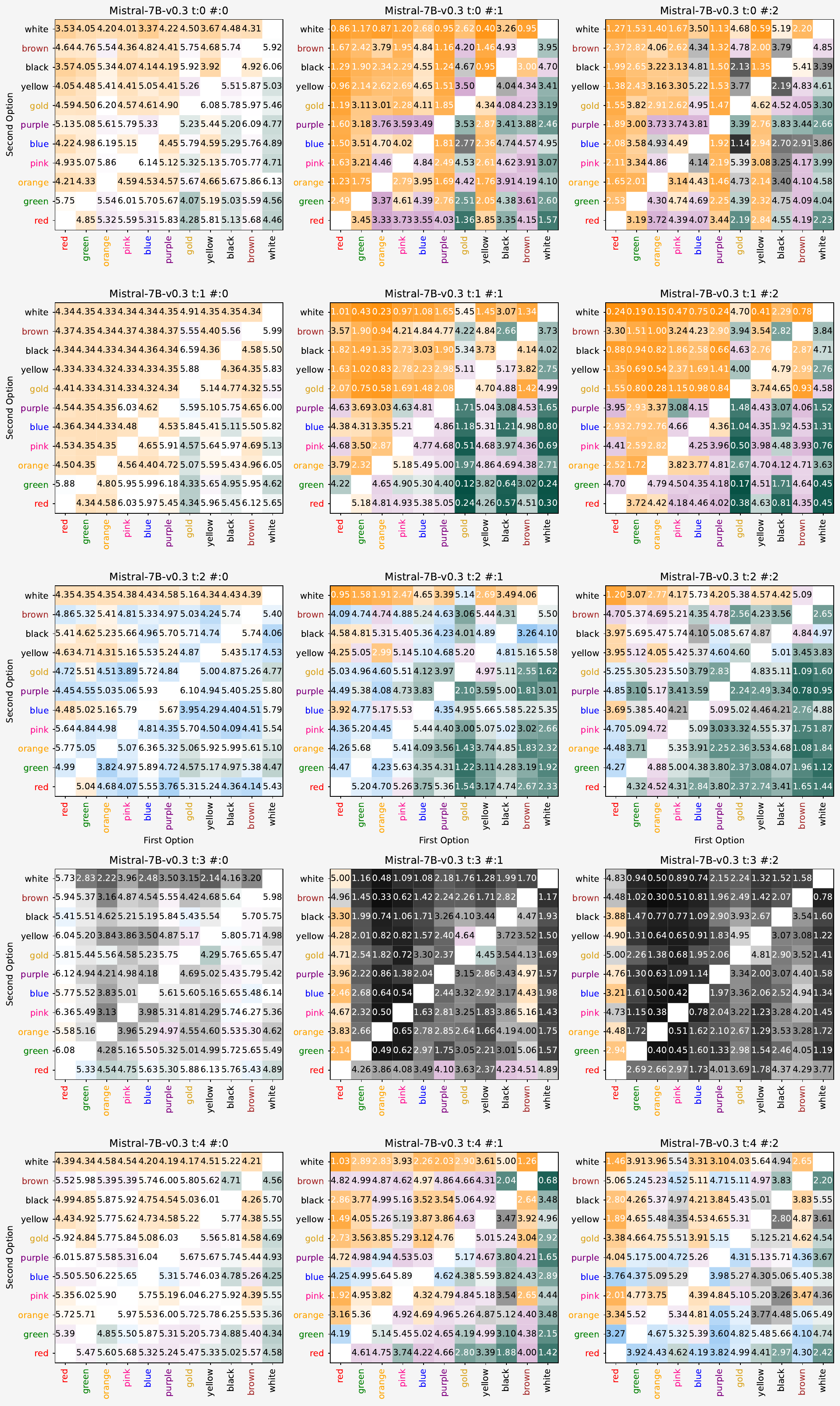}}
 \frame{\includegraphics[width=0.24\linewidth]{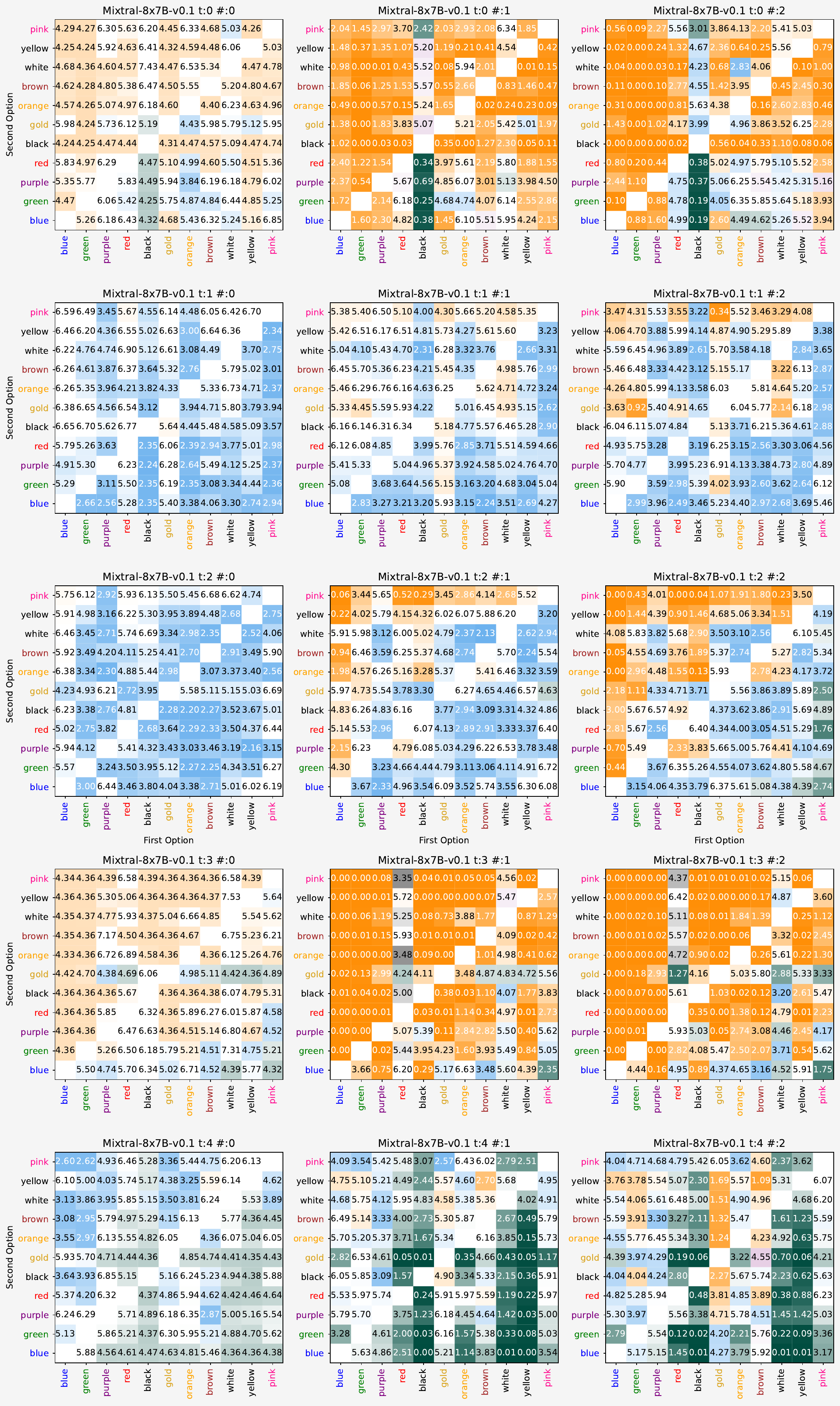}}
 \frame{\includegraphics[width=0.16\linewidth]{appendix/figures-keynote/big-big-col-labels-a.pdf}}
 \\
 \vspace{0.05cm}
\frame{\includegraphics[width=0.24\linewidth]{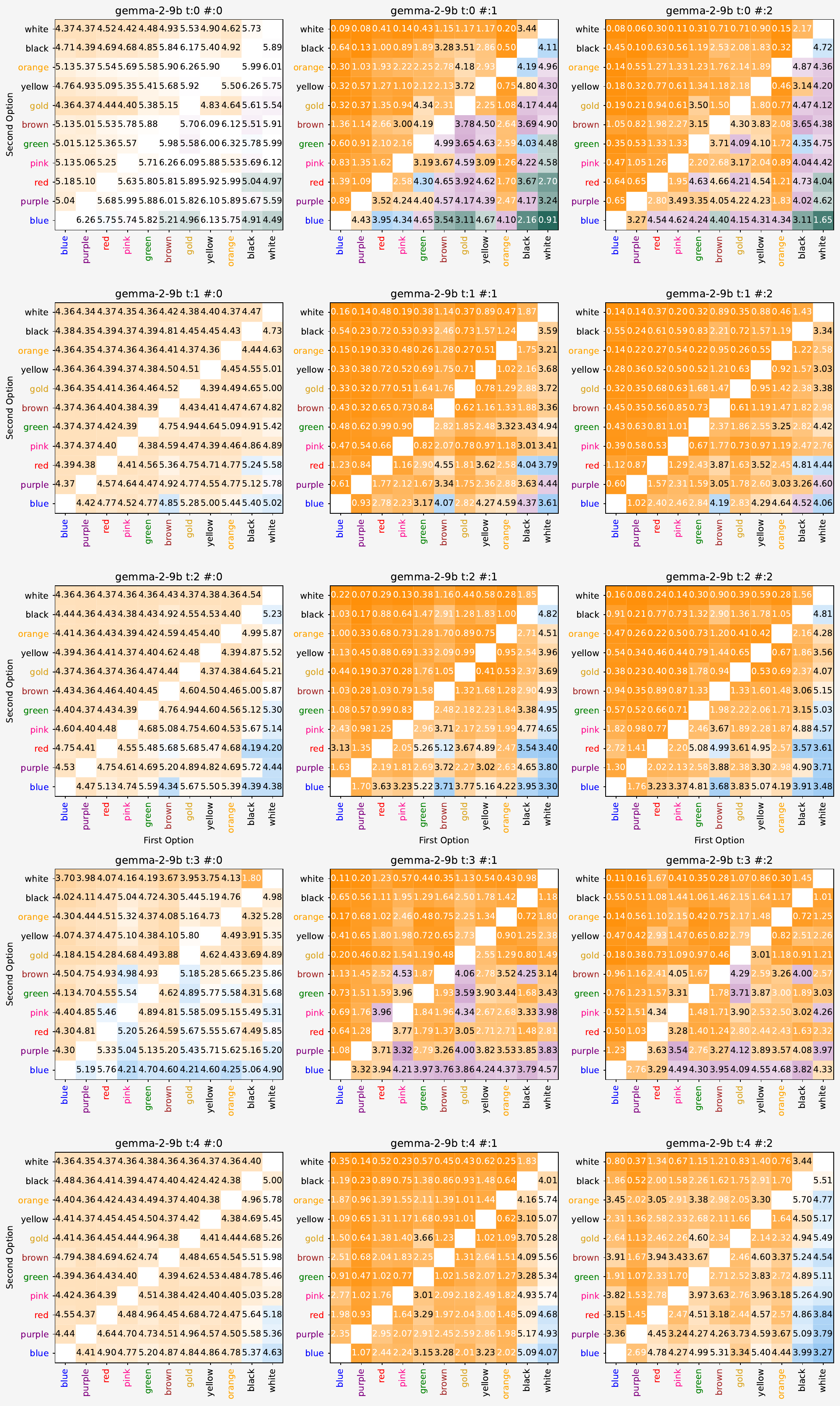}}
\frame{\includegraphics[width=0.24\linewidth]{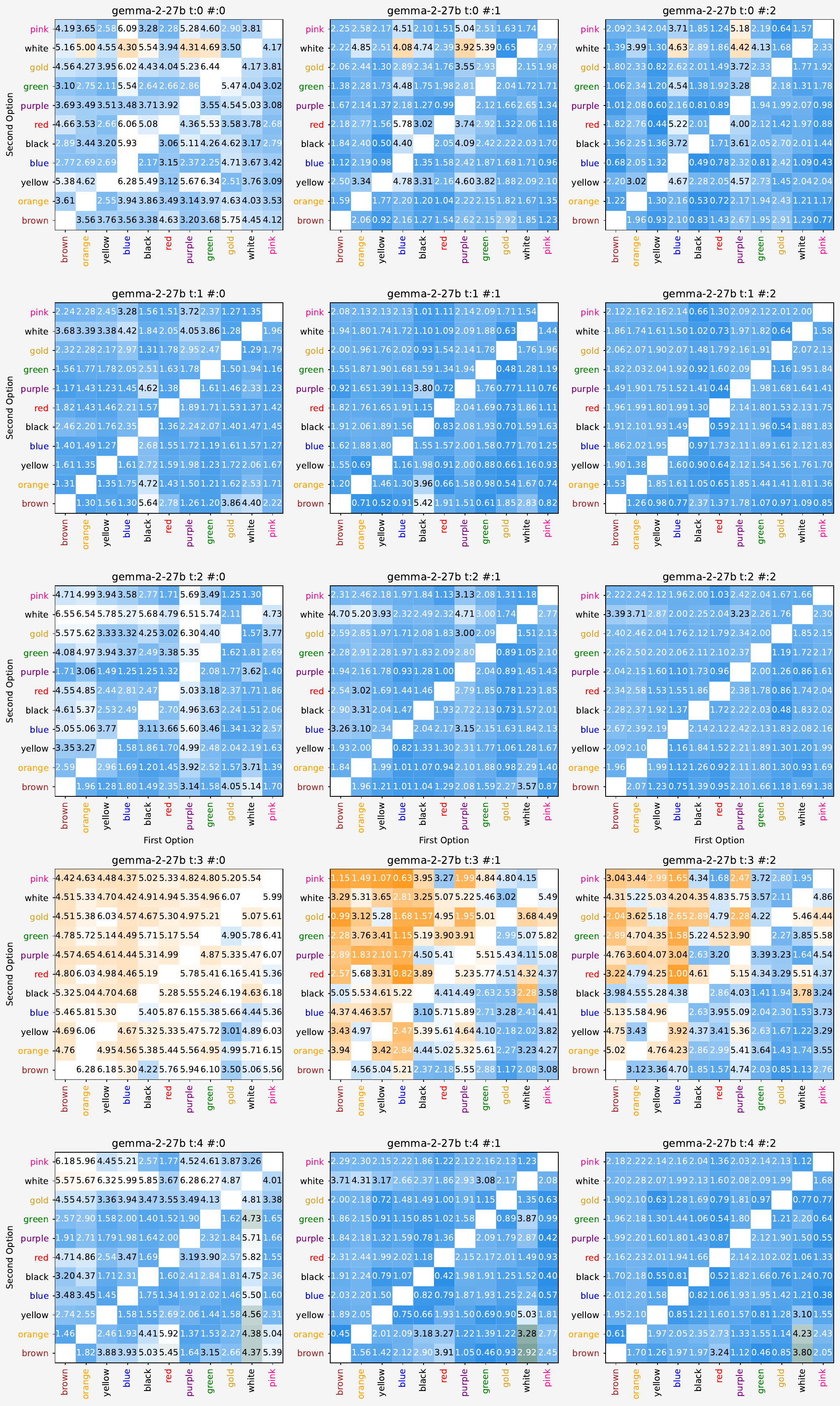}}
\frame{\includegraphics[width=0.24\linewidth]{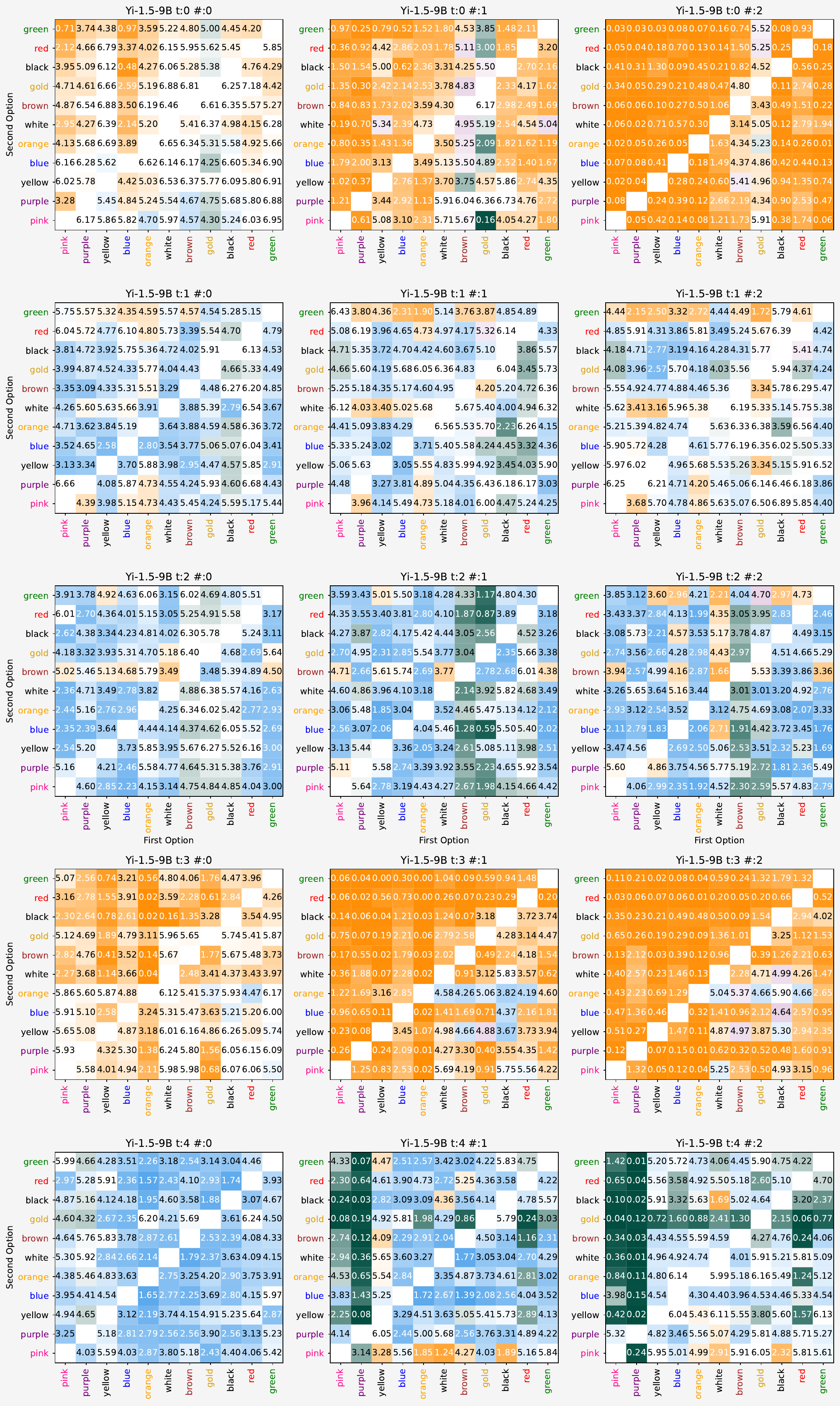}}
 \frame{\includegraphics[width=0.16\linewidth]{appendix/figures-keynote/big-big-col-labels-b.pdf}}
\\
 \vspace{0.05cm}
\frame{\includegraphics[width=0.24\linewidth]{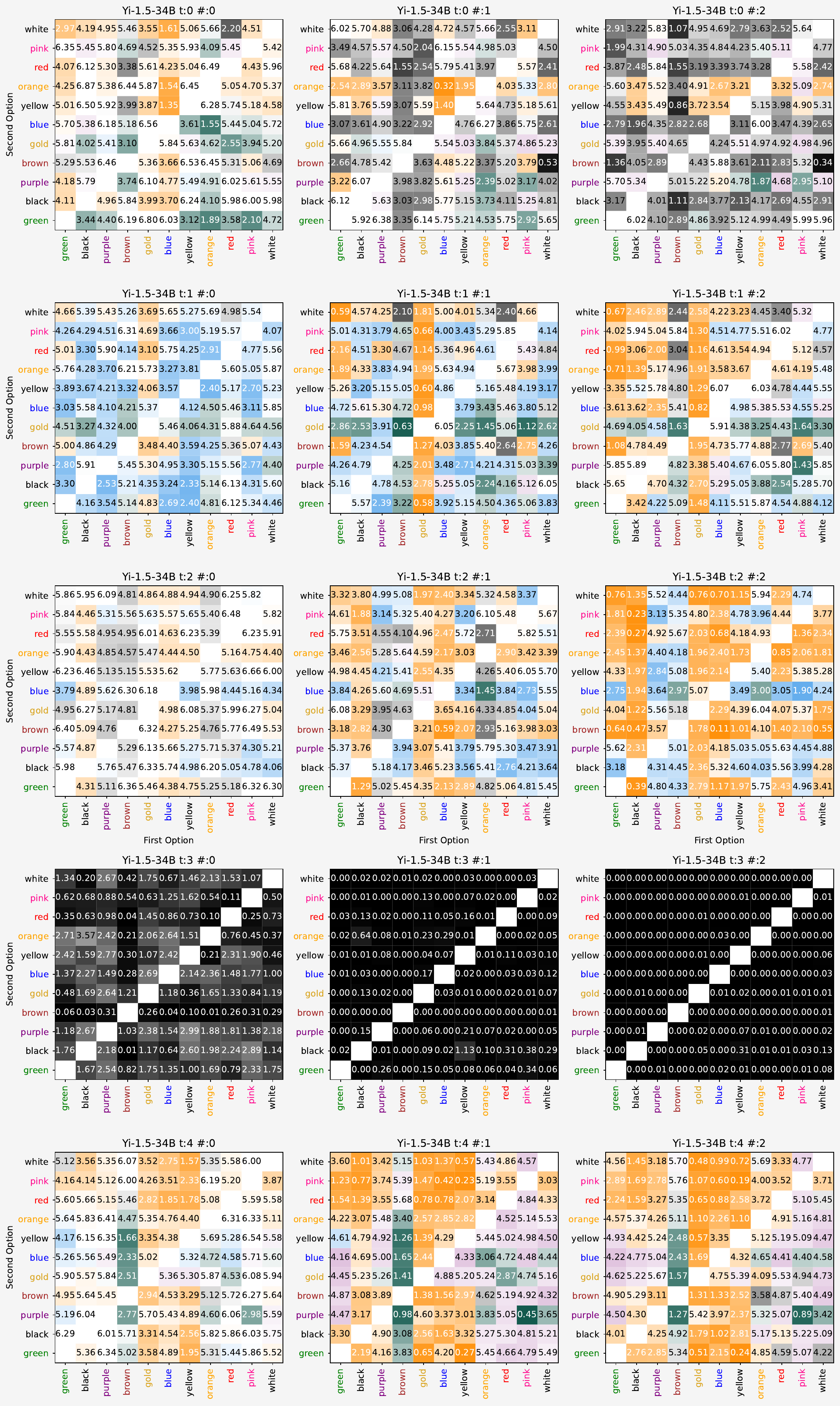}}
\frame{\includegraphics[width=0.24\linewidth]{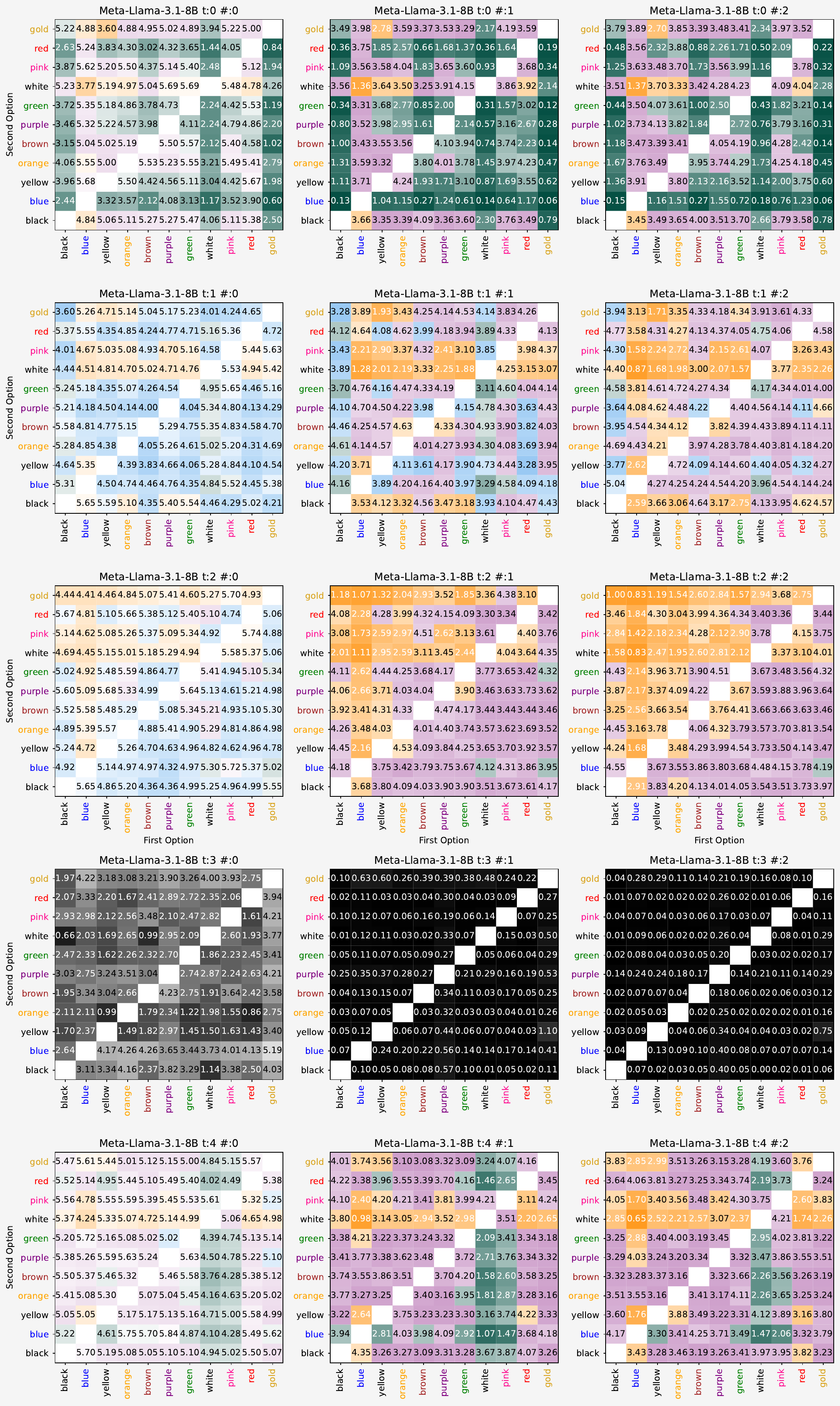}}
\frame{\includegraphics[width=0.24\linewidth]{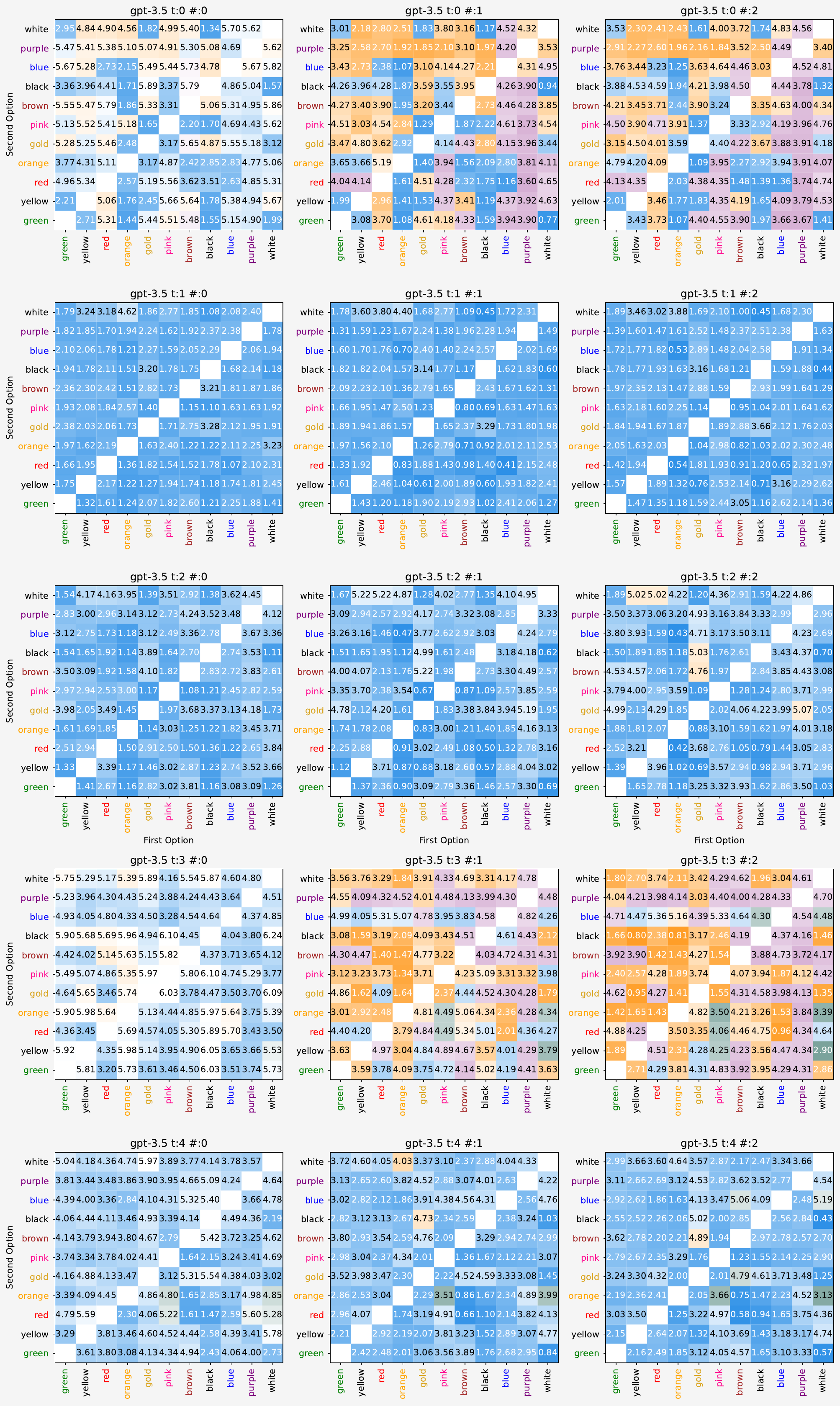}}
 \frame{\includegraphics[width=0.16\linewidth]{appendix/figures-keynote/big-big-col-labels-c.pdf}}
\\
 \vspace{0.05cm}
\frame{\includegraphics[width=0.24\linewidth]{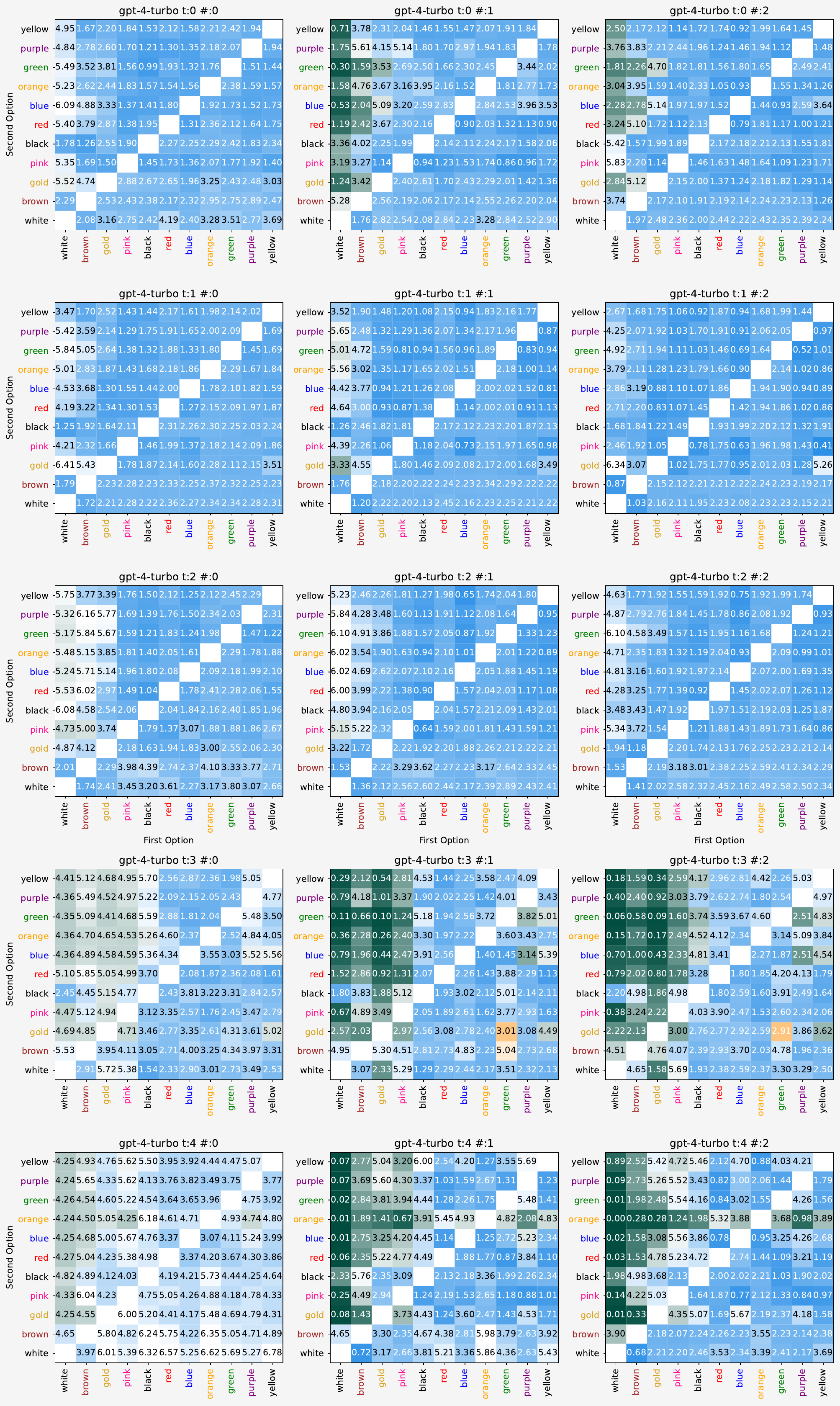}}
\frame{\includegraphics[width=0.24\linewidth]{appendix/figures-heatmaps-distance/distance-summary-gpt-4o-mini-all-no_norm-portrait.pdf}}
\frame{\includegraphics[width=0.24\linewidth]{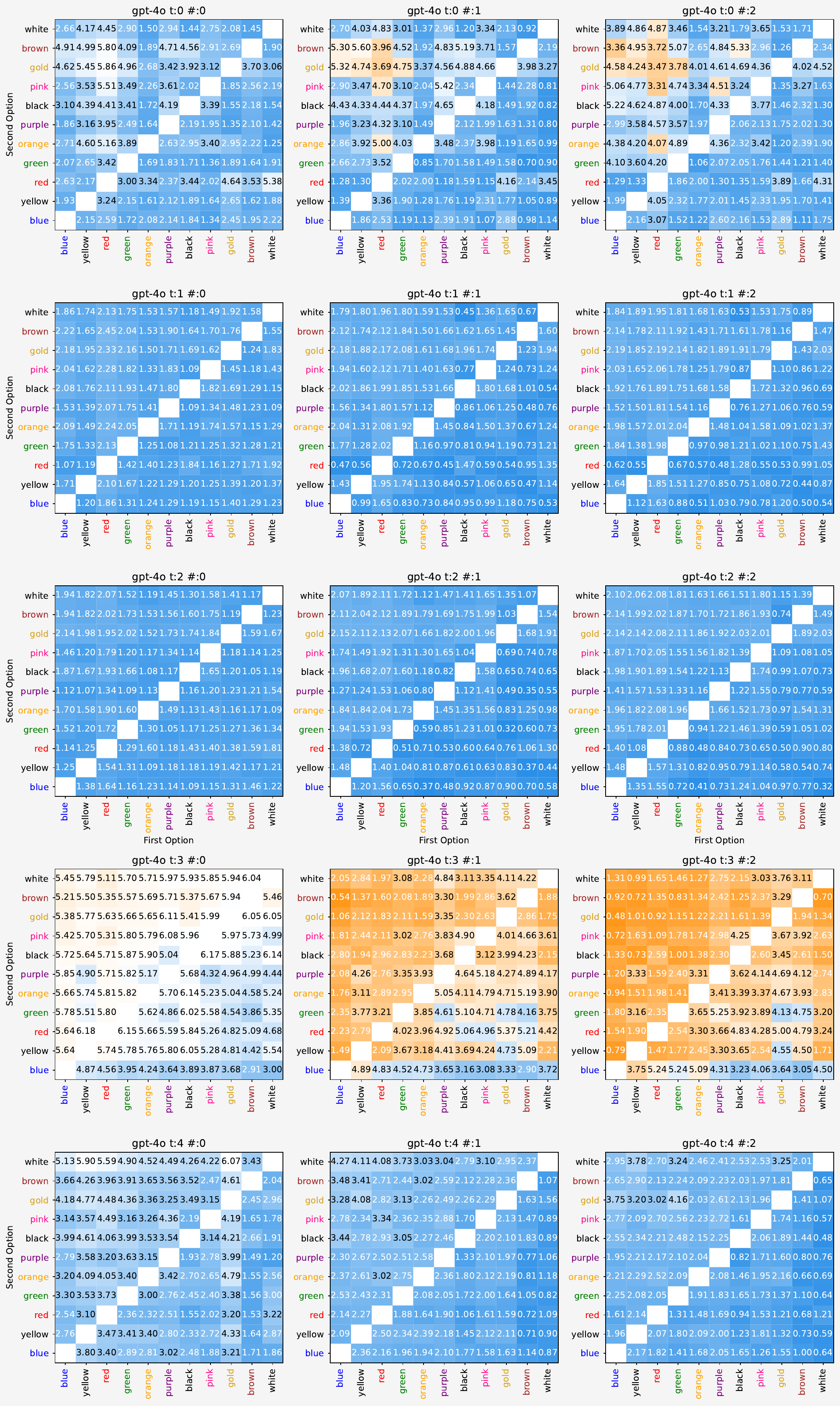}}
 \frame{\includegraphics[width=0.16\linewidth]{appendix/figures-keynote/big-big-col-labels-d.pdf}}
 \caption{\textbf{Space Station View of Model Behaviors.}  }
  \label{app:fig:summary-summary:distance}
\end{figure*}

\paragraph{We categorize model output behavior over sets of results by using computing the distance to reference behaviors.}
\Cref{fig:model_behavior_bar_plot} computes the proportion of each behavior by classifying a set of results (across an number scale, template, and option pair) as one of the reference points. This is done by measuring the WD between the output probabilities and reference point. We additionally classify any grid where the average PM is less than 0.5 as Null. Meaning, anything where more than 50\% of the probability mass is on tokens outside of $T$ are considered out of scope. This is how the results in \Cref{app:fig:summary-summary:distance}, \Cref{app:fig:gpt-4o-mini-distance}, and \Cref{app:fig:legend:distance} are computed.

\begin{figure*}[ht!]
 \centering
 \includegraphics[width=0.95\linewidth]{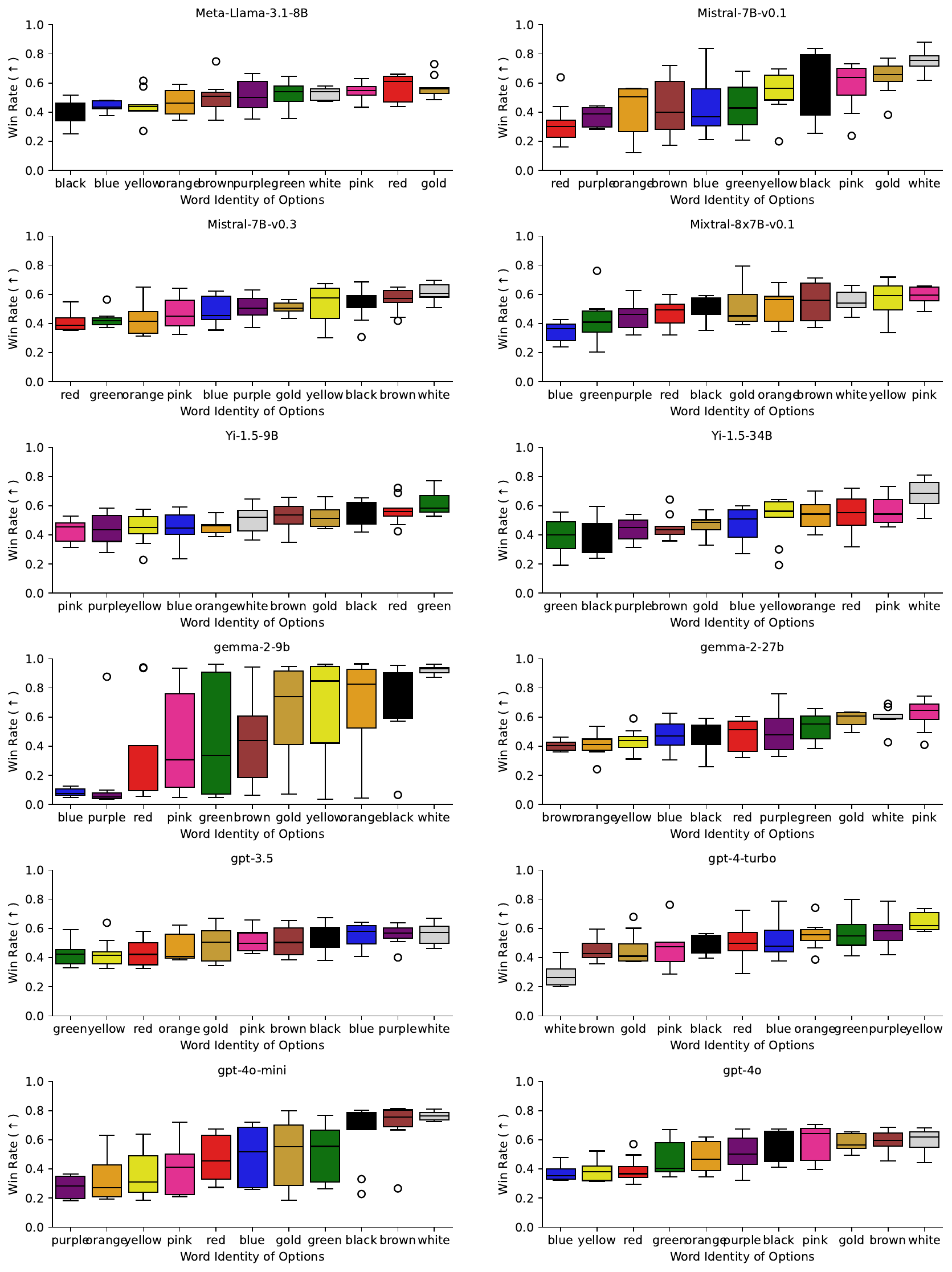}
 \caption\textbf{{Win rates correspond to option order.}}
 \label{app:fig:color_preferences:wins}
\end{figure*}

\begin{figure*}[ht!]
 \centering
 \includegraphics[width=0.95\linewidth]{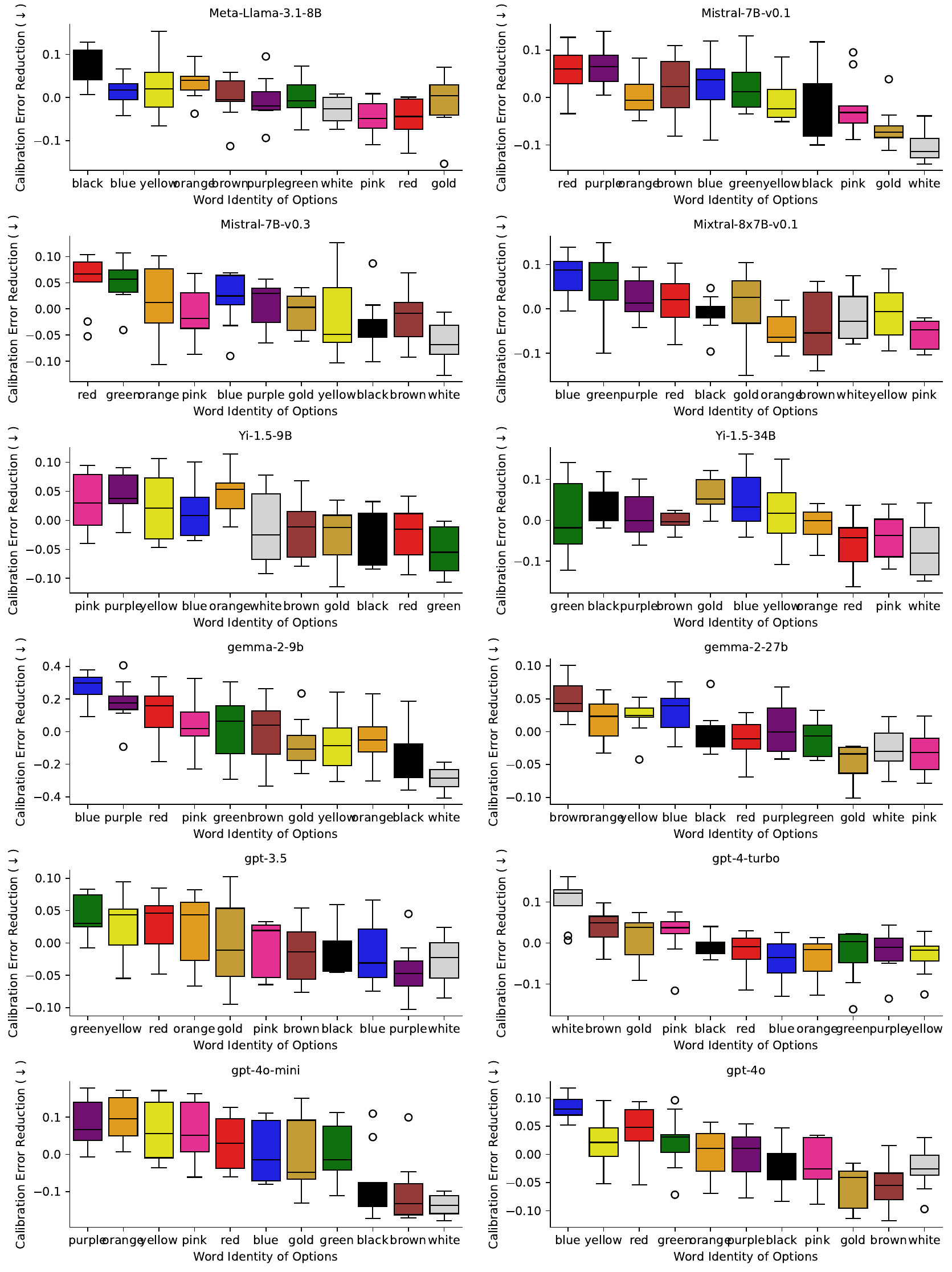}
 \label{app:fig:color_preferences:wd}
 \caption{\textbf{Reductions in error that correspond to option order.} 
 }

 \label{fig:color_preferences}
\end{figure*}

\paragraph{The bias across for individual option words varies model to model, but there is some shared ordering.}
\Cref{app:fig:color_preferences:wins} shows how win rates change when a color is placed before vs after other colors. If the color did not matter we would expect the win rates to be around 50\% (with more variance). Instead, there are clear rank orders. \Cref{app:fig:color_preferences:wd} shows the average change in calibration (lower is better, or a reduction in calibration error).

When looking at the hierarchies formed by the options on the X-axes of \Cref{app:fig:color_preferences:wins} there appears to be some shared structure across models. For example, for most models, listing the color white first leads to an improvement in calibration. Notably, this does not mean the models prefer this color or pick it. Always picking it would not be particularly calibrated. The example we saw in the main body of the paper was the \gpt{4o-mini} was biased against picking white when it was second and tended to Pick Higher when it was first.

\Cref{app:tab:rankings_corr} shows the correlation of the hierarchy orderings across models. The structure similarity (and dissimilarity) is sometimes far beyond random. Similar models generally have more correlated orderings. \gpt{4-turbo} is the most dissimilar from others.

\begin{figure*}[ht!]
 \centering
 \includegraphics[width=0.55\linewidth]{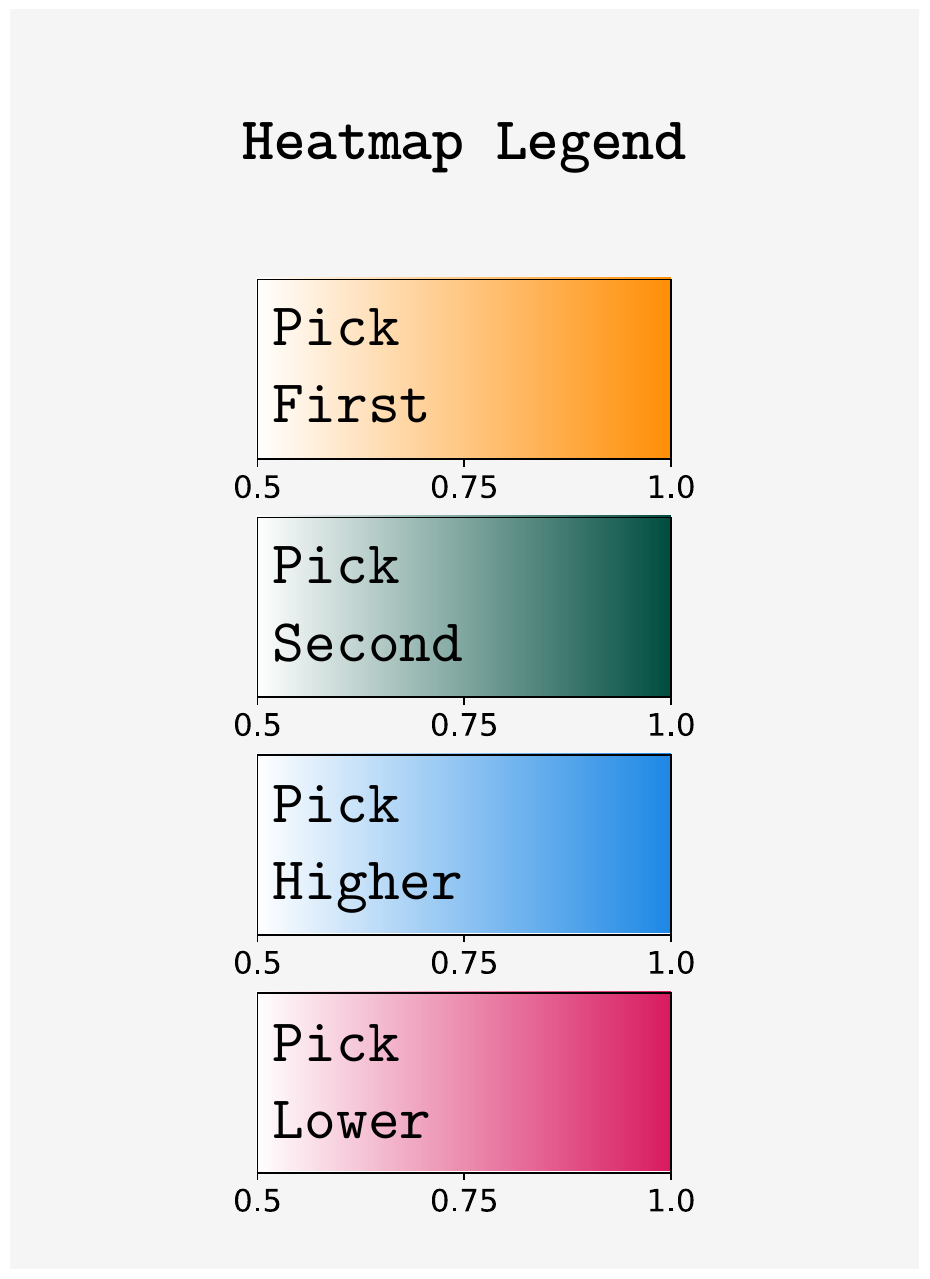}
 \caption{\textbf{Legend for heatmaps on the following page.} 
 }
 \label{app:fig:legend}
\end{figure*}

\begin{figure*}[ht]
 \centering
 \frame{\includegraphics[width=0.24\linewidth]{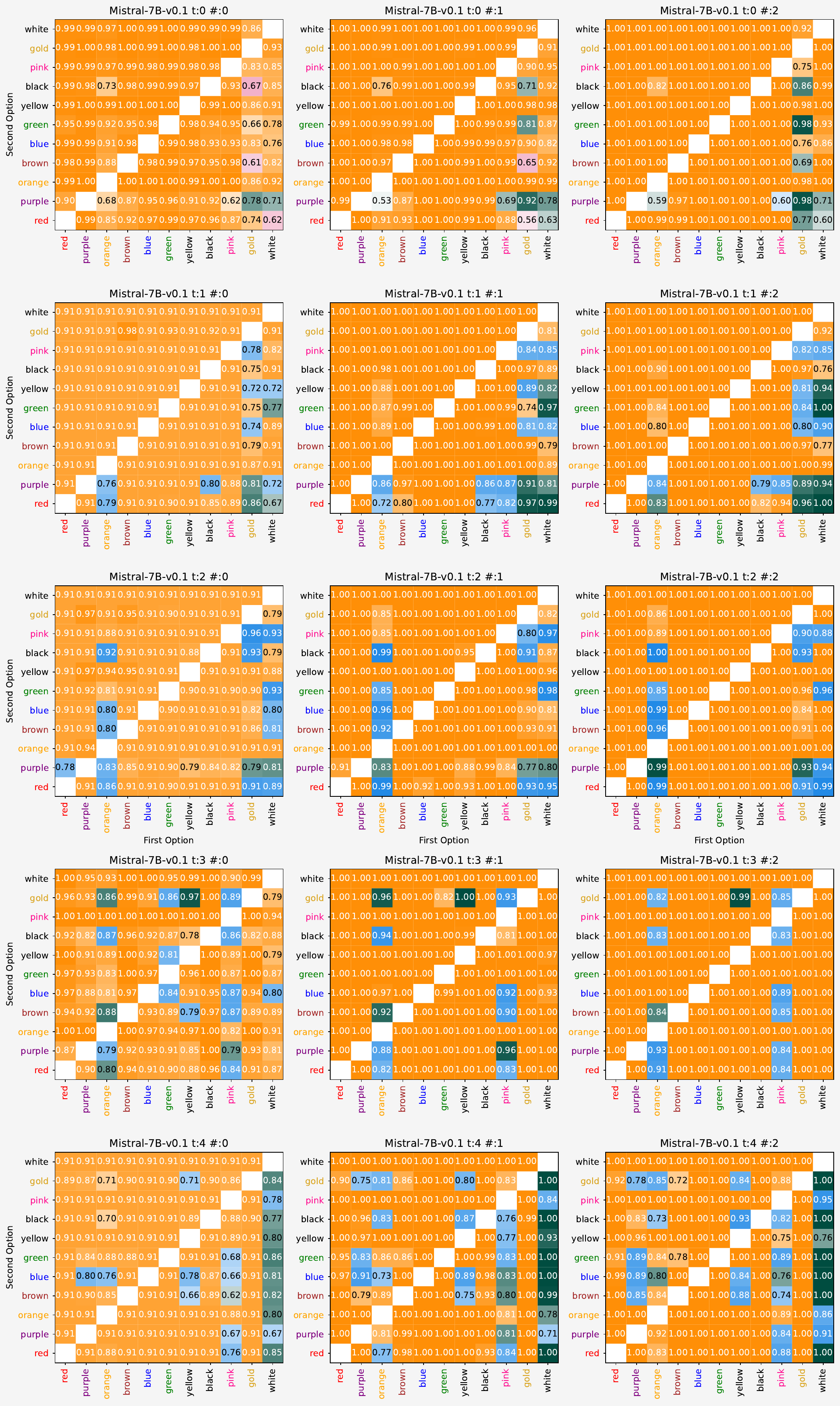}}
 \frame{\includegraphics[width=0.24\linewidth]{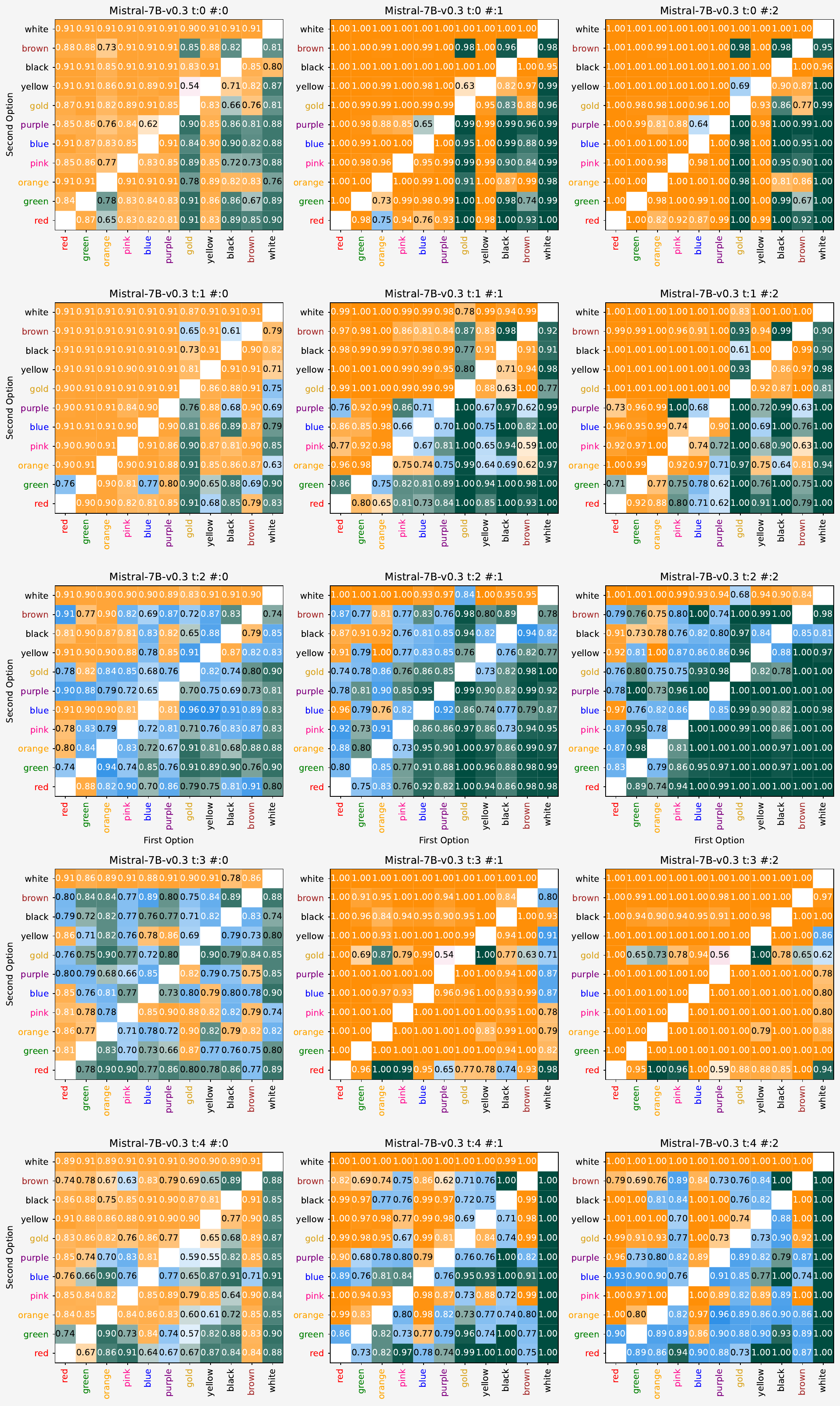}}
 \frame{\includegraphics[width=0.24\linewidth]{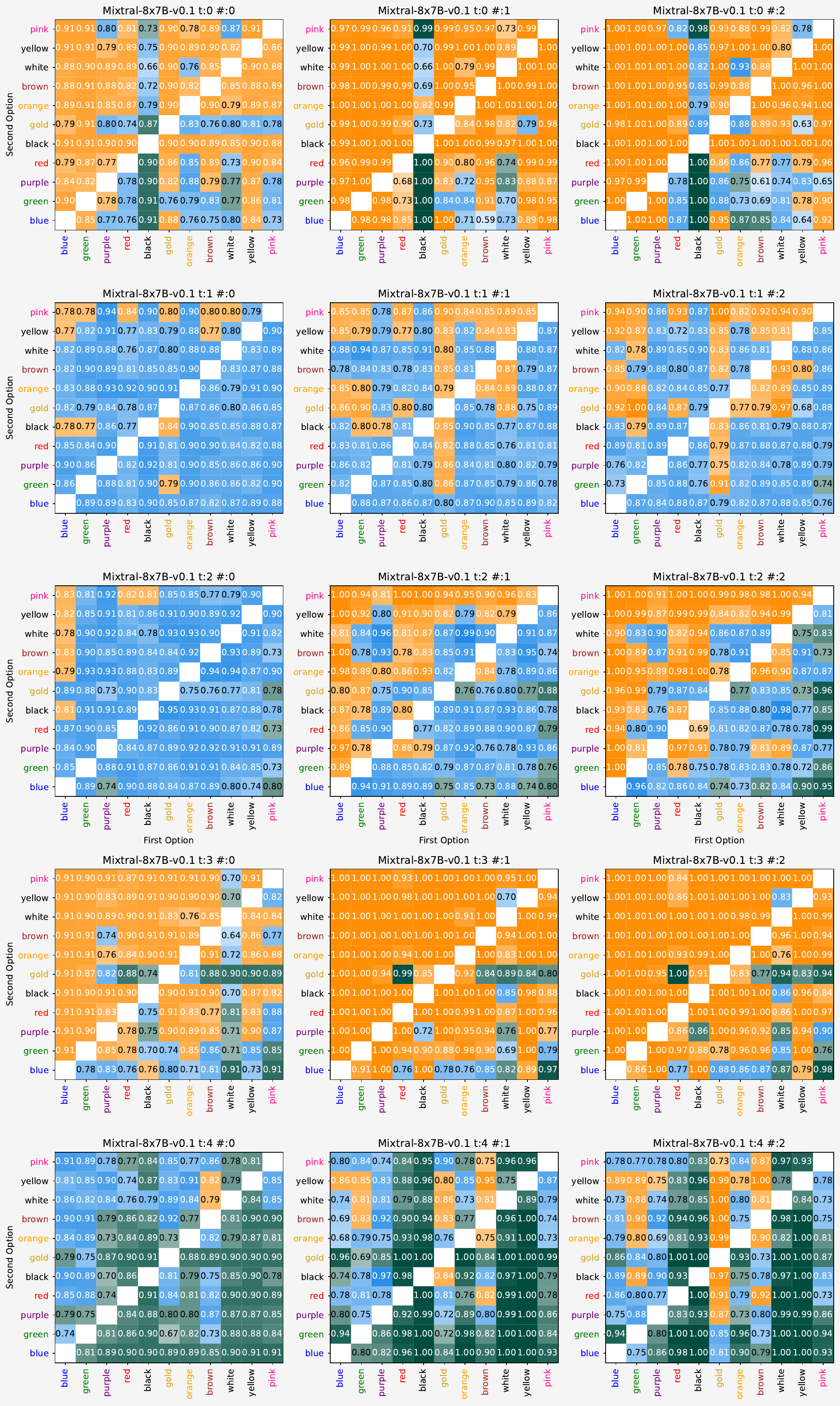}}
 \frame{\includegraphics[width=0.16\linewidth]{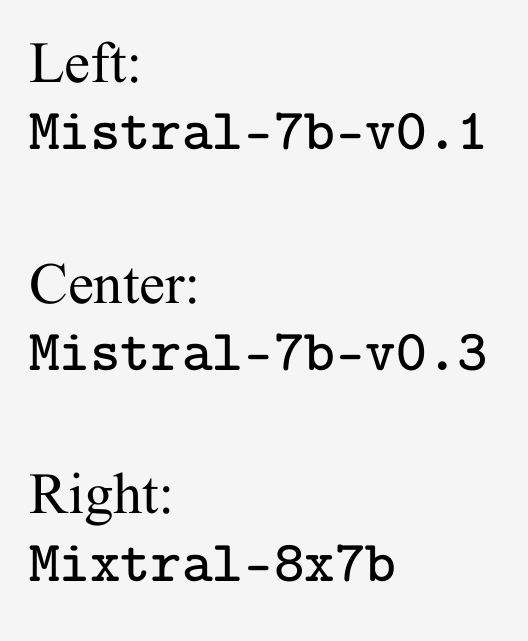}}
 \\
 \vspace{0.05cm}
 \frame{\includegraphics[width=0.24\linewidth]{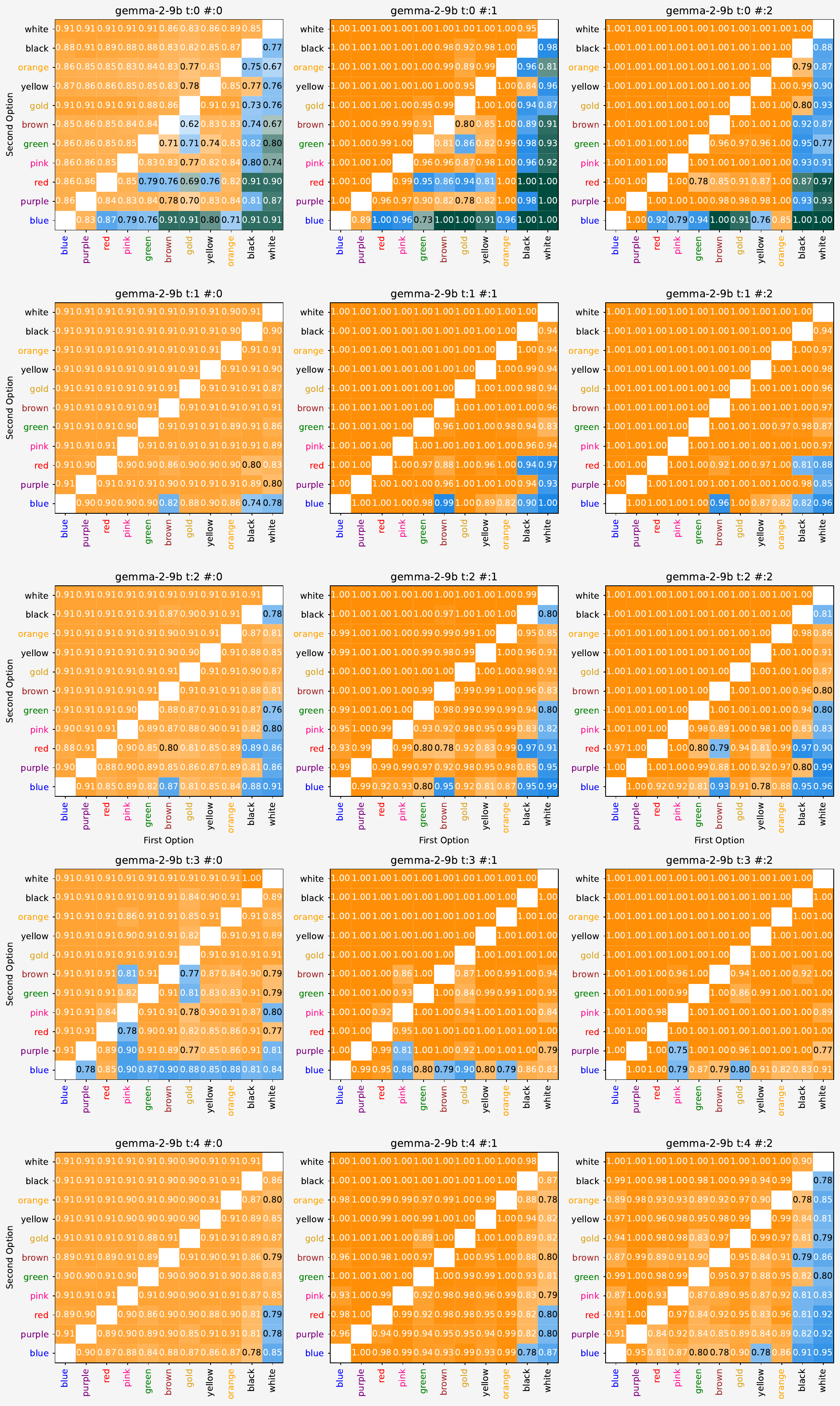}}
 \frame{\includegraphics[width=0.24\linewidth]{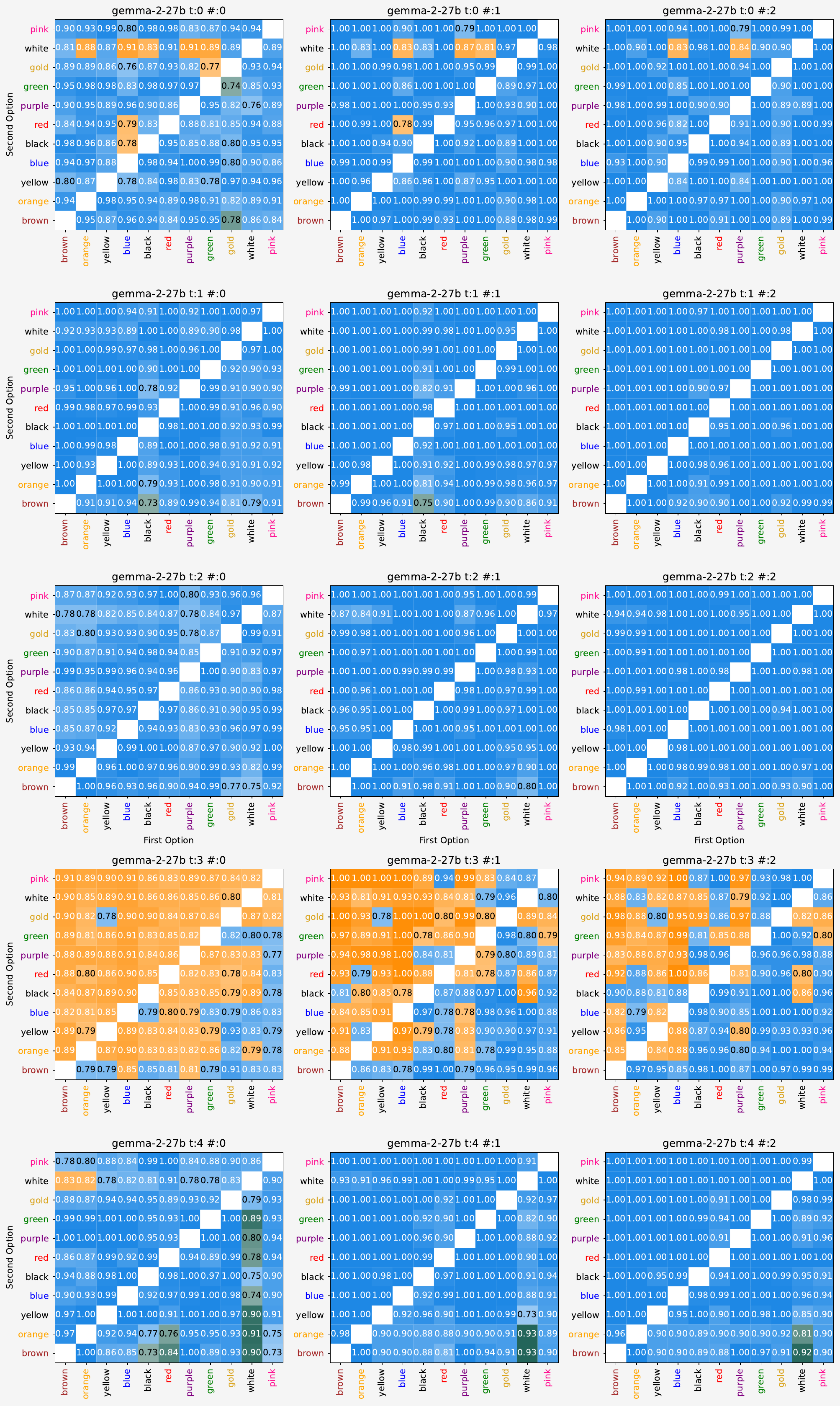}}
 \frame{\includegraphics[width=0.24\linewidth]{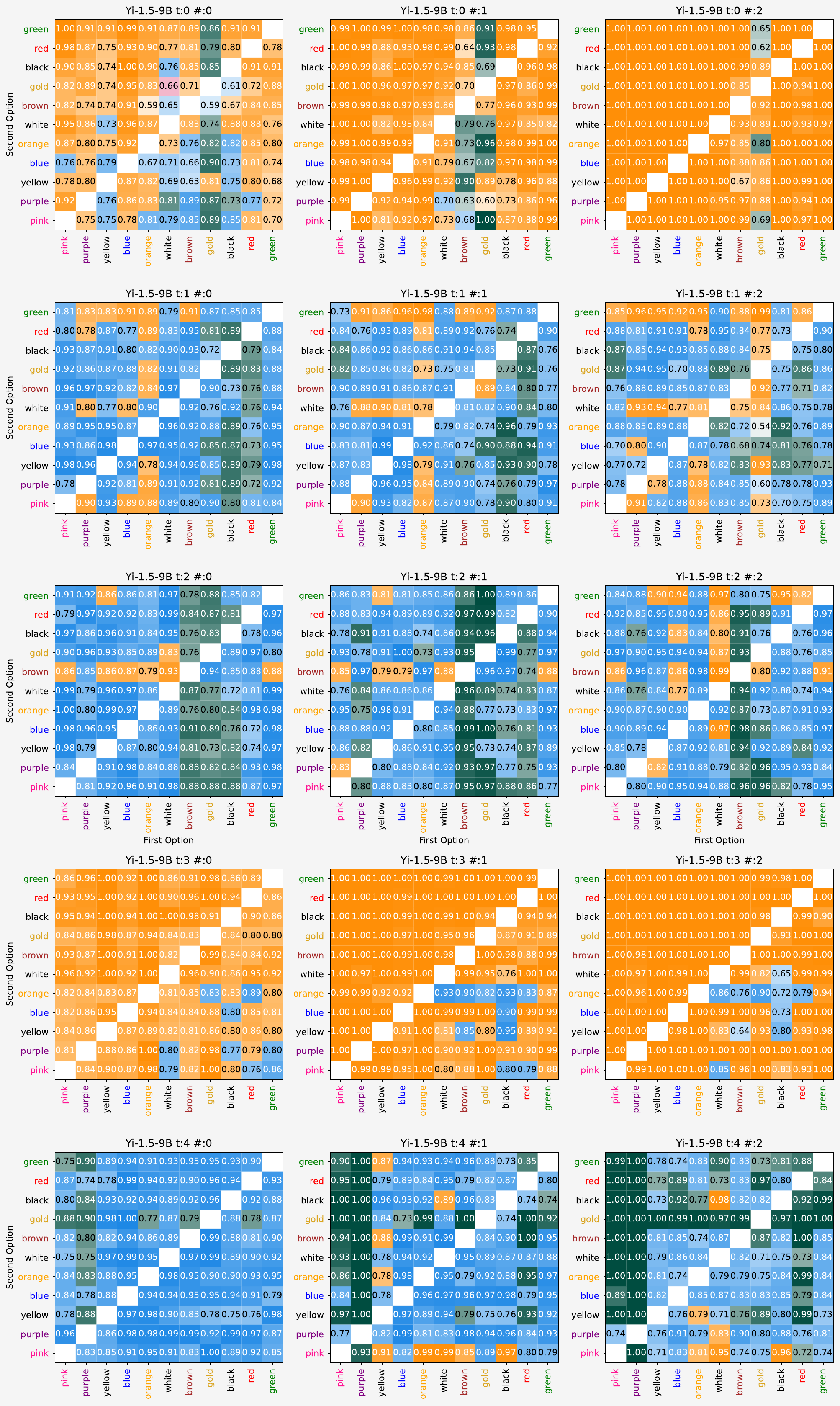}}
 \frame{\includegraphics[width=0.16\linewidth]{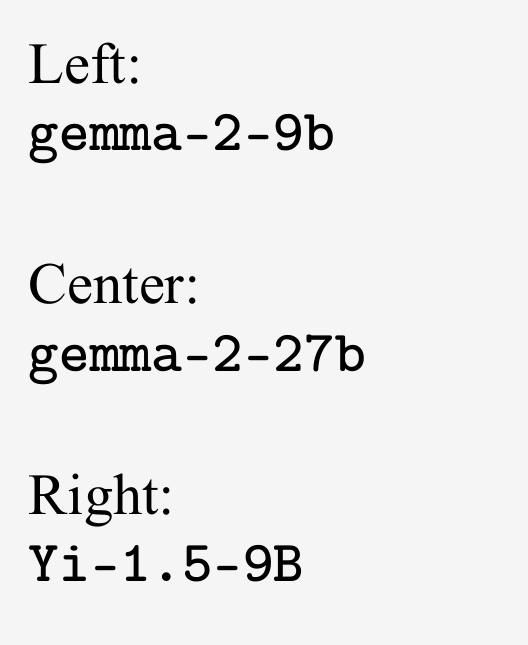}}
\\
 \vspace{0.05cm}
  \frame{\includegraphics[width=0.24\linewidth]{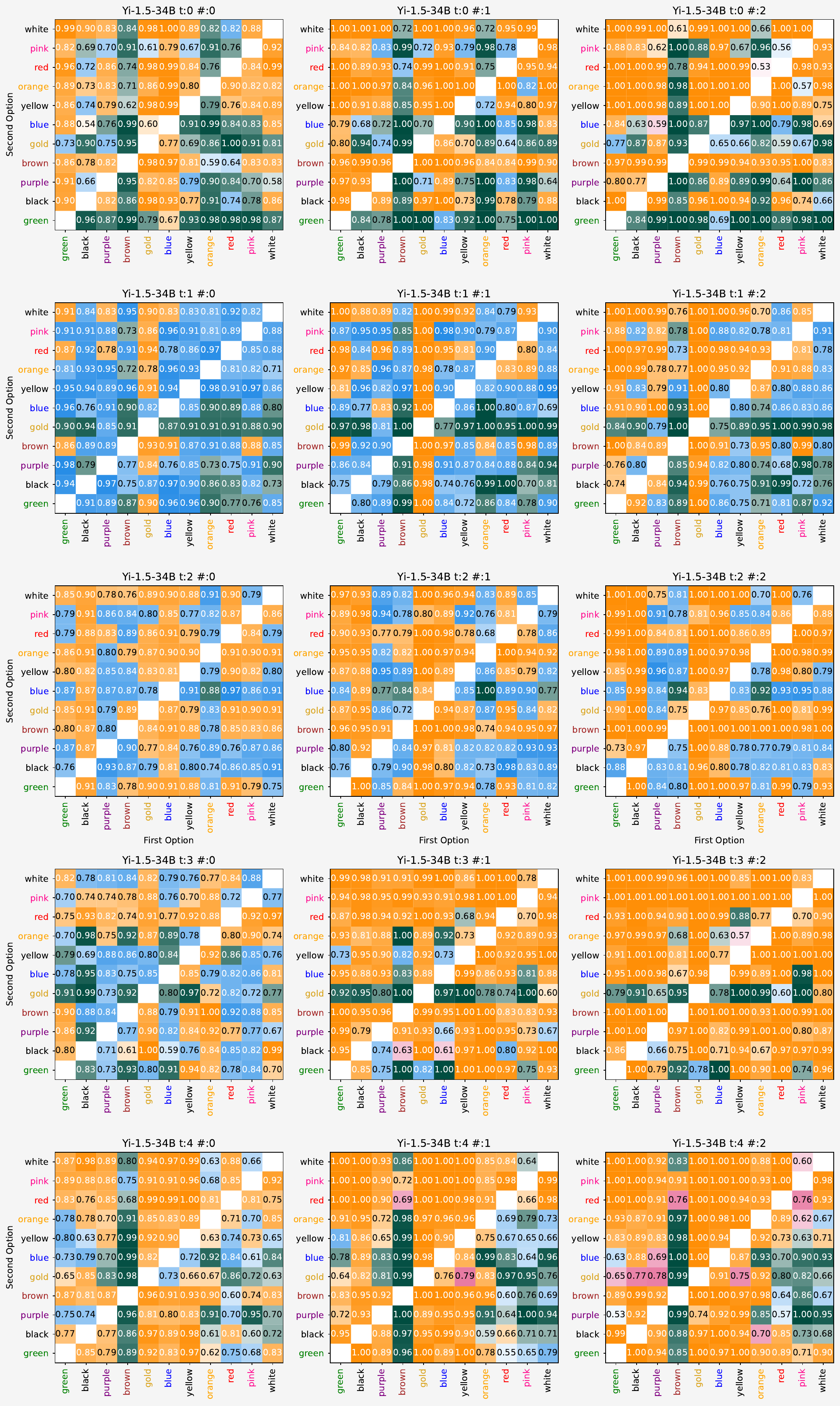}}
  \frame{\includegraphics[width=0.24\linewidth]{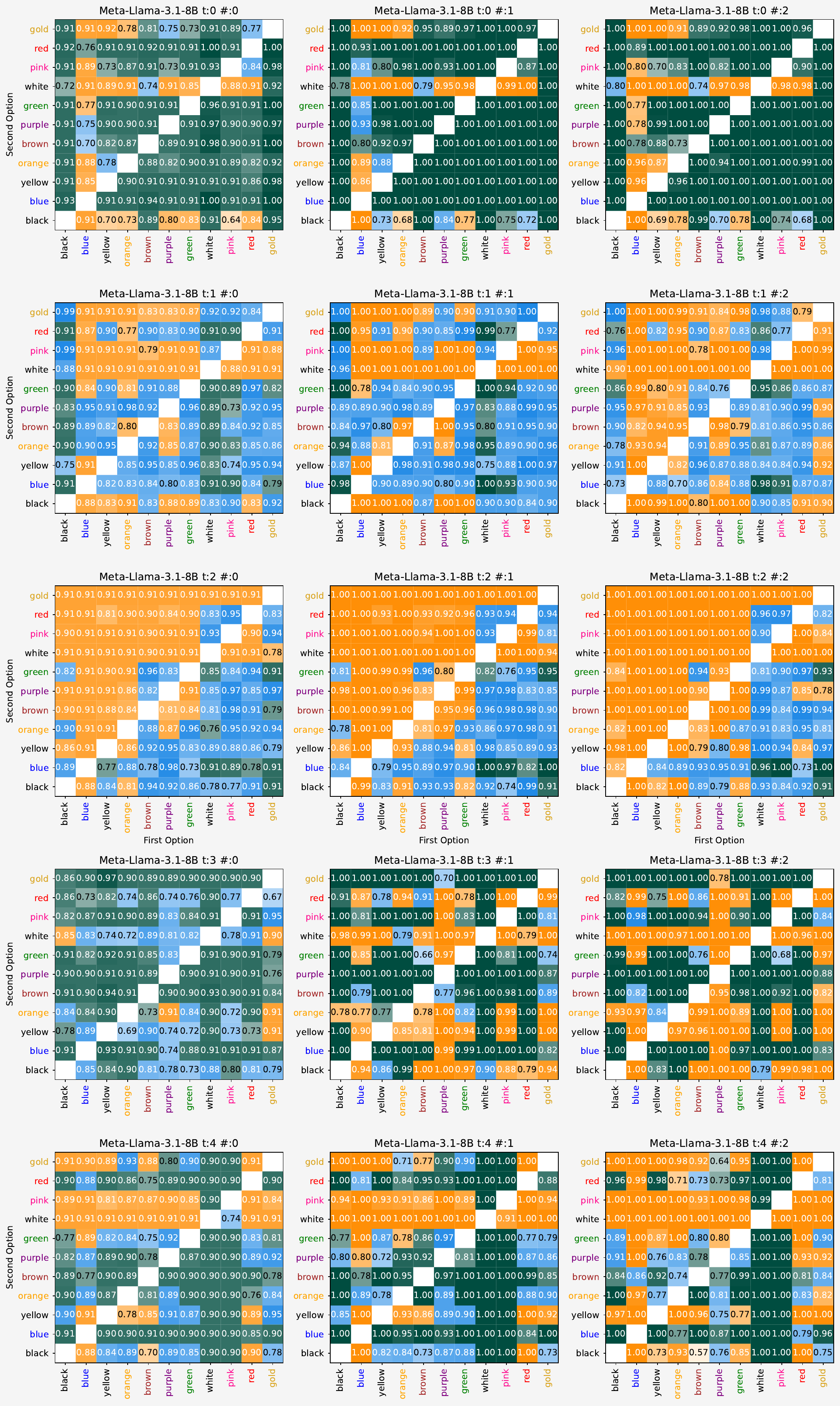}}
  \frame{\includegraphics[width=0.24\linewidth]{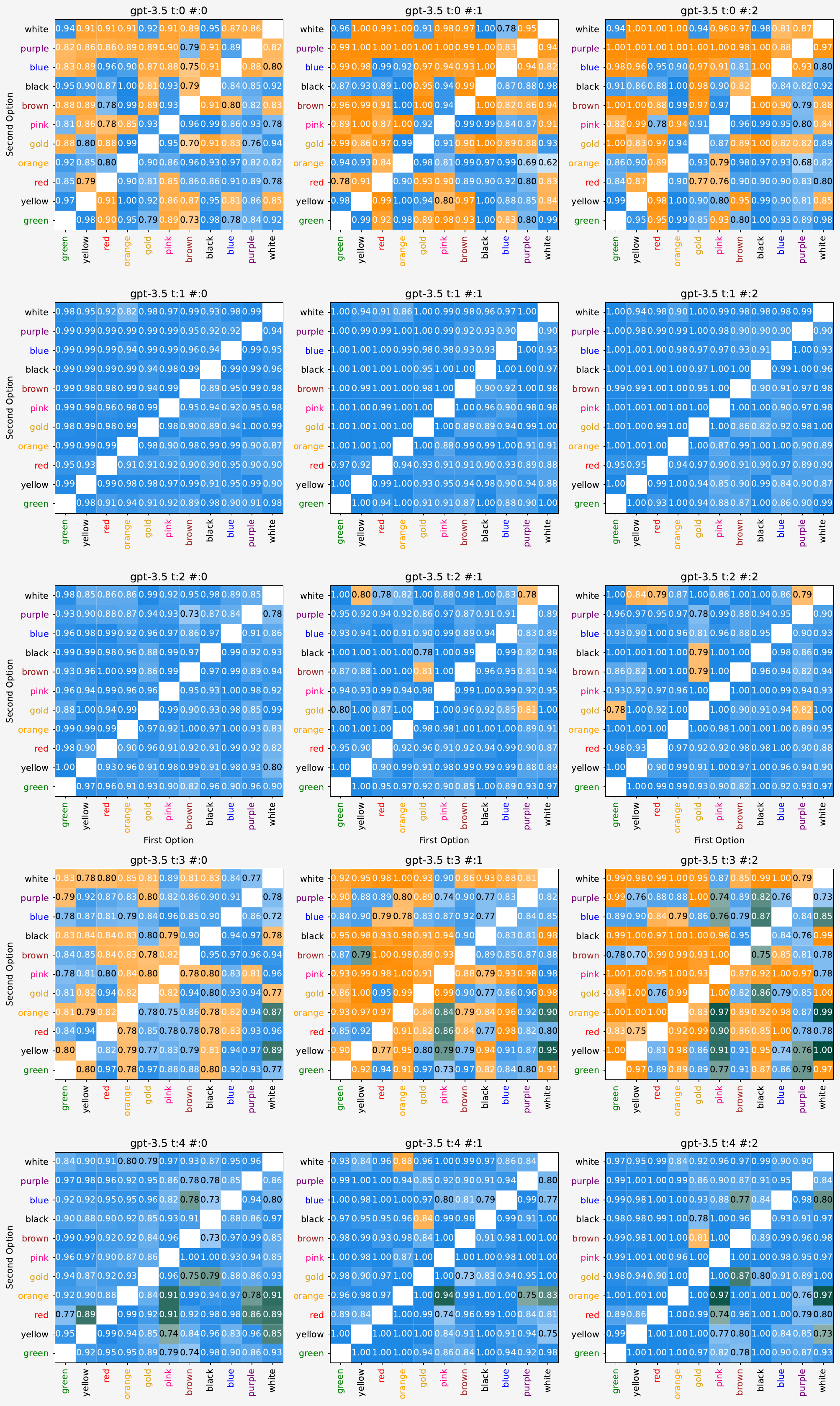}}
 \frame{\includegraphics[width=0.16\linewidth]{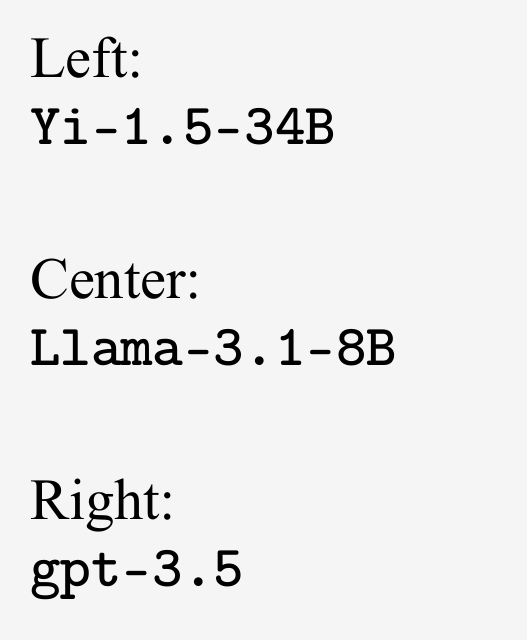}}
\\
 \vspace{0.05cm}
  \frame{\includegraphics[width=0.24\linewidth]{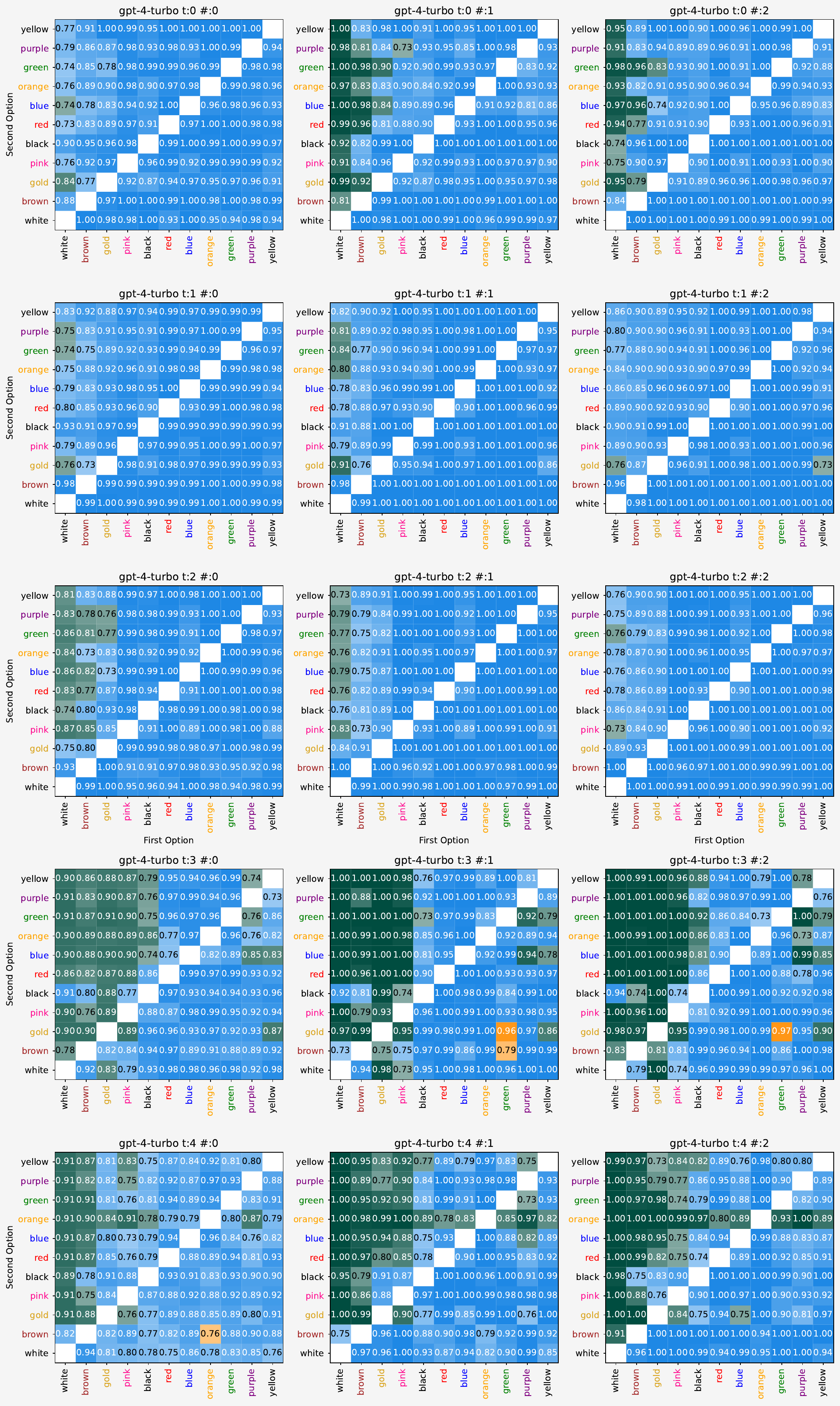}}
  \frame{\includegraphics[width=0.24\linewidth]{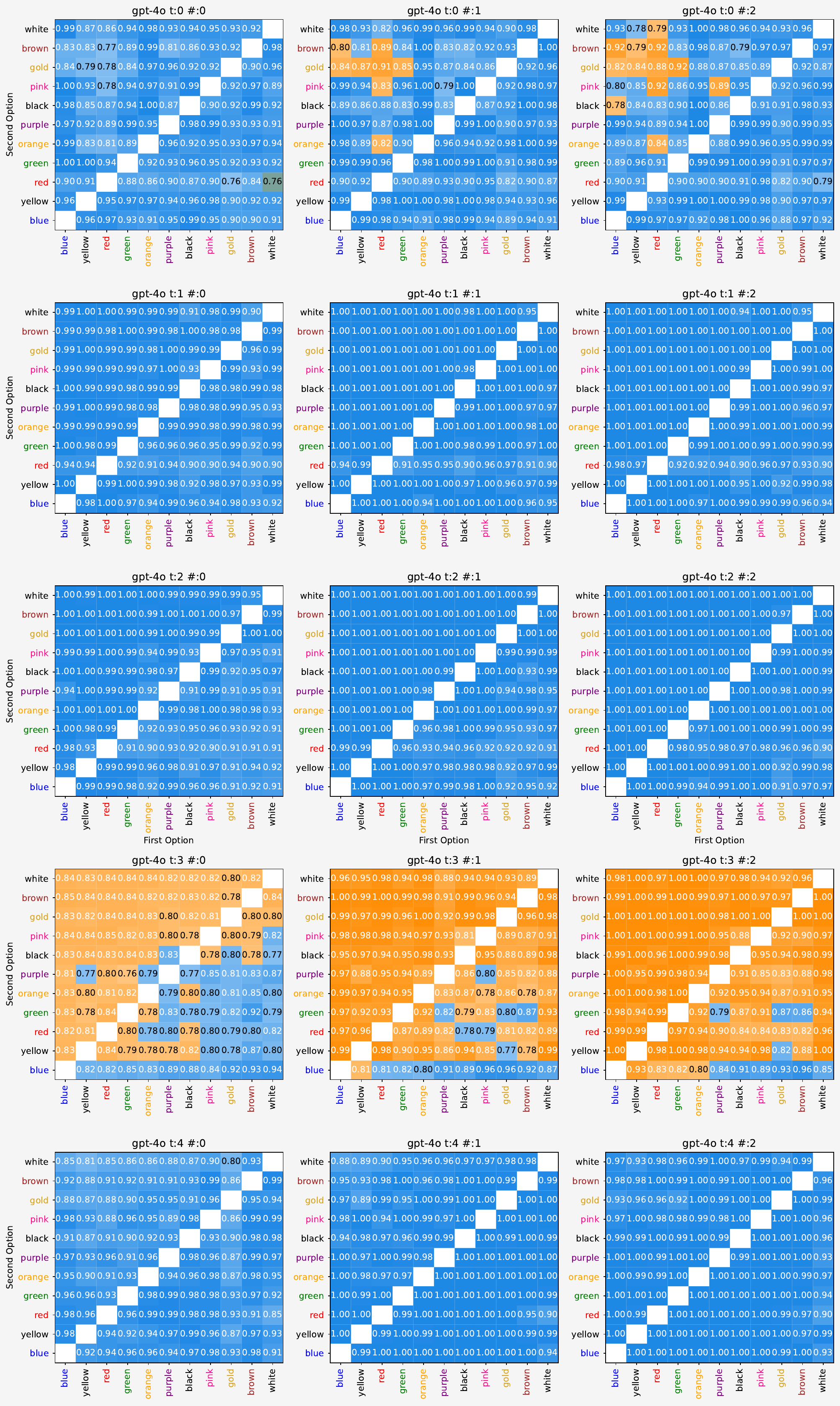}}
 \frame{\includegraphics[width=0.24\linewidth]{appendix/figures-heatmaps-rate/heatmap-gpt-4o-mini-rates.pdf}}
 \frame{\includegraphics[width=0.16\linewidth]{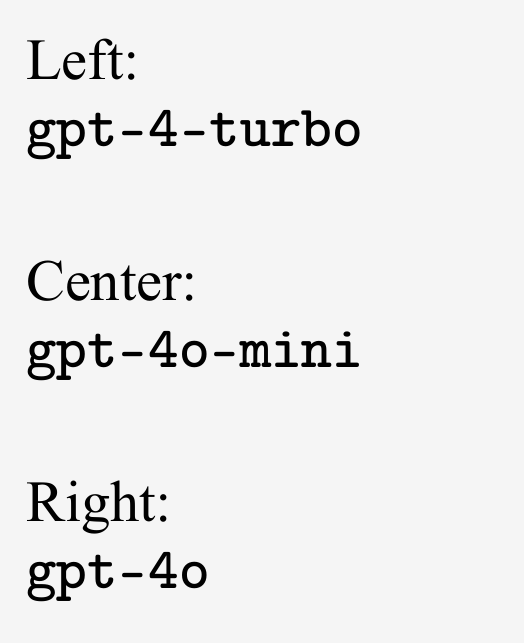}}
 \caption{\textbf{Space Station View of Model Behaviors.}  }
  \label{app:fig:summary-summary}
\end{figure*}

\paragraph{We show a pictorial/imagistic view of all the data via collections  of heatmaps.} The sets of heatmaps are computed in two different ways.
\Cref{app:fig:summary-summary} shows a \textit{space-station-view} of all the hierarchical, diagonal patterns; \Cref{app:fig:legend} is the legend showing what the colors correspond to. The 15 heatmaps per model are the variations across different templates and number scales. The number scales appear to have relatively little impact on the patterns. The templates, for some models, have a larger impact. The behaviors are computed as in the main body of the paper; only among the references, we report the highest rate of behavior compatible with the model's outputs.

\Cref{app:fig:summary-summary:distance} shows the same results but we create the heatmaps using a different, more high-fidelity approach. We also provide a more zoomed in look at \Cref{app:fig:gpt-4o-mini-distance}. To assign a behavior per cell we now use the distance to each type of behavior. The cells now show distance instead of rate. (Lower is better). This is reflected by the updated legend, \Cref{app:fig:legend:distance}. 

\section{{More Results on Frequency Effects}}\label{app:results:freq}
Here we present more results on the frequency effects and consider alternate independent variables. Firstly, we designed our experiments, in the main, to highlight how the word identity and order impacted calibration. Experiments that examine a wider range of numbers and therefore frequencies would better study this problem. Nonetheless, we find clear and strong effects and report them here.

\begin{figure*}[ht!]
 \centering
 \includegraphics[width=0.95\linewidth]{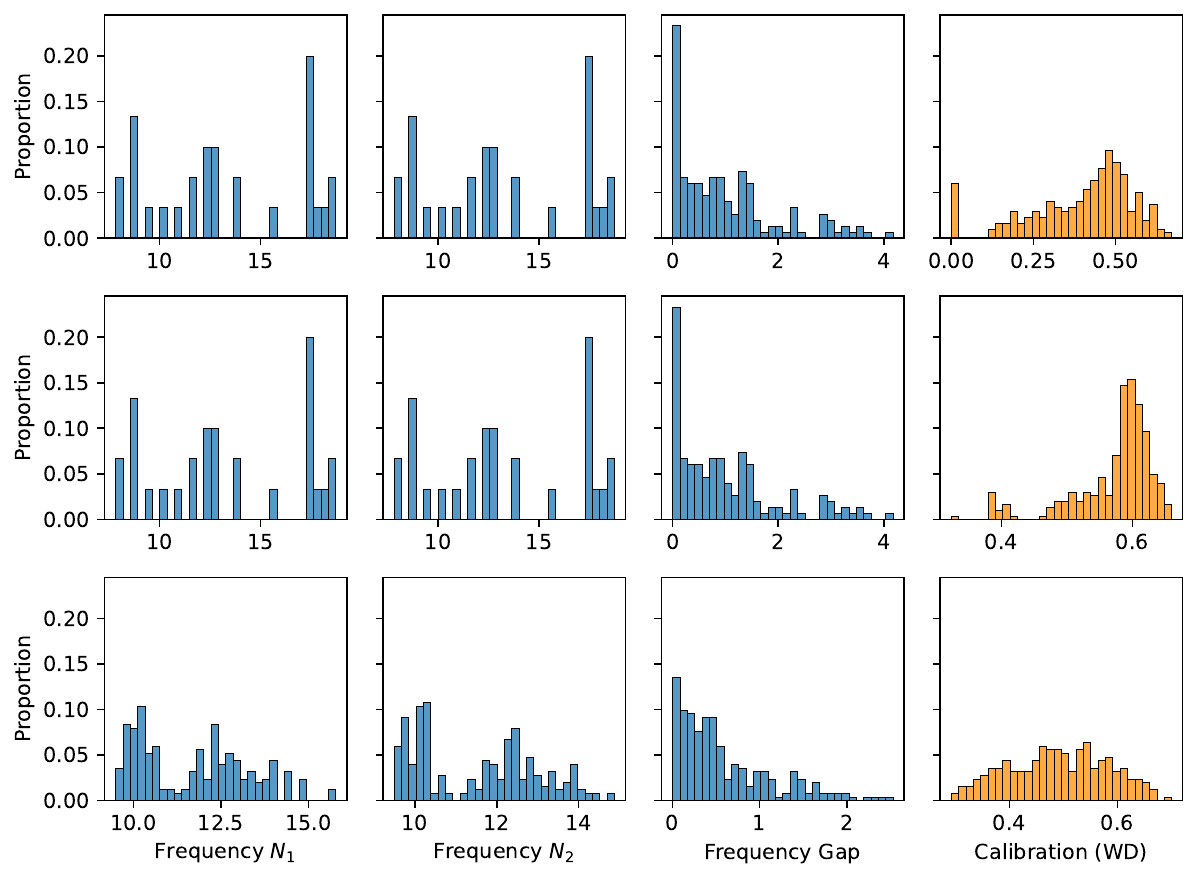}
 \caption{This plot shows dataset frequency statistics proportions for the number tokens used across the different datasets. We also present, highlighted orange, \gpt{4o}'s calibration score distribution. }
 \label{app:fig:frequency:stats}
\end{figure*}

\begin{figure*}[ht!]
 \centering
 \includegraphics[width=0.95\linewidth]{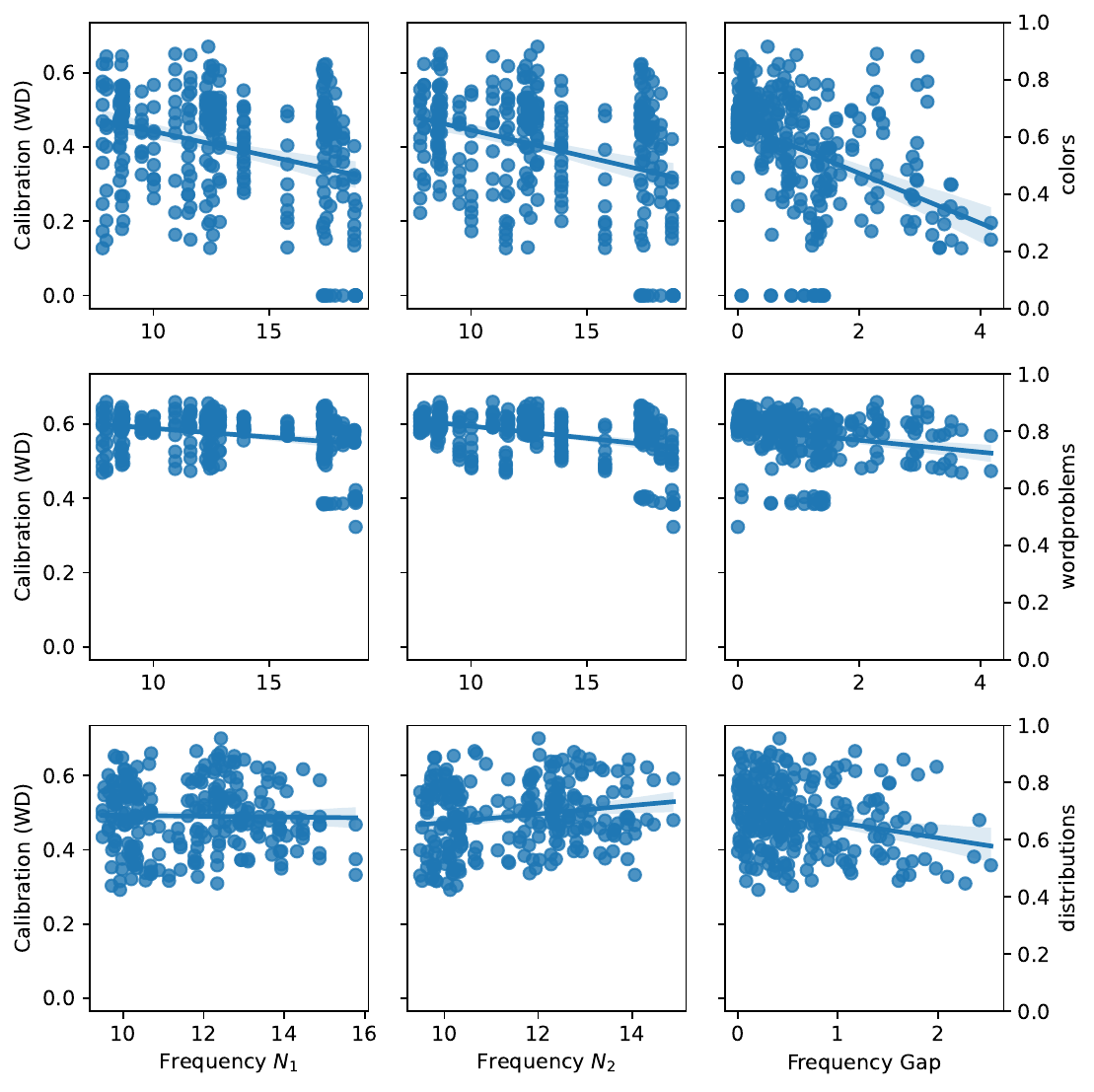}
 \caption{This plot shows \gpt{4o} data comparing calibration performance and different frequency data. Each row of results is across a different dataset.}
 \label{app:fig:frequency:corr}
\end{figure*}

\Cref{app:fig:frequency:stats} shows the frequency statistics of the number tokens used in our different datasets. \Cref{app:fig:frequency:corr} shows how those frequency stats relate to the calibration scores for \gpt{4o}. Not shown here is there is also an order effect (which number is more frequent).

\section{Human Study: Model vs Human Randomness}\label{app:human}
Our experiments demonstrated that language models are poorly calibrated. Here, we provide further context by comparing the proprietary model and human behavior, asking if models are as random (or not) as people. Previous work suggests that both would be \textit{not} random (see our Related Work section). We test this again. To get at this question, we use a variation of the two-player game Matching Pennies. We focus on a variation of the game similar to \citet{eliaz2011edgar}. Each round, the ``guesser'' aims to match her chosen coin face with ``misleader's'' choice, earning a +1 reward if their coin faces match, and -1 otherwise. The Nash Equilibrium is for both agents to act randomly. A calibrated model can trivially achieve random behavior.

\subsection{Experimental Design}\label{sec:exp:exp4}
This experiment is based on a simple heads/tails game. Our motivation for including this experiment is to allow a human baseline in a setting where random behavior is optimal. This section makes no distinction between calibrated probabilities and unbiased outputs because we don't have access to this information for humans. However, to ``perform'' well, or appear calibrated, requires either the model/person to (1) produce calibrated probabilities which when sampled from will yield random outputs, (2) use an internal random process to produce uncalibrated probabilities that yield random outputs, or (3) have a systematic process for producing outputs that \textit{appear} random enough. (Previous work would suggest that humans are limited in this regard and will likely not be random (and therefore not calibrated).) This leaves us with a one-sided problem. If the humans/models were proficient at producing random outcomes we would have to adjudicate between the above possibilities.

In a single-round game, the Nash Equilibrium is to select a coin face randomly. We set there to be 100 rounds of play. To focus on calibration, we frame the tested model as the ``misleader.'' The misleader is prompted to submit answers, and the  ``guesser'' is set to predict the misleader's answers. The misleader submits $m_1, m_2, \dots, m_{100}$ answers, and the guesser bases each of its $g_i$ answers incrementally based upon previous misleader's answers, $\{m_j\,|\, 0 < j < i\}$. We use a n-gram\footnote{Motivated by the strategy here: \url{https://www.expunctis.com/2019/03/07/Not-so-random.html}.} based strategy as our guesser.

\paragraph{Models} We focus on \texttt{gpt-*} models sampling temperature $\tau = 1$. Human behavior is drawn from a sample of 44 games of 100 rounds from $10$ different volunteers with technical backgrounds up to $10$ different games each. 

\Cref{lst:human:ngram} reports our the implementation for the n-gram model. We find that most n from 1 to 5 do similarly well. In the paper we report a 4-gram model.

\paragraph{Additional Metrics} Whereas previous metrics rely on model probabilities, this experiment assumes access only to outputs, meaning we cannot use the metrics used in previous sections.

\textbf{Player Win Rate (PWR)} is the rate the misleader wins, averaging over 100 games each for up to 100 rounds. The higher this rate, the more random the outputs of the misleader, and thus the more calibrated it is at an outcome level. 

\textbf{Randomness Testing (WW)}. Wald–Wolfowitz \citep{wald1940} tests if the elements of a sequence are mutually independent. Using this test at a significance level of 0.05, we report the proportion of samples that cannot statistically be rejected as non-random, labeling this value as \textbf{WW}. For reference, as shown in Table \ref{tab:exp4_arena} when we sampled random sequences using \texttt{python}, $98\%$ of the sequences could not be rejected; only $7\%$ of human-generated samples could not be rejected. 

\subsection{Results}
\begin{table}[ht!]
    \centering
    \begin{tabular}{lrr}
    \toprule
{} & $\textrm{PWR}_\uparrow$ & {$\textrm{WW}_\uparrow$} \\
    \midrule
    Random          &                    {0.50} & 0.98 \\
    Human           &                   {0.33} &  {0.07} \\
    \midrule
    gpt-3.5         &   \textbf{0.40} &  \textbf{0.43} \\
    gpt-4-turbo         &                 0.33 &    0.25 \\
    gpt-4o          &                   0.36 & 0.19 \\
    \bottomrule
    \end{tabular}
    \caption{\textbf{Human Study Results: Player win rates (PWR) and statistical tests for randomness (WW).} Higher values for both metrics suggest that models are more able to produce random (and thus more calibrated) outputs.
    \textbf{Takeaways:} \textbf{(1)} \texttt{gpt-3.5} stands out as the most random model; \textbf{(2)} Models pass a greater number of the statistical tests for randomness than humans, though are still far from random; \textbf{(3)} Model sequences are predictable at rates similar to humans. 
    }
    \label{tab:exp4_arena}
\end{table}


Both best-performing proprietary models and human behavior are not random. Models are prompted with the game rules and we explicitly instruct the models (and humans) to be random. Not doing so reduces model performance to near 0. PWR for models and humans are similar, WW for models is higher, but still low.

\Cref{tab:exp4_arena} reports the proportion of samples (where each sample is 100 coin flips) the result could not be considered non-random ($p = 0.05$): Higher WW values suggest the model is more random. More model- than human-generated samples were considered random: $19-43\%$ vs. $7\%$.

\clearpage
\onecolumn
\section*{\textsc{Continued Figures and Tables}}

\section{Figures: Reproducibility}

\begin{table}[ht!]
\small
 \centering
 \begin{tabular}{l|ll}
 \toprule
{Our Name} & HF Model Key (Base) & HF Model Key (Chat)  \\
\midrule
 Yi-1.5-9B   &  01-ai/Yi-1.5-9B &   01-ai/Yi-1.5-9B-Chat	\\
 Yi-1.5-34B   &  01-ai/Yi-1.5-34B &   01-ai/Yi-1.5-34B-Chat	\\
 Llama-3.1-8B   &  meta-llama/Llama-3.1-8B &   meta-llama/Llama-3.1-8B-Instruct	\\
 Mistral-7B-v0.1   &   mistralai/Mistral-7B-v0.1 &   mistralai/Mistral-7B-Instruct-v0.1	\\
 Mistral-7B-v0.3   &   mistralai/Mistral-7B-v0.3 &   mistralai/Mistral-7B-Instruct-v0.3	\\
 Mixtral-8x7B-v0.1   &   mistralai/Mixtral-8x7B-v0.1 &   mistralai/Mixtral-8x7B-Instruct-v0.1	\\
 gemma-2-9b & google/gemma-2-9b &  google/gemma-2-9b-it	\\
 gemma-2-27b & google/gemma-2-27b &  google/gemma-2-27b-it	\\
 \midrule
    
 & &OpenAI Model Key \\
\midrule
 gpt-3.5   & &  gpt-3.5-turbo-0125	\\
 gpt-4-turbo    &     &    gpt-4-turbo-2024-04-09	 \\
 gpt-4o    & & gpt-4o-2024-08-06	\\
 gpt-4o-mini  &  & gpt-4o-mini-2024-07-18	\\
 \bottomrule
 \end{tabular}
 \caption{\textbf{Models}}
 \label{tab:models}
\end{table}
\begin{figure*}[ht!]
\centering
\begin{tcolorbox}[enhanced,size=small,colback=black!5!white,colframe=RoyalBlue,flip title={interior hidden},title={We use the message format shared by OpenAI and HuggingFace. The problem \{text\} is injected into the prompt message.}]
\begin{lstlisting}[label={tab:prompts:prob}, captionpos=b, breaklines=true, language=Python]
# prompt 0
[
 {
 "role": "user",
 "content": "You are a perfectly calibrated language model that produces tokens with probabilities that correspond to values in the context and prompt. I will give you a scenario. Produce only a single word and nothing else. The word will be one of the options in context.",
 },
 {
 "role": "assistant",
 "content": "Excellent. I will continue by producing a single word after your next message.",
 },
 {"role": "user", "content": text},]
# prompt 1
[
 {
 "role": "user",
 "content": "You are a perfectly calibrated language model that produces tokens with probabilities that correspond to values in the context and prompt. I will give you a context where there are some probabilities involved. Produce one token.",
 },
 {
 "role": "assistant",
 "content": "Excellent. I will generate a single token calibrated to the statistics of the context.",
 },
 {"role": "user", "content": text},]
# prompt 2
[
 {
 "role": "user",
 "content": "You are a master of statistics and math. Produce tokens perfectly calibrated to the numbers in context. Produce one token.",
 },
 {"role": "assistant", "content": "Excellent. Ask away!"},
 {"role": "user", "content": text},]
# prompt 3
[
 {
 "role": "user",
 "content": "Produce tokens with probability calibrated to the numbers in context. Produce a single token.",
 },
 {"role": "assistant", "content": "Excellent. Ask away!"},
 {"role": "user", "content": text},]
# prompt 4
[
 {
 "role": "user",
 "content": "Produce tokens with probability calibrated to the statistics implicit and explicit in context. Produce only a single token.",
 },
 {"role": "assistant", "content": "Excellent. Ask away!"},
 {"role": "user", "content": text}, ],
\end{lstlisting}
\end{tcolorbox}
\caption{\textbf{Prompts For Instruction-Tuned Models; \texttt{Colors/Wordproblems}.}}
\end{figure*}
\begin{figure*}[ht!]
\centering
\begin{tcolorbox}[enhanced,size=small,colback=black!5!white,colframe=RoyalBlue,flip title={interior hidden},title={We use the message format shared by OpenAI and HuggingFace. The problem \{text\} is injected into the prompt message.}]
\begin{lstlisting}[label={tab:prompts:dist}, captionpos=b, breaklines=true, language=Python]
# prompt 0
[ 
 {
 "role": "user",
 "content": "You are a perfectly calibrated language model that produces tokens with probabilities that correspond to values in the context and prompt. I will give you a sentence. Finish the sentence by producing only a number and nothing else.",
 },
 {
 "role": "assistant",
 "content": "Excellent. I will continue by producing a single number after your next message.",
 },
 {"role": "user", "content": @text},
]
# prompt 1
[ 
 {
 "role": "user",
 "content": "You are a master of statistics and math. Produce tokens perfectly calibrated to the numbers in context.  I will give you a sentence. Produce one number to finish the sentence.",
 },
 {"role": "assistant", "content": "Excellent. Ask away!"},
 {"role": "user", "content": @text},
]
# prompt 2
[ 
 {
 "role": "user",
 "content": "Produce tokens with probability calibrated to the numbers in context. Given the scenario, produce a number.",
 },
 {"role": "assistant", "content": "Excellent. Go ahead!"},
 {"role": "user", "content": @text},
]
\end{lstlisting}
\end{tcolorbox}
\caption{\textbf{Prompts For Instruction-Tuned Models; \texttt{Distributions}.}}
\end{figure*}

\begin{figure*}[ht!]
\centering
\begin{tcolorbox}[enhanced,size=small,colback=black!5!white,colframe=RoyalBlue,flip title={interior hidden},title={The variables are injected into the text; \{number\_1\}, \{number\_2\}, \{option\_1\}, \{option\_2\}.}]
\begin{lstlisting}[label={tab:templates:prob}, captionpos=b, breaklines=true, language=Python]
# template 0
"There were {number_1} {option_1} marbles and {number_2} {option_2} marbles in a bag. Jane randomly picked a marble and saw it was the color"
# template 1
"Billy got to pick one marble from a big urn with many marbles. There were {number_1} {option_1} marbles and {number_2} {option_2} marbles in an urn. The color of the marble Billy randomly picked was"
# template 2
"Amanda had a huge pile of shirts. There were {number_1} {option_1} shirts and {number_2} {option_2} shirts. Without looking, she picked one by chance. The color of the shirt was"
# template 3
"Bill and Rick went to the hardwore store for paint in a hurry. The store had {number_1} shades of {option_1} and {number_2} shades of {option_2}. They didn't have any time to test out colors so they randomly grabbed a can. The color they grabbed turned out to be"
# template 4
"Kids at soccer practice randomly grabbed pinnies from a bag. There were {number_1} {option_1} pinnies and {number_2} {option_2} pinnies. Tommy's pinny was the color"
\end{lstlisting}
\end{tcolorbox}
\caption{\textbf{Templates for \texttt{Probabilities}.}}
\end{figure*}
\begin{figure*}[h!]
\centering
\begin{tcolorbox}[enhanced,size=small,colback=black!5!white,colframe=RoyalBlue,flip title={interior hidden},title={The variables are injected into the text; \{number\_1\}, \{number\_2\}, \{prefix\}. \{prefix\} is always set to empty; early pilot experiments suggest that using an ellipsis leads to a similar pattern of results.}]
\begin{lstlisting}[label={tab:templates:dist}, captionpos=b, breaklines=true, language=Python]
# template 0
"Picking an integer between {number_1} inclusive and {number_2} exclusive, the integer I got was{prefix}"
# template 1
"Picking an integer between {number_1} and {number_2}, exclusive on both sides, the integer I got was{prefix}"
# template 2
"I sampled an integer from a random variable distributed uniformly between {number_1} inclusive and {number_2} exclusive, getting the value{prefix}"
# template 3
"I sampled an integer from a random variable distributed uniformly between {number_1} and {number_2}, exclusive on both ends, getting the value{prefix}"
\end{lstlisting}
\end{tcolorbox}
\caption{\textbf{Templates for \texttt{Distributions}.}}
\end{figure*}

\clearpage
\section{Figures: What Do Metric Values Mean?}\label{app:results:demo:figures}
\begin{table}[ht!]
    \centering
    \small
\begin{tabular}{lrrrrr}
\toprule
&\multicolumn{5}{c}{\textit{Settings: }\texttt{51 \textcolor{PURPLE}{\textbf{purple}}, 98 \textbf{white}}}\\
\midrule
{} &    $P(t_1)$ &    $P(t_2)$ &  PM &  RE &  WD \\
\midrule
Ideal & 0.342 & 0.657 & 1.000& 1.000& 0.000 \\
\midrule
Pick First                &  1.000 &  0.000 &      1.000&            -0.927 &          0.930 \\
Pick Second               &  0.000 &  1.000 &      1.000&            -0.927 &          0.484 \\
Pick Higher$_{p =0.7}$    &  0.300 &  0.700 &      1.000&            -0.046 &          0.060 \\
Pick Higher$_{p =0.6}$    &  0.400 &  0.600 &      1.000&             0.044 &          0.082 \\
Pick Higher$_{p =0.8}$    &  0.200 &  0.800 &      1.000&            -0.205 &          0.201 \\
Pick Higher$_{p =0.9}$    &  0.100 &  0.900 &      1.000&            -0.458 &          0.343 \\
Pick Higher               &  0.000 &  1.000 &      1.000&            -0.927 &          0.484 \\
Pick Lower                &  1.000 &  0.000 &      1.000&            -0.927 &          0.930 \\
\midrule
Meta-Llama-3.1-8B &  0.687 &  0.279 &    0.966 &            -0.060 &          0.512 \\
Mistral-7B-v0.1   &  0.698 &  0.024 &    0.723 &            -0.715 &          0.727 \\
Mistral-7B-v0.3   &  0.938 &  0.000 &    0.938 &            -0.926 &          0.887 \\
Mixtral-8x7B-v0.1 &  1.000 &  0.000 &    1.000 &            -0.927 &          0.930 \\
Yi-1.5-34B        &  0.581 &  0.010 &    0.591 &            -0.800 &          0.690 \\
Yi-1.5-9B         &  0.999 &  0.000 &    0.999 &            -0.924 &          0.929 \\
gpt-4-turbo       &  0.014 &  0.985 &    1.000 &            -0.818 &          0.464 \\
gpt-4o            &  0.245 &  0.755 &    1.000 &            -0.124 &          0.137 \\
gpt-4o-mini       &  0.998 &  0.002 &    1.000 &            -0.902 &          0.927 \\
\bottomrule
\end{tabular}
\end{table}

\begin{table}[ht!]
    \centering
    \small
\begin{tabular}{lrrrrr}
\toprule
&\multicolumn{5}{c}{\textit{Settings: }\texttt{51 \textbf{white}, 98 \textcolor{PURPLE}{\textbf{purple}}}}\\
\midrule
{} &    $P(t_1)$ &    $P(t_2)$ &  PM &  RE &  WD \\
\midrule
Ideal & 0.342 & 0.657 & 1.000& 1.000& 0.000 \\
\midrule
Meta-Llama-3.1-8B &  0.027 &  0.955 &    0.982 &            -0.747 &          0.433 \\
Mistral-7B-v0.1   &  0.304 &  0.361 &    0.665 &             0.068 &          0.299 \\
Mistral-7B-v0.3   &  0.035 &  0.886 &    0.921 &            -0.694 &          0.383 \\
Mixtral-8x7B-v0.1 &  0.101 &  0.899 &    1.000 &            -0.455 &          0.341 \\
Yi-1.5-34B        &  0.179 &  0.147 &    0.326 &             0.066 &          0.536 \\
Yi-1.5-9B         &  0.951 &  0.048 &    0.999 &            -0.649 &          0.862 \\
gpt-4-turbo       &  0.000 &  1.000 &    1.000 &            -0.927 &          0.484 \\
gpt-4o            &  0.095 &  0.905 &    1.000 &            -0.473 &          0.349 \\
gpt-4o-mini       &  0.095 &  0.905 &    1.000 &            -0.473 &          0.349 \\
\bottomrule
\end{tabular}
\caption{Each row is a single result for a single setting. The first row shows the ideal, calibrated result. The first section of each tables shows baseline values, which are helpful for understanding the practical ranges of RE and WD values. The subsequent rows show how different models behave. These values, see in particular \gpt{4o-mini}, line up with the summary behaviors seen in \Cref{fig:hard-fig} and Appendix \Cref{app:fig:gpt-4o-mini-distance}. In the second section we invert the order of the color options. No model produces a well-calibrated result for this problem.}
\label{app:tab:more:best:b}
\end{table}

\clearpage
\section{Figures: More Results on Is It Possible For The Models To Do Well?}\label{app:results:demo:metrics}

\clearpage

\section{{Figures: More Results on Calibration And Low Probability Mass}}\label{app:results:lowpm:figures}

\begin{table*}[ht!]
\small
\centering
\begin{tabular}{lll|ll|ll}
\toprule
                            &   \multicolumn{2}{c}{\textit{colors}} &                   \multicolumn{2}{c}{\textit{wordproblems}}       &               \multicolumn{2}{c}{\textit{distributions}}                \\
\midrule
Pick Higher      &  \multicolumn{2}{c|}{0.47$_{\pm0.01}$} &  \multicolumn{2}{c|}{0.44$_{\pm0.00}$} &                                   \multicolumn{2}{c}{-} \\
Pick $\textrm{Higher}_\textrm{p=0.7}$ &  \multicolumn{2}{c|}{0.15$_{\pm0.01}$} &  \multicolumn{2}{c|}{0.17$_{\pm0.00}$} &                                   \multicolumn{2}{c}{-} \\
Pick Lower       &  \multicolumn{2}{c|}{0.95$_{\pm0.01}$} &  \multicolumn{2}{c|}{0.98$_{\pm0.00}$} &                                   \multicolumn{2}{c}{-} \\
Pick First/Second       &  \multicolumn{2}{c|}{0.71$_{\pm0.02}$} &  \multicolumn{2}{c|}{0.71$_{\pm0.00}$} &                                   \multicolumn{2}{c}{0.38$_{\pm0.00}$} \\
$\textrm{Random}_{\tau = 0.01}$    &  \multicolumn{2}{c|}{0.71$_{\pm0.02}$} &  \multicolumn{2}{c|}{0.69$_{\pm0.00}$} &  \multicolumn{2}{c}{0.86$_{\pm0.00}$} \\
$\textrm{Random}_{\tau = 1.0}$    &  \multicolumn{2}{c|}{0.27$_{\pm0.01}$} &  \multicolumn{2}{c|}{0.29$_{\pm0.00}$} &  \multicolumn{2}{c}{0.38$_{\pm0.00}$} \\
\midrule

& Base & Chat & Base & Chat & Base & Chat \\
\midrule
Llama-3.1-8B & 0.55$_{\pm0.00}$ & 0.40$_{\pm0.01}$ & 0.53$_{\pm0.03}$ & 0.48$_{\pm0.02}$ & 0.17$_{\pm0.00}$ & 0.43$_{\pm0.01}$ \\
  Mistral-7B-v0.1 & 0.60$_{\pm0.00}$ & 0.50$_{\pm0.01}$ & 0.55$_{\pm0.03}$ & 0.49$_{\pm0.01}$ & 0.13$_{\pm0.00}$ & 0.22$_{\pm0.01}$ \\
  Mistral-7B-v0.3 & 0.58$_{\pm0.00}$ & 0.48$_{\pm0.02}$ & 0.57$_{\pm0.03}$ & 0.55$_{\pm0.01}$ & 0.13$_{\pm0.00}$ & 0.56$_{\pm0.01}$ \\
Mixtral-8x7B-v0.1 & 0.56$_{\pm0.00}$ & 0.51$_{\pm0.01}$ & 0.51$_{\pm0.03}$ & 0.60$_{\pm0.03}$ & 0.16$_{\pm0.01}$ & 0.71$_{\pm0.01}$ \\
       Yi-1.5-34B & 0.54$_{\pm0.01}$ & 0.55$_{\pm0.01}$ & 0.56$_{\pm0.04}$ & 0.57$_{\pm0.01}$ & 0.21$_{\pm0.01}$ & 0.62$_{\pm0.01}$ \\
        Yi-1.5-9B & 0.60$_{\pm0.01}$ & 0.49$_{\pm0.01}$ & 0.53$_{\pm0.02}$ & 0.56$_{\pm0.01}$ & 0.18$_{\pm0.01}$ & 0.61$_{\pm0.01}$ \\
      gemma-2-27b & 0.47$_{\pm0.00}$ & 0.40$_{\pm0.00}$ & 0.50$_{\pm0.04}$ & 0.48$_{\pm0.01}$ & 0.13$_{\pm0.01}$ & 0.59$_{\pm0.01}$ \\
       gemma-2-9b & 0.54$_{\pm0.00}$ & 0.50$_{\pm0.01}$ & 0.48$_{\pm0.03}$ & 0.52$_{\pm0.02}$ & 0.14$_{\pm0.00}$ & 0.59$_{\pm0.01}$ \\
          gpt-3.5 &                 - & 0.30$_{\pm0.01}$ &                 - & 0.57$_{\pm0.02}$ &                 - & 0.42$_{\pm0.01}$ \\
      gpt-4-turbo &                 - & 0.42$_{\pm0.01}$ &                 - & 0.62$_{\pm0.02}$ &                 - & 0.69$_{\pm0.01}$ \\
           gpt-4o &                 - & 0.40$_{\pm0.00}$ &                 - & 0.57$_{\pm0.02}$ &                 - & 0.49$_{\pm0.01}$ \\
      gpt-4o-mini &                 - & 0.40$_{\pm0.01}$ &                 - & 0.57$_{\pm0.02}$ &                 - & 0.57$_{\pm0.01}$ \\
\bottomrule
\end{tabular}

\caption{\textbf{Calibration Results.} Across two datasets, this plot shows the calibration scores (WassersteinSistance, WD) for chat versions of models along with 95\% confidence intervals. (Lower Better). For \textit{colors}, six models have a statistically better calibration than the baseline of always picking the option associated with the higher number, denoted \spade. For \textit{wordproblems}, no model is better than this simple baseline. For \textit{distributions}, the Base models do better, but this partially because models are have low probability mass on all the options.
}
\label{tab:results:wd:all}
\end{table*}

\clearpage
\section{{Figures: More Results on Mode Collapse}}\label{app:results:mode:figures}

\begin{table*}[ht!]
\small
\centering

\begin{tabular}{llll}
\toprule
{} &               Base &               Chat &                 CI($\Delta$) \\
\midrule
\textit{colors}\\
Meta-Llama-3.1-8B &   0.08$_{\pm0.01}$ &  -0.11$_{\pm0.02}$ &  {\scriptsize(-0.21, -0.17)} \\
Mistral-7B-v0.1   &   0.10$_{\pm0.01}$ &  -0.21$_{\pm0.02}$ &  {\scriptsize(-0.33, -0.29)} \\
Mistral-7B-v0.3   &   0.09$_{\pm0.01}$ &  -0.36$_{\pm0.03}$ &  {\scriptsize(-0.47, -0.42)} \\
Mixtral-8x7B-v0.1 &   0.06$_{\pm0.01}$ &  -0.67$_{\pm0.01}$ &  {\scriptsize(-0.74, -0.72)} \\
Yi-1.5-9B         &   0.06$_{\pm0.02}$ &  -0.55$_{\pm0.01}$ &  {\scriptsize(-0.63, -0.59)} \\
Yi-1.5-34B        &   0.02$_{\pm0.02}$ &  -0.45$_{\pm0.02}$ &  {\scriptsize(-0.49, -0.45)} \\
gemma-2-9b        &   0.13$_{\pm0.01}$ &  -0.44$_{\pm0.01}$ &  {\scriptsize(-0.58, -0.56)} \\
gemma-2-27b       &   0.06$_{\pm0.01}$ &  -0.63$_{\pm0.01}$ &   {\scriptsize(-0.7, -0.69)} \\
gpt-3.5           &                  - &  -0.35$_{\pm0.03}$ &               {\scriptsize-} \\
gpt-4-turbo       &                  - &  -0.65$_{\pm0.01}$ &               {\scriptsize-} \\
gpt-4o-mini       &                  - &  -0.53$_{\pm0.01}$ &               {\scriptsize-} \\
gpt-4o            &                  - &  -0.61$_{\pm0.00}$ &               {\scriptsize-} \\
\midrule
\textit{wordproblems}\\
Meta-Llama-3.1-8B &  -0.11$_{\pm0.04}$ &  -0.33$_{\pm0.06}$ &  {\scriptsize(-0.27, -0.17)} \\
Mistral-7B-v0.1   &  -0.10$_{\pm0.05}$ &  -0.24$_{\pm0.04}$ &   {\scriptsize(-0.18, -0.1)} \\
Mistral-7B-v0.3   &  -0.12$_{\pm0.05}$ &  -0.47$_{\pm0.04}$ &  {\scriptsize(-0.37, -0.33)} \\
Mixtral-8x7B-v0.1 &  -0.13$_{\pm0.07}$ &  -0.65$_{\pm0.05}$ &  {\scriptsize(-0.59, -0.47)} \\
Yi-1.5-9B         &  -0.11$_{\pm0.05}$ &  -0.55$_{\pm0.03}$ &   {\scriptsize(-0.48, -0.4)} \\
Yi-1.5-34B        &  -0.17$_{\pm0.05}$ &  -0.53$_{\pm0.04}$ &   {\scriptsize(-0.4, -0.32)} \\
gemma-2-9b        &  -0.07$_{\pm0.04}$ &  -0.45$_{\pm0.03}$ &  {\scriptsize(-0.41, -0.34)} \\
gemma-2-27b       &  -0.07$_{\pm0.06}$ &  -0.57$_{\pm0.02}$ &  {\scriptsize(-0.56, -0.44)} \\
gpt-3.5           &                  - &  -0.68$_{\pm0.04}$ &               {\scriptsize-} \\
gpt-4-turbo       &                  - &  -0.69$_{\pm0.04}$ &               {\scriptsize-} \\
gpt-4o-mini       &                  - &  -0.58$_{\pm0.05}$ &               {\scriptsize-} \\
gpt-4o            &                  - &  -0.63$_{\pm0.04}$ &               {\scriptsize-} \\
\midrule
\textit{distributions}\\
Meta-Llama-3.1-8B &   0.34$_{\pm0.02}$ &  -0.70$_{\pm0.03}$ &  {\scriptsize(-1.06, -1.01)} \\
Mistral-7B-v0.1   &   0.24$_{\pm0.01}$ &   0.05$_{\pm0.02}$ &  {\scriptsize(-0.21, -0.18)} \\
Mistral-7B-v0.3   &   0.25$_{\pm0.01}$ &  -1.07$_{\pm0.04}$ &  {\scriptsize(-1.36, -1.28)} \\
Mixtral-8x7B-v0.1 &   0.04$_{\pm0.01}$ &  -1.59$_{\pm0.04}$ &  {\scriptsize(-1.67, -1.59)} \\
Yi-1.5-9B         &   0.19$_{\pm0.01}$ &  -1.13$_{\pm0.05}$ &  {\scriptsize(-1.36, -1.27)} \\
Yi-1.5-34B        &   0.04$_{\pm0.01}$ &  -1.32$_{\pm0.04}$ &   {\scriptsize(-1.4, -1.33)} \\
gemma-2-9b        &   0.34$_{\pm0.02}$ &  -1.10$_{\pm0.04}$ &  {\scriptsize(-1.49, -1.39)} \\
gemma-2-27b       &   0.09$_{\pm0.01}$ &  -1.14$_{\pm0.05}$ &  {\scriptsize(-1.28, -1.19)} \\
gpt-3.5           &                  - &  -0.69$_{\pm0.02}$ &               {\scriptsize-} \\
gpt-4-turbo       &                  - &  -1.54$_{\pm0.03}$ &               {\scriptsize-} \\
gpt-4o-mini       &                  - &  -1.30$_{\pm0.02}$ &               {\scriptsize-} \\
gpt-4o            &                  - &  -0.92$_{\pm0.02}$ &               {\scriptsize-} \\
\bottomrule
\end{tabular}
\caption{\textbf{Relative Entropy (RE) Results.} The RE is far below a calibrated level across all datasets. That being said, on all datasets, placing all probability mass on a single option (Pick First) has lower RE than models' behavior, suggesting that models have some success. This table also shows how this large drop in entropy occurred with instruction tuning. See \Cref{tab:results:re} for the subset of results we show in the main body of the paper. \textbf{\Cref{tab:results:re} captured differences in entropy relative to the target calibrated entropy. Here, in this table, the rightmost column captures the difference in entropy between the Base and Chat models.} 
}
\label{app:tab:results:re}
\end{table*}
\begin{table*}[ht!]
\small
\centering
\begin{tabular}{lrrr}
\toprule
{} & \multicolumn{2}{c}{Mode Prob} &  \multicolumn{1}{c}{Mode} \\
&         Base & Chat &  Stability \\
\midrule
Idealized & 0.23 & 0.23 & 0.05\\
\midrule
Meta-Llama-3.1-8B &  0.25 & 0.60 & 0.48\\
Mistral-7B-v0.1   &  0.26 & 0.33 & 0.58 \\
Mistral-7B-v0.3   &            0.26 &          0.65 &            0.35\\
Mixtral-8x7B-v0.1 &       0.32 &          0.82 &            0.52\\
Yi-1.5-34B         &          0.75 &            0.53 \\
Yi-1.5-9B         &        0.30 &          0.75 &            0.49\\
gemma-2-27b       &            0.29 &          0.68 &            0.46 \\
gemma-2-9b        &            0.22 &          0.66 &            0.40 \\
\midrule
$\textrm{AVG}_{\textrm{OpenSource}}$ &0.28 & 0.66 & 0.48\\
\midrule
gpt-3.5           & - & 0.53 & -\\
gpt-4-turbo       & - & 0.79 & - \\
gpt-4o            &  - & 0.58 & - \\
gpt-4o-mini       &  - & 0.73 & - \\
\midrule
$\textrm{AVG}_{\textrm{Proprietary}}$  & - & 0.66 & - \\
\bottomrule
\end{tabular}
\caption{\textbf{\textit{Distributions} Results.} Mode-related metrics. \textbf{(1)} Averaged, the probability mass on the top-token for chat models is $43\%$ above the calibrated ideal, showing where relative entropy increased. \textbf{(2)} The top-tokens remain the same across base and chat models for 48\% of instances. These results together suggest mode collapse.}
\label{app:tab:dist:mode}
\end{table*}
\clearpage

\section{{Figures: More Results on Ordering and Colors}}\label{app:results:ordering:figures}

\begin{table*}[ht!]
\small
\centering

\begin{tabular}{@{}l@{\hspace{4pt}}l@{\hspace{4pt}}r@{\hspace{4pt}}r@{\hspace{4pt}}r@{\hspace{4pt}}r@{}}
\toprule

&& Spearman's Corr & PValue & Kendall's Tau & PValue \\
\midrule
      gpt-4-turbo &            gpt-4o &               \RC{-0.61} &         8.83e-13 &                \RC{-0.45} &          2.08e-06 \\
      gpt-4-turbo &       gpt-4o-mini &               \RC{-0.58} &         3.37e-11 &                \RC{-0.35} &          3.10e-04 \\
       gemma-2-9b &       gpt-4-turbo &               \RC{-0.49} &         6.90e-08 &                \RC{-0.27} &          4.41e-03 \\
  Mistral-7B-v0.1 & Mixtral-8x7B-v0.1 &               \RC{-0.44} &         1.46e-06 &                \RC{-0.35} &          3.10e-04 \\
      gemma-2-27b &       gpt-4-turbo &               \RC{-0.40} &         1.51e-05 &                \RC{-0.42} &          1.27e-05 \\
Mixtral-8x7B-v0.1 &       gpt-4-turbo &               \RC{-0.36} &         1.10e-04 &                \RC{-0.16} &          8.76e-02 \\
Meta-Llama-3.1-8B &       gpt-4-turbo &               \RC{-0.34} &         2.64e-04 &                \RC{-0.31} &          1.25e-03 \\
  Mistral-7B-v0.3 &        gemma-2-9b &               \RC{-0.34} &         3.10e-04 &                \RC{-0.09} &          3.43e-01 \\
  Mistral-7B-v0.3 &        Yi-1.5-34B &               \RC{-0.32} &         6.85e-04 &                \RC{-0.38} &          6.71e-05 \\
Mixtral-8x7B-v0.1 &         Yi-1.5-9B &               \RC{-0.30} &         1.41e-03 &                \RC{-0.20} &          3.68e-02 \\
  Mistral-7B-v0.3 &       gemma-2-27b &               \RC{-0.29} &         1.89e-03 &                \RC{-0.38} &          6.71e-05 \\
Meta-Llama-3.1-8B &        gemma-2-9b &               \RC{-0.26} &         6.73e-03 &                \RC{-0.35} &          3.10e-04 \\
  Mistral-7B-v0.1 &        Yi-1.5-34B &               \RC{-0.25} &         7.97e-03 &                \RC{-0.31} &          1.25e-03 \\
        Yi-1.5-9B &        gemma-2-9b &                 \GC{+0.26} &         6.30e-03 &                  \GC{+0.20} &          3.68e-02 \\
          gpt-3.5 &            gpt-4o &                 \GC{+0.28} &         3.37e-03 &                  \GC{+0.31} &          1.25e-03 \\
Meta-Llama-3.1-8B &        Yi-1.5-34B &                 \GC{+0.33} &         4.45e-04 &                  \GC{+0.31} &          1.25e-03 \\
Meta-Llama-3.1-8B &       gemma-2-27b &                 \GC{+0.33} &         3.97e-04 &                  \GC{+0.09} &          3.43e-01 \\
       gemma-2-9b &       gemma-2-27b &                 \GC{+0.33} &         3.69e-04 &                  \GC{+0.20} &          3.68e-02 \\
      gemma-2-27b &            gpt-4o &                 \GC{+0.34} &         2.83e-04 &                  \GC{+0.31} &          1.25e-03 \\
        Yi-1.5-9B &       gpt-4o-mini &                 \GC{+0.42} &         4.90e-06 &                  \GC{+0.42} &          1.27e-05 \\
  Mistral-7B-v0.1 &         Yi-1.5-9B &                 \GC{+0.42} &         4.08e-06 &                  \GC{+0.27} &          4.41e-03 \\
Mixtral-8x7B-v0.1 &        gemma-2-9b &                 \GC{+0.42} &         3.88e-06 &                  \GC{+0.31} &          1.25e-03 \\
      gpt-4o-mini &            gpt-4o &                 \GC{+0.46} &         4.08e-07 &                  \GC{+0.24} &          1.36e-02 \\
Mixtral-8x7B-v0.1 &            gpt-4o &                 \GC{+0.47} &         2.04e-07 &                  \GC{+0.42} &          1.27e-05 \\
  Mistral-7B-v0.1 &       gpt-4o-mini &                 \GC{+0.48} &         1.50e-07 &                  \GC{+0.56} &          3.99e-09 \\
       gemma-2-9b &       gpt-4o-mini &                 \GC{+0.51} &         1.10e-08 &                  \GC{+0.42} &          1.27e-05 \\
       Yi-1.5-34B &       gemma-2-27b &                 \GC{+0.57} &         7.59e-11 &                  \GC{+0.42} &          1.27e-05 \\
       gemma-2-9b &            gpt-4o &                 \GC{+0.62} &         4.33e-13 &                  \GC{+0.53} &          3.69e-08 \\
\bottomrule
\end{tabular}
\caption{Correlation matrix between model rankings. We show only the statistically significant correlations, others had both lower effect sizes and were not significant. In most cases both Kendall Tau and Spearman were significant; there are two exceptions where for $p=0.01$ Kendall Tau was not significant where Spearman was significant.}
\label{app:tab:rankings_corr}
\end{table*}

\begin{figure*}[ht!]
 \centering \includegraphics[width=0.85\linewidth]{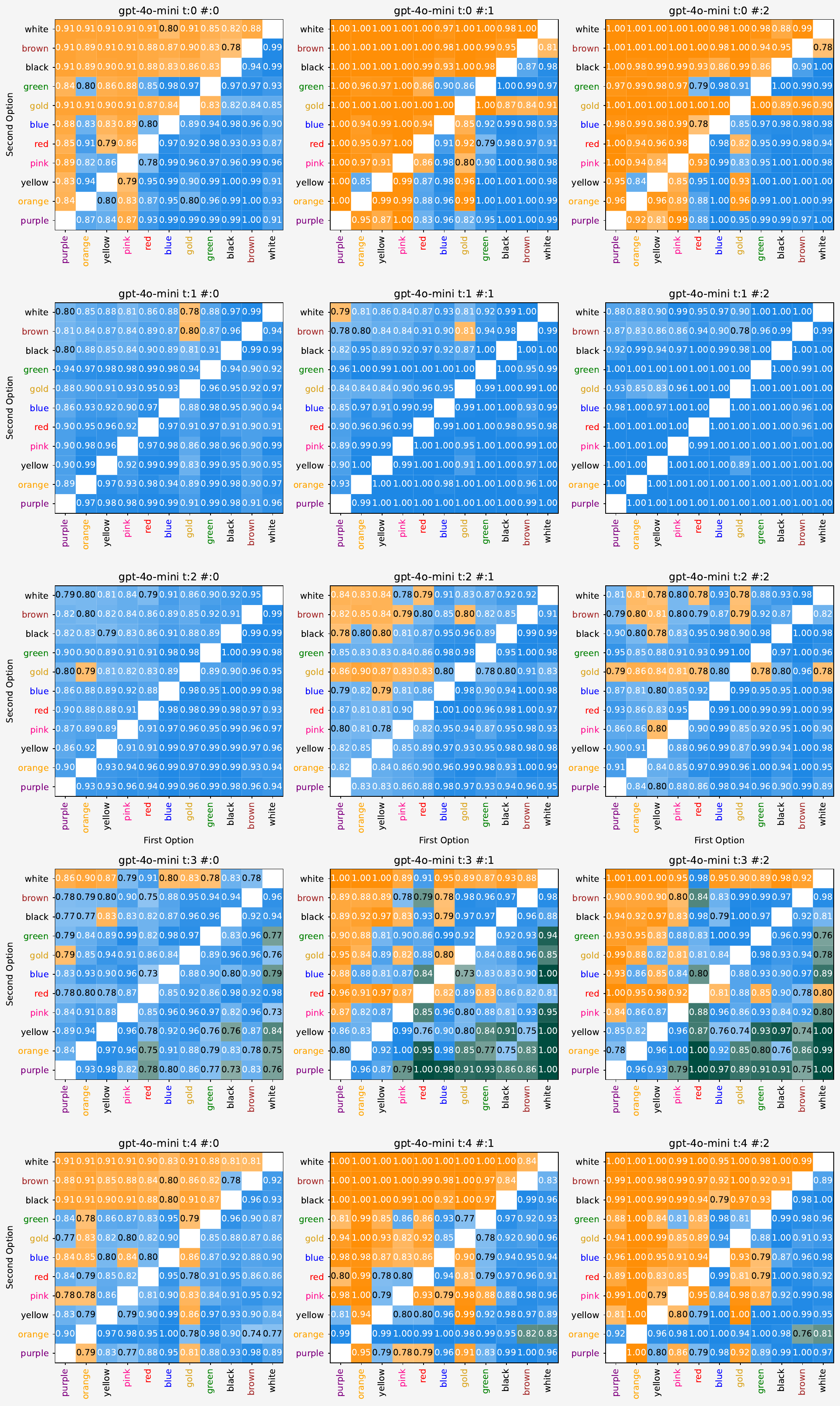}
 \label{app:fig:gpt-4o-mini-rate}
 \caption{\textbf{Behavior Compatibility-based Heatmap for \gpt{4o-mini}.} 
 }
 \end{figure*}

\clearpage
\section{{Figures: More Results on Number Frequency}}\label{app:results:frequency:figures}

\clearpage
\section{{Figures: Human Experiments}}\label{app:results:hhuman:figures}

\begin{figure*}[h!]
\centering
\begin{tcolorbox}[enhanced,size=small,colback=black!5!white,colframe=RoyalBlue,flip title={interior hidden},title={n-gram model implementation}]
\begin{lstlisting}[captionpos=b, breaklines=true, language=Python]
CoinFace = str, H: CoinFace = "H", T: CoinFace = "T"
def flip(face: CoinFace) -> CoinFace:
    return {H: T, T: H}[face]
def mdp_1gram(sequence: List[CoinFace]) -> CoinFace:
    return _mdp_ngram(sequence, 1)
def mdp_2gram(sequence: List[CoinFace]) -> CoinFace:
    return _mdp_ngram(sequence, 2)
def mdp_3gram(sequence: List[CoinFace]) -> CoinFace:
    return _mdp_ngram(sequence, 3)
def mdp_4gram(sequence: List[CoinFace]) -> CoinFace:
    return _mdp_ngram(sequence, 4)
def mdp_5gram(sequence: List[CoinFace]) -> CoinFace:
    return _mdp_ngram(sequence, 5)
def _mdp_ngram(sequence: List[CoinFace], ngram_size: int) -> CoinFace:
    """Picks the most common continuation for a given sequence based upon all n_grams of the current size.

    Defaults to H if sequence is empty. Flips previous coinface if the length of the sequence is 1.
    """
    assert ngram_size >= 1

    # Handle short sequences
    if len(sequence) == 0:
        return H
    elif len(sequence) == 1:
        if sequence[0] == H:
            return T
        else:
            return H
    if len(sequence) <= ngram_size:
        return _mdp_ngram(sequence, min(ngram_size - 1, len(sequence) - 1))

    map_ = {H: 0, T: 1}
    sequence_numbers = [map_[face] for face in sequence]
    mdp = np.zeros([2 for _ in range(ngram_size + 1)])
    for i in range(0, len(sequence_numbers) - ngram_size):
        index = tuple(sequence_numbers[i : i + (ngram_size + 1)])
        mdp[index] = mdp[index] + 1

    last = tuple(sequence_numbers[-ngram_size:])
    transitions = mdp[last]
    if transitions[0] > transitions[1]:
        return H
    elif transitions[0] == transitions[1]:
        # NOTE: Alt, we could pick which ever is more common
        # or fall back to a lower ngram or sample.
        return H
    else:
        return T
\end{lstlisting}
\end{tcolorbox}
\caption{\textbf{\texttt{n-gram} model implementation.}}
\label{lst:human:ngram}
\end{figure*}

\end{document}